\documentclass[a4paper]{fairmeta}

\usepackage{hyperref}
\usepackage{url}
\usepackage{float}
\usepackage{booktabs}
\usepackage{tabularx}
\usepackage{multirow}
\usepackage{xspace}
\usepackage{graphicx}
\usepackage{wrapfig}
\usepackage{sidecap}

 \crefname{figure}{Figure}{figures}
 \crefname{section}{Sec.}{Secs.}
 \Crefname{section}{Section}{Sections}
 \crefname{table}{Table}{tables}
\crefname{appendix}{Appendix}{Appendices}
\crefname{equation}{Equation}{equations}

\def\mypar#1{\noindent\textbf{#1.}\hspace{0mm}}

\usepackage{amsmath,amsfonts,bm}

\def\eqref#1{equation~\ref{#1}}

\def\1{\bm{1}}

\def\vh{{\bm{h}}}

\def\vs{{\bm{s}}}

\def\vv{{\bm{v}}}

\def\vx{{\bm{x}}}

\def\vz{{\bm{z}}}

\def\mH{{\bm{H}}}

\DeclareMathAlphabet{\mathsfit}{\encodingdefault}{\sfdefault}{m}{sl}
\SetMathAlphabet{\mathsfit}{bold}{\encodingdefault}{\sfdefault}{bx}{n}

\newcommand{\E}{\mathbb{E}}

\makeatletter
\DeclareRobustCommand\onedot{\futurelet\@let@token\@onedot}
\def\@onedot{\ifx\@let@token.\else.\null\fi\xspace}

\def\eg{\emph{e.g}\onedot} 
\def\ie{\emph{i.e}\onedot} 
 
 \def\vs{\emph{vs}\onedot}
\def\wrt{w.r.t\onedot} 

\makeatother

\def \ours {{PerCo}\xspace}

\title{Towards image compression with perfect realism at ultra-low bitrates}

\author[1,2]{Marlène Careil}
\author[1]{Matthew J. Muckley}
\author[1]{Jakob Verbeek}
\author[2]{Stéphane Lathuilière}

\affiliation[1]{FAIR at Meta}
\affiliation[2]{LTCI, Télécom Paris, IP Paris}

\abstract{Image  codecs are typically optimized to trade-off bitrate \vs distortion metrics.
At low bitrates, this leads to  compression artefacts which are easily perceptible, even  when training with  perceptual  or adversarial losses.
To improve image quality and remove dependency on the bitrate 
 we propose to decode with iterative diffusion models.
 We condition the decoding process on a vector-quantized image representation, as well as a global image description to provide additional context.
We dub our model ``\ours'' for ``perceptual compression'', and compare it to state-of-the-art codecs at rates from 0.1 down to  0.003 bits per pixel.
The latter rate  is more than an order of magnitude smaller than those considered in most prior work, compressing a $512\!\times\!768$ Kodak image with less than 153 bytes. Despite this ultra-low bitrate, our approach maintains the ability to reconstruct realistic images.
We find that our model leads to reconstructions with state-of-the-art visual quality as measured by FID and KID.
As predicted by rate-distortion-perception theory, visual quality is less dependent on the bitrate than previous methods.

}

\date{\today}
\correspondence{First Author at \email{marlene.careil@protonmail.com}}

\metadata[Note]{To appear at ICLR 2024.}

\begin{document}

\maketitle

\section{Introduction}

Traditional image and video codecs are optimized for the rate-distortion function~\citep{shannon1948mathematical}, which minimizes the expected size of the data under a distortion constraint such as mean-squared error.
Recent research has developed neural image compression methods that surpass handcrafted image compression codecs in terms of rate-distortion performance~\citep{balle2017end,balle2018variational,cheng2020joint,he2022elic}.
However, optimization for the rate-distortion function comes at a cost of ``realism'', where realism is mathematically related to the statistical fidelity or $f$-divergence between the compressed image distribution and the true image distribution~\citep{blau2018perception,blau2019rethinking}.
The typical qualitative manifestation of unrealistic images is blurring. 

\begin{figure}
\def\myim#1{ \includegraphics[width=.195\textwidth,height=.13\textwidth]{figs/main_fig/kodak/#1}}
\centering
\setlength\tabcolsep{0.5 pt}
\renewcommand{\arraystretch}{0.2}
\scriptsize 	
\begin{tabular}{ccccc}
Original & Text-Sketch, 0.0281 bpp &  VTM, 0.0242 bpp & MS-ILLM, 0.0065 bpp  & \ours (ours), 0.0032 bpp  \\ 
\myim{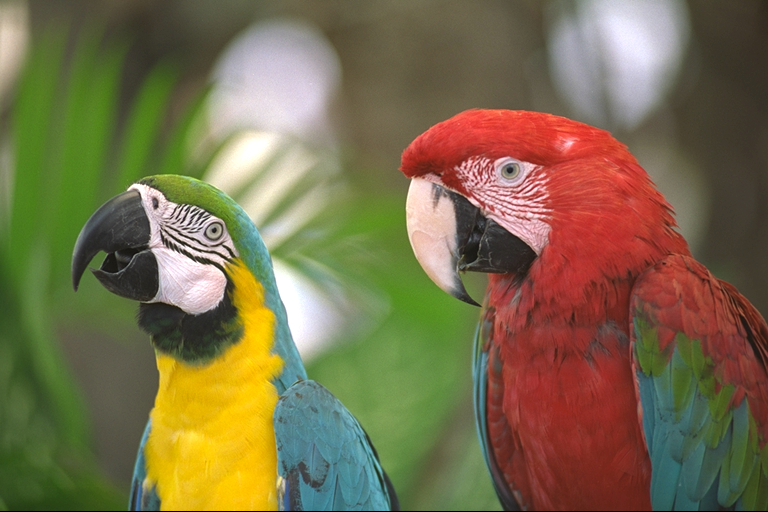} & \myim{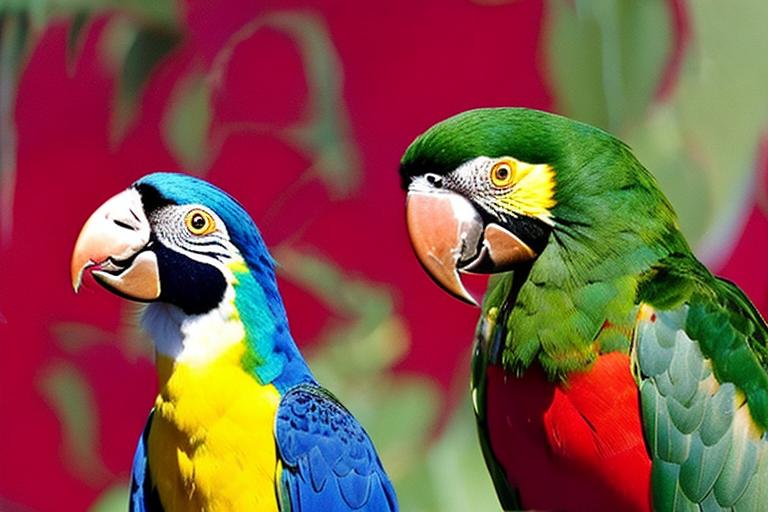}  & \myim{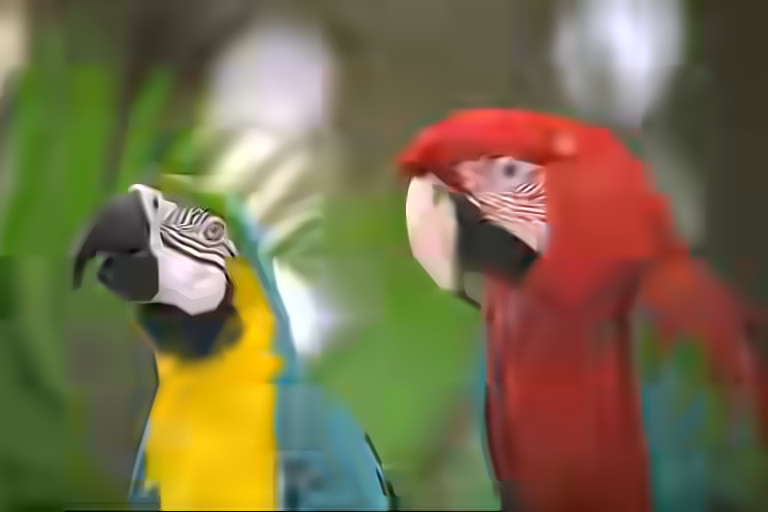}&  \myim{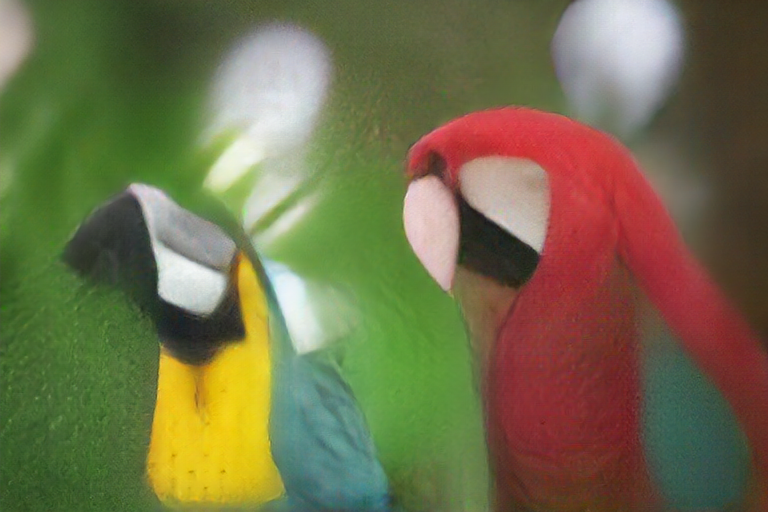} & 
\myim{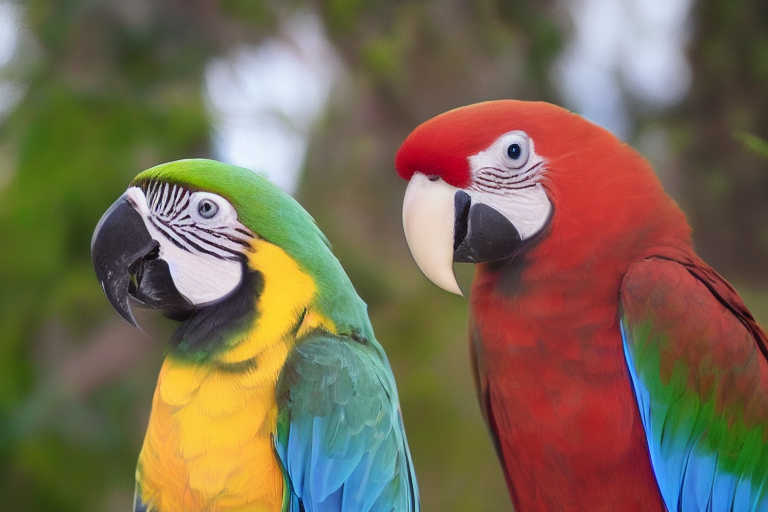}  \\
\myim{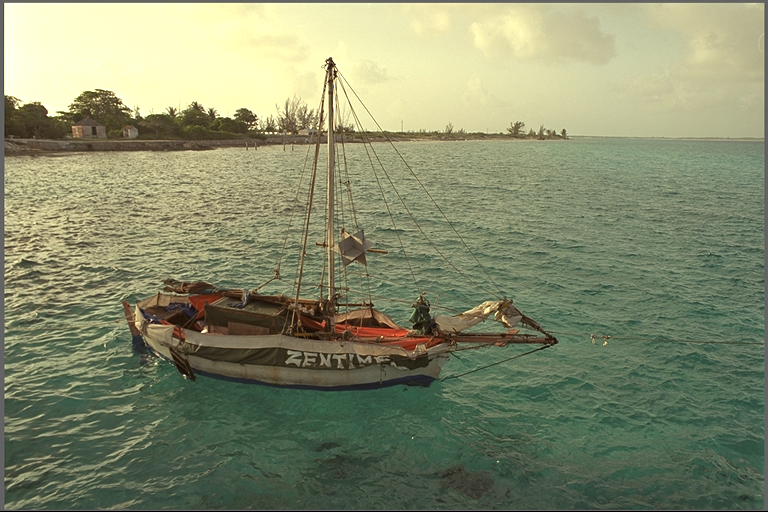} &  \myim{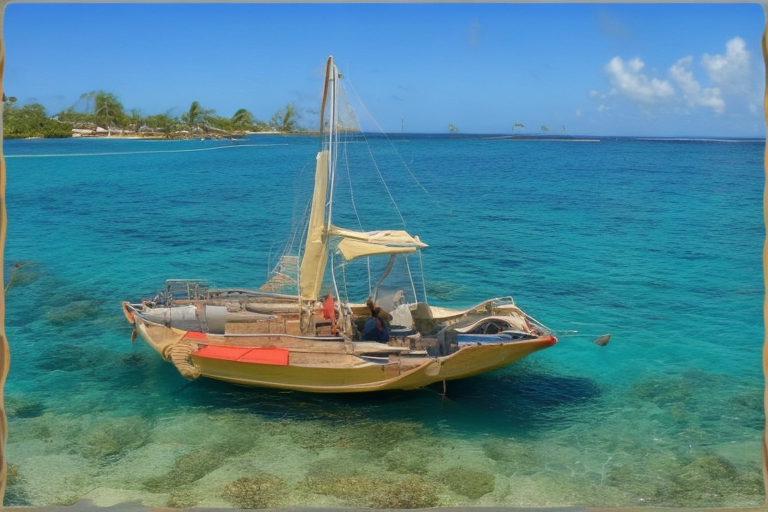}  &    \myim{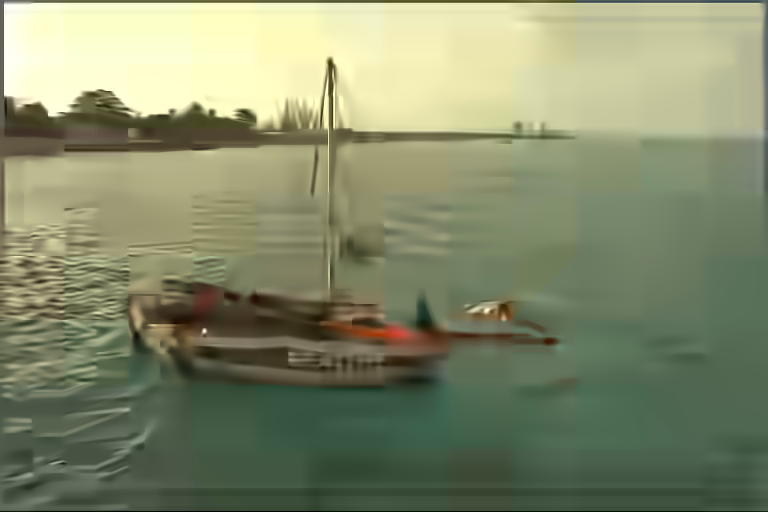}  & \myim{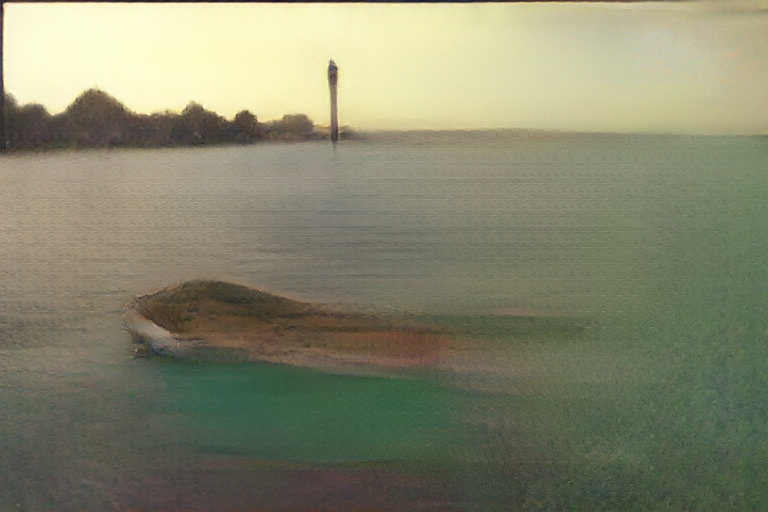} &

\myim{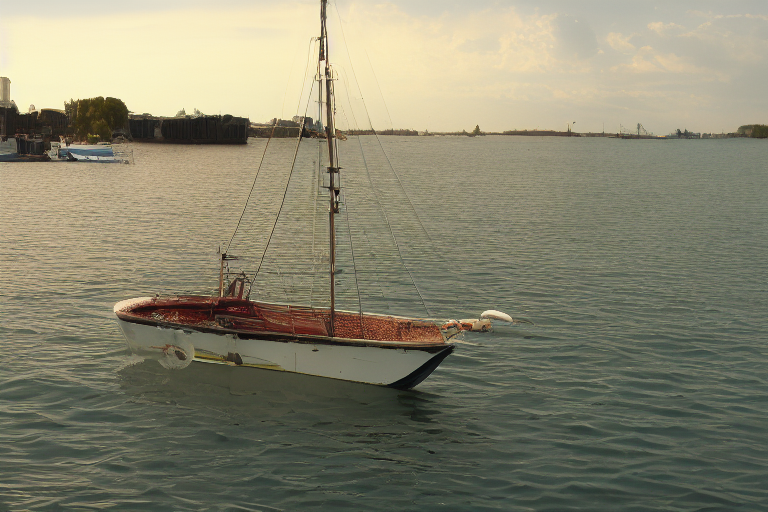}  \\
 \myim{16_gt.png} & \myim{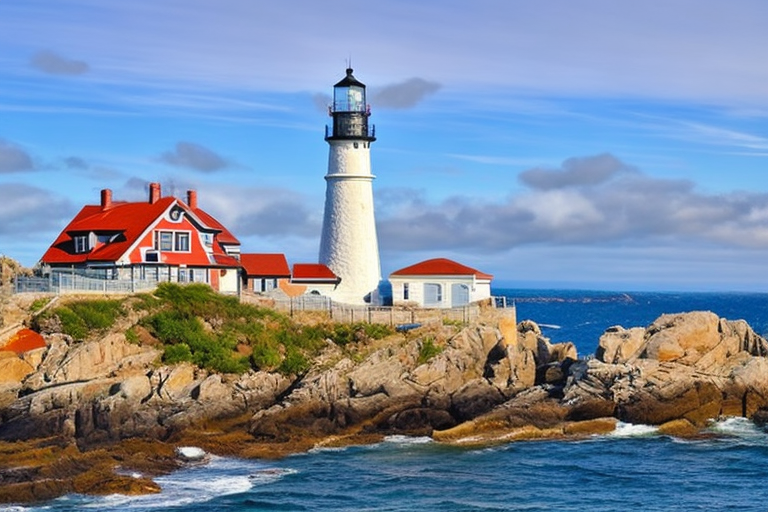}  & \myim{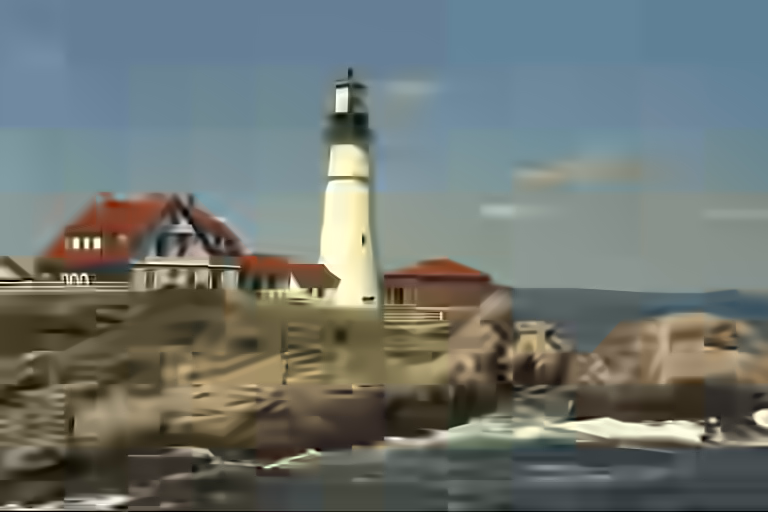}  &\myim{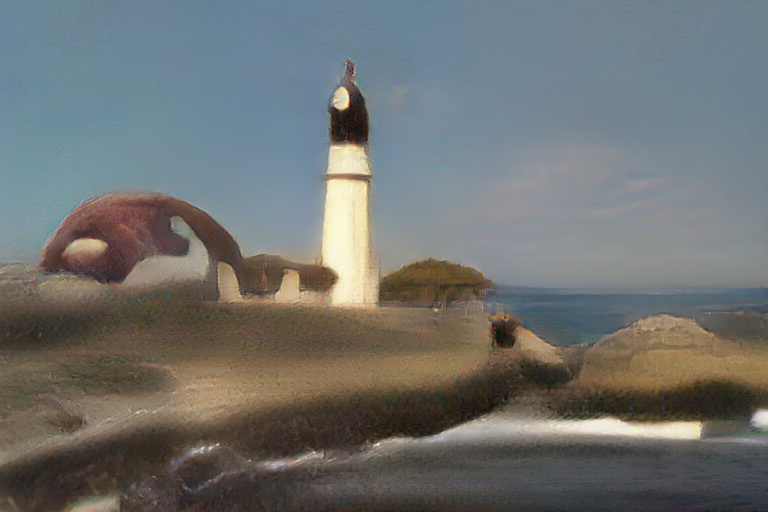} & 
 
 \myim{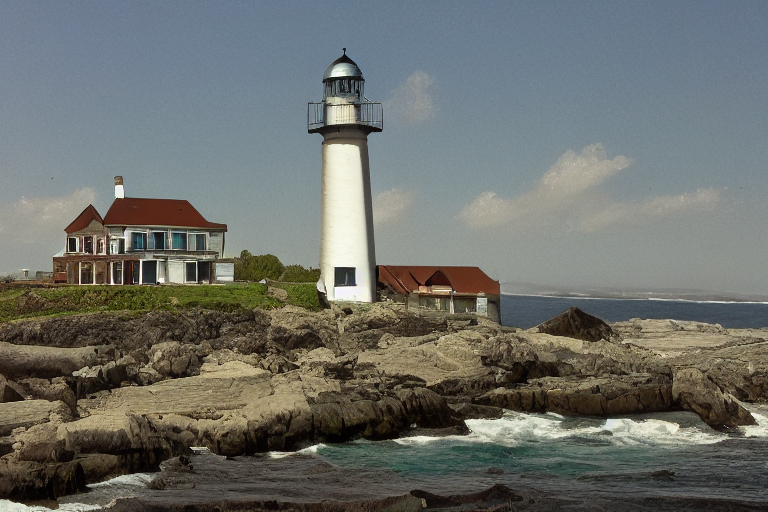}  \\
\end{tabular}

\caption{Kodak images compressed with the diffusion-based Text-Sketch approach PICS~\citep{lei23ncw}, the hand-crafted codec VTM~\citep{vtm}, MS-ILLM~\citep{muckley2023improving} which leverages an adversarial loss, and with \ours (ours). Taking the lowest available bitrate for each method. }

\label{fig:teaser}
\end{figure}

Generative modeling compensates for artefacts such as  blurring by introducing a  divergence term~\citep{agustsson2019generative,mentzer2020high,muckley2023improving}, typically in the form of an adversarial discriminator loss,  which improves human-perceptual performance~\citep{mentzer2020high}.
Such codecs are called ``generative compression'' codecs, and are evaluated in terms of the rate-distortion-realism tradeoff~\citep{blau2019rethinking}.
A further result of rate-distortion-realism theory is the theoretical possibility of a perfect realism codec, \ie, a codec with no $f$-divergence, and zero FID  across all rates, with no more than twofold increase in mean-squared error from the rate-distortion optimal codec. 
This  result motivates research on perfect realism codecs, where original and reconstruction are different from each other, but it is not possible to tell which is which. 
Such codecs are particularly interesting for extremely low bitrate settings, where existing codecs introduce severe  artefacts that are easily perceptible, see \cref{fig:teaser}.
So far, the primary effort towards building such a codec was the work of \citet{theis2022lossy}, who demonstrated lower FID scores than  HiFiC~\citep{mentzer2020high} 
via a combination of a pretrained variational diffusion model and reverse channel coding.
However, computational constraints prohibited application to images larger than $64\!\times\!64$ pixels.

In this paper we make further progress towards  perfect-realism codecs.
Similar to \citet{theis2022lossy}, we  leverage a pretrained text-to-image diffusion model; however, 
rather than using the diffusion model for compression directly, we use it as a decoder in a VQ-VAE-like autoencoder.
The (approximate) log-likelihood loss of diffusion models offers an alternative to the use of distortions such as MSE or LPIPS used to train most neural compression models that make strong assumptions on the conditonal distribution on images given the latents which are inappropriate at  low bitrates, and lead to compression artefacts.
For decoding, we sample from the conditional diffusion model, which allows us to obtain a set of reconstructions that reflect the uncertainty about the original  source image. We augment the quantized image representation with a global image description that captures the high-level semantic information of the image. To this end, we use an automatically generated textual image description, \eg using BLIP~\citep{li23icml}. Alternatively, this can take the form of  a global image feature extracted from a pre-trained image backbone.
We demonstrate that this high-level information is helpful for compression performance at extremely low rates.

Our work is closely related to that of \citet{lei23ncw}, who explored text-conditioned diffusion decoders for image compression, and the only prior work that we are aware of that considers the same ultra-low bitrates as we do.
In contrast, their work uses (automatically generated) binary contour maps as a spatial conditioning signal which carry little detail on  colors and textures, and leads to reconstructions that are not faithful to the original.
Experimentally we observe significantly improved FID and KID scores compared to  competing methods. Moreover, we find that FID and KID are much more stable across bitrates than other methods, which aligns with our goal of image compression with perfect realism.

In sum, our contributions are as follows:
 \vspace{-2\topsep}
\begin{itemize}
  \setlength\itemsep{0em}
    \item We develop a novel diffusion model called \ours for image compression that is conditioned on both a vector-quantized latent image representation and a textual image description.
    \item We obtain realistic reconstructions at bitrates as low as 0.003 bits per pixel, significantly improving over previous work, see \Cref{fig:teaser}.
    \item We obtain state-of-the-art FID and KID performance on the MS-COCO 30k dataset; and observe no significant degradation of FID when reducing the bitrate.
\end{itemize}

\section{Related work}

\mypar{Neural image compression codecs}
\citet{balle2017end} demonstrated that an end-to-end neural image compression codec could outperform the classical JPEG codec in terms of rate-distortion performance.
\citet{balle2018variational} enhance this approach by conditioning the latent on a ``hyperprior'' that encodes high-level information in the image, greatly improving performance.
Follow-up works improved the latent conditioning mechanism~\citep{minnen2018joint,minnen2020channel,cheng2020joint,he2022elic}.
Optimization for rate-distortion alone leads to distributional mismatch between the compressed image distribution and the natural image distribution~\citep{blau2018perception,blau2019rethinking}.
\citet{agustsson2019generative} and \citet{mentzer2020high} added an adversarial loss term and demonstrated its usefulness for human perception. Similarly \citet{he2022po} builds upon ELIC~\citep{he2022elic} by adding perceptual losses including adversarial ones.
Follow-up works have improved the discriminator architecture~\citep{muckley2023improving}, decode-time realism guidance~\citep{agustsson2023multi}, or investigated alternative autoencoders with discrete entropy models~\citep{elnouby2023image}. \cite{xu2023conditional} introduce the notion of conditional perceptual quality, where perceptual quality is dependent on side information. 
Our work is related to this, as the global textual or visual representation on which we condition the decoder can be seen as a form of side information.

\mypar{Diffusion models}
\citet{ho2020denoising} improved the original diffusion probabilistic model~\citep{sohl2015deep} that gave it impressive performance on image generation benchmarks.
\citet{dhariwal2021diffusion} demonstrated superior performance to GANs for image generation with classifier guidance, while \citet{kingma2021variational} improved diffusion model likelihood estimation. 
Conditioning diffusion models on large text-encoder networks enables these models to generate realistic images from natural language prompts~\citep{nichol21arxiv,saharia2022image}, and defining the diffusion process in an autoencoder latent space reduces the computational cost of the diffusion process. %
In our work we build upon large-scale pretrained text-conditioned diffusion models  to benefit from the image generation capabilities embedded in them.

\mypar{Diffusion generative compressors}
\citet{ho2020denoising} considered compression with diffusion models and reverse channel coding and the corresponding rate-distortion performance, but reported only experiments on $32\!\times\!32$ CIFAR-10 images and MSE distortion. 
\citet{theis2022lossy} demonstrated that the reverse channel coding approach gave FID numbers superior to HiFiC for $64\!\times\!64$ ImageNet images.
\citet{saharia2022image} introduced a  super-resolution diffusion model that removed artefacts from JPEG images.
\citet{ghouse2023residual} developed a complete version of a residual diffusion codec with images of high quality for a suite of codecs.
\citet{hoogeboom2023high} showed similar results using a neural autoencoder baseline codec, and using a separate diffusion model that does not require any additional bits to refine the reconstruction.
\citet{yang2022lossy} developed an alternative approach, where a hyperprior-based neural encoder was trained jointly with a diffusion decoder, also showing improved performance to HiFiC.

In \citet{pan2022extreme}, they optimize a textual embedding on top of a pretrained text-to-image diffusion model. They also design a compression guidance method used at each denoising step to better reconstruct images.
\citet{lei23ncw} leverage text-conditioned diffusion models for image compression at very low bitrates, using prompt inversion to encode the image into text, and adding spatial detail via compressed binary contour maps.  Our work is similar, but we replace computationally costly  prompt inversion~\citep{wen23arxiv} with fast  feed-forward image captioning~\citep{li23icml}, and replace binary contour maps which carry little appearance information with  end-to-end learned vector-quantized image features. 
\section{Perceptual compression with a diffusion decoder}

\subsection{Overall framework}

The foundation of most lossy compression  lies in rate-distortion theory~\citep{shannon1948mathematical}, which  quantifies the balance between the bitrate needed to transmit a signal \vs the distortion of the signal.
Let us assume an input signal $\vx$, with its corresponding quantized representation $\vz$ and reconstruction $\hat{\vx}(\vz)$. Neural compression is commonly achieved via the minimization of a training objective formulated as a linear combination of a rate  and a distortion term:
\begin{equation}
\label{eq:rateDistortionori}
\mathcal{L}_\textrm{RD}=\E_{P_\vx}[\E_{P_{\vz|\vx}}\mathcal{L}_\textrm{R}(\vz)+\lambda \mathcal{L}_\textrm{D}(\hat{\vx}(\vz),\vx)],
\end{equation}

where $P_\vx$ is the data distribution and $P_{\vz|\vx}$ is the posterior distribution of the quantized codes.
The rate term $\mathcal{L}_\textrm{R}(\vz)$ 

estimates the bitrate either by means of scalar quantization and a continuous entropy model~\citep{balle2017end}, or using vector-quantization in combination with discrete entropy models~\citep{elnouby2023image}. 
For simplicity, in \ours, we employ vector-quantization combined with a uniform entropy model leading to a constant rate controlled through two hyper-parameters: the number of elements that are quantized, and the codebook size. Therefore, the rate $\mathcal{L}_\textrm{R}$ can be considered fixed when training our models and can be dropped from the optimization formulation. 
Next, we introduce a new formulation of the distortion term, $\mathcal{L}_\textrm{D}$,  enabling the integration of a pre-trained diffusion model, thereby achieving a higher level of perceptual quality. Our aim with this formulation is to leverage a large pre-trained diffusion model as a robust image prior, enabling realistic image reconstruction even at extremely low bitrates.

\begin{figure}
    \centering    \includegraphics[width=\textwidth]{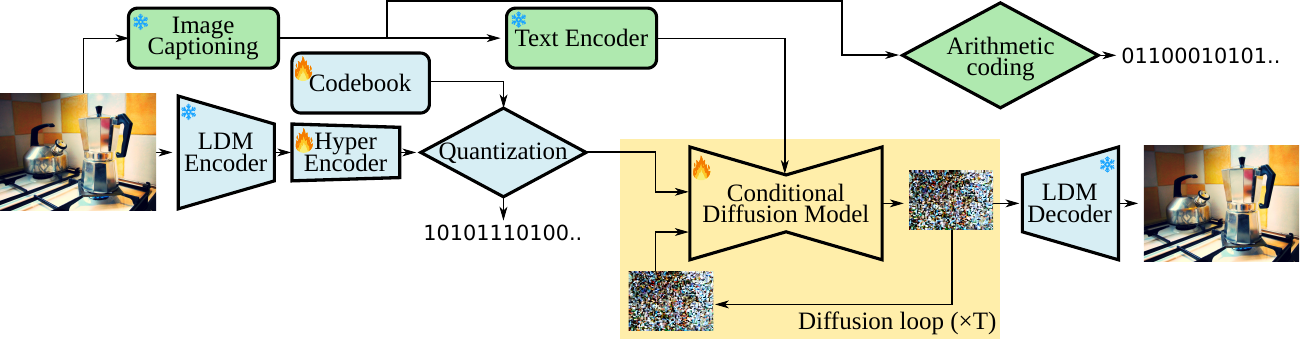}
        \caption{
    Overview of \ours.
    The LDM encoder maps an RGB image into the latent space of the diffusion model. 
    The ``hyper encoder'' then maps the image to a hyper-latent with smaller spatial resolution, which is then vector-quantized and represented as a bitstream using uniform coding.
    The image captioning model generates a textual description of the input image, which is  losslessly compressed, and processed by a text encoder to condition the diffusion model. 
    The  diffusion model  reconstructs the input image in its latent space conditioned on the output of the text encoder and the hyper encoder. 
    Finally, the LDM decoder maps the latent reconstruction back to RGB pixel space. 
   }
    \label{fig:flowchart}
\end{figure}

Diffusion models~\citep{sohl2015deep,ho2020denoising}  approximate a data distribution by gradually denoising a unit Gaussian random variable. This is achieved by training a denoising function to reverse a diffusion process that takes an initial sample, denoted as $\vx_0$, from the data distribution and adds small Gaussian noise over a large number of  steps. 
In practice, a noise estimator  $\epsilon_\theta(\vx_t,t,\vz)$ implemented as a neural network is trained to denoise input data points at timesteps  \mbox{$t \in[0,T]$}, where $\vx_t\!=\!\sqrt{\alpha_t} \vx_0\!+\!\sqrt{1-\alpha_t} \epsilon $ is a corrupted version of the original data point $\vx_0$ with Gaussian noise $\epsilon$. 
The scalar $\alpha_t$  fixes the noise strength, varying from $\alpha_0\!\simeq\!1$ for no noise to $\alpha_T\!\simeq\!0$ for pure Gaussian noise. 
Finally, $\vz$ is an optional condition, such as a text-prompt in text-to-image models.
In our case, $\vz$  is the quantized representation of the input image. 
Once the noise estimator is trained, the model can generate samples by drawing    $\vx_T \sim \mathcal{N}(0,I)$, and applying iterative denoising for $t\!=\!T,\dots,1$.
To reduce the computational cost, we use a latent diffusion model (LDM) where   the variables $\vx_t$ are defined in the latent space of an autoencoder.

We formulate our distortion loss within the probabilistic diffusion framework. 
For every the diffusion step $t$, we compute an estimation $\hat{\vx}_{t-1}$ of $\vx_{t-1}$ from $\vz$ and $\vx_{t}$ and minimize its error:
\begin{equation}
\label{eq:rateDistortionoriDif}
\mathcal{L}_\textrm{Diff}^t=\E_{P_\vx}
\E_{P_{\vz,\vx_t|\vx}}\rVert\vx_{t-1}-\hat{\vx}_{t-1}(\vx_{t},\vz)\big\rVert_{2}^{2}.
\end{equation}
Up to a   multiplicative constant, this loss can be rewritten as 
\begin{equation}
\label{eq:loss_dist}
 \mathcal{L}_\textrm{Diff}^t \propto   \E_{P_\vx} 
\E_{P_{\vz,\vx_t|\vx}}\mathop{\mathbb{E}}_{\epsilon\sim \mathcal{N}(0,1)}  \lVert  \bm{\epsilon} -  \bm{\epsilon}_\theta(\vx_t,\vz,t) \rVert_{2}^{2}.
\end{equation}
We  obtain the final  loss by  taking the expectation of \cref{eq:loss_dist} \wrt the steps $t$, and   use  $\vv$ prediction objective of  \cite{salimans22iclr}, which is more stable during training. 

The  loss   can be augmented with any image-level loss such as LPIPS~\citep{zhang2018unreasonable} to improve perceptual image similarity.

\subsection{Encoding local and global context}

We encode the input image as $\vz=(\vz_l, \vz_g)$, where  $\vz_l$ and $\vz_g$ carry complementary local and global context, respectively. 
The overall structure of our compression scheme is illustrated in \Cref{fig:flowchart}.
 
\mypar{Local spatial encoding}
We encode the input image into a quantized tensor  proceeding in three stages. First, we employ the LDM encoder $E(\cdot)$  to achieve a first dimension reduction. Assuming an input image of resolution $512\!\times\!512$, we obtain a feature map of dimension $4\!\times\!64\!\times\!64$. 
Second, we add a lightweight ``hyper encoder'' $H$ composed of several convolutional layers to project the LDM features to a lower spatial resolution tensor $\mH_s$. 
The spatial resolution, $h\!\times\!w$, of $\mH_s$ is chosen according to the target bitrate. Third, we proceed to quantization of $\mH_s$ to obtain $\vz_l$ via vector-quantization. As in VQ-VAE~\citep{oord2017neural,razavi19nips}, each of the $h\!\times\!w$ vectors $\vh_s$ in $\mH_s$  is mapped to an element in a codebook learned using the vector quantization loss:
\begin{equation} 
 \mathcal{L}_\textrm{VQ} = \E_{\vh_s} \left[ \lVert sg(\vh_s)-\vz_q \rVert_{2}^{2} + \lVert sg(\vz_q)-\vh_s \rVert_{2}^{2} \right],
 \label{eq:vqloss}
\end{equation}
where $sg(.)$ is the stop-gradient operation, and $\vz_{q}$ is the mapping of $\vh_s$ to its closest codebook entry. 
For better codebook usage,  we follow  \citet{yu22iclr} and set the output of the hyper encoder to be relatively  low dimensional: 32  in our case.
We also found it beneficial to replace the codebook loss by exponential moving average on the codebooks and thus the first term in \cref{eq:vqloss} is no longer used.
The quantized encoder output is then upsampled to $64\!\times\!64$ resolution, to be channel-wise concatenated to the input of the noise estimation network of the diffusion model. The hyper-network is trained end-to-end while finetuning the diffusion model.
For simplicity, we opt for    uniform coding  to map the quantized latents to a bitstream.
Using $\log_2 V$ bits to encode each element in the $h\!\times\!w$ sized hyper latent encoding, where $V$ is the codebook size,   results in a total of $hw\log_2V$ bits.

\mypar{Global encoding with image captioning}
While $\vz_l$ can accurately encode local information, our experiments show that relying on $\vz_l$ alone leads to unsatisfying realism. 
Therefore, we add  a global encoding of the image $\vz_g$ which provides additional context. For this purpose, we use an off-the-shelf state-of-the-art image captioning model, such as BLIP-2~\citep{li23icml} or IDEFICS~\citep{idefics}, which we keep fixed when training our model. Note that user-generated captions can also be used, as explored in our experiments. 
Similar to~\citet{lei23ncw}, we  losslessly compress the caption using  Lempel-Ziv coding as implemented in the zlib library~\citep{zlib} to obtain $\vz_g$. Alternatively, we also explore the use of global image features extracted using an image backbone network.

\mypar{Decoding with a diffusion model}
The quantized local image features in $\vz_l$ are  fed to the denoising U-Net of the diffusion model via concatenation with the latent features $\vx_t$ of the current time step. To adapt to this change of dimension, we extend the kernel of the first convolutional layer of the U-Net by adding a number of channels that corresponds to the channel dimension of $\vz_l$ which is randomly initialized. The global encoding $\vz_g$ is losslessly decoded and passed to the diffusion model through cross-attention layers of the pre-trained diffusion model. Furthermore, we found it beneficial to use classifier-free guidance~\citep{ho21cfg}, which we apply at inference time for the  text conditioning $\vz_g$, 
\ie we contrast conditioning on local featues $\vz_l$  alone \vs conditioning on both $\vz_l$  and $\vz_g$.
This leads to the noise estimate:
\begin{equation} 
  \hat{\epsilon}_\theta = \epsilon_\theta(\vx_t,(\vz_l,\emptyset),t) + \lambda_s~\big( \epsilon_\theta(\vx_t,(\vz_l, \vz_g),t) - \epsilon_\theta(\vx_t,(\vz_l,\emptyset),t)\big),
\end{equation}
where $\lambda_s$ is the guidance scale. We empirically found $\lambda_s=3$ to work well. To enable classifier-free guidance, we train \ours by dropping the text-conditioning in 10$\%$ of the training iterations. 
When dropping the text-conditioning we use a constant learned text-embedding instead.
\section{Experiments}

\subsection{Experimental setup}

\mypar{Implementation details}
We base our model off a  text-conditioned latent diffusion model, with latent space provided by a  convolutional autoencoder and diffusion model with an architecture similar to GLIDE~\citep{nichol21arxiv}. 
The model is pre-trained on a proprietary dataset consisting of around 300M image-caption pairs.
We also experiment with an image-conditioned model.
We use BLIP-2~\citep{li23icml} to obtain image captions for conditioning the decoder, and limit the  maximum caption length to 32 tokens. 
The hyper encoder consists of nine residual convolutional blocks, and includes a number of  downsampling layers depending on the spatial map resolution selected for the target bitrate. 
In total, the hyper encoder contains between 4M and 8M parameters. During all our experiments, the autoencoder weights are frozen.
We train the hyper encoder and finetune  the diffusion model on OpenImages~\citep{kuznetsova2020open}, similar to \citet{lei23ncw} and  \citet{muckley2023improving}.
We use random $512\!\times\!512$ crops, and instead of finetuning the full U-Net we found it beneficial to only finetune  the linear layers present of the diffusion model, representing around 15\% of all weights.

\mypar{Datasets} 
For evaluation, we use the \textbf{Kodak} dataset~\citep{kodak} as well as \textbf{MS-COCO 30k}. On COCO we evaluate at resolution $256\!\times\!256$ by selecting the same images from the 2014 validation set~\citep{lin14eccv} as \cite{hoogeboom2023high} and \cite{agustsson2023multi}. We evaluate at resolution $512\!\times\!512$ on the 2017 training set~\citep{caesar2018coco}, which is the same resolution used for evaluation  by \cite{lei23ncw}, and use captions and label maps for some metrics.

\textbf{Metrics.} 
To quantify image quality we use  \textbf{FID}~\citep{heusel2017gans} and \textbf{KID}~\citep{binkowski2018demystifying}, which  match feature distributions between sets of original images and their reconstructions. We measure distortion using \textbf{MS-SSIM}~\citep{wang2003multiscale} and \textbf{LPIPS}~\citep{zhang2018unreasonable}.
In \cref{app:quant} we also report \textbf{PSNR}, but find these low-level distortion metrics to be less meaningful for low rates, see 
\cref{fig:psnr2}.
Therefore we include other more semantic-level metrics. We compute the \textbf{CLIP score} to measure  global alignment of reconstructed samples with ground truth captions~\citep{hessel21emnlp}. 
Unless otherwise specified,  we condition our model on  captions generated by BLIP2, which are different from the ground-truth captions used to compute the CLIP score. 
Second, we adopt the mean Intersection over Union (\textbf{mIoU}) metric to measure the local semantic preservation, see \eg \cite{sushko2022oasis}. 
 In this case, we pass reconstructed images through a pre-trained semantic segmentation model,  and  compare the predicted semantic maps to the ground truth ones.
We take the pretrained ViT-Adapter segmentation network~\citep{chen2022vitadapter}, 
which was trained on the COCO train split.
For Kodak we only report LPIPS and MS-SSIM, as the dataset is too small (24 images) to reliably compute FID and KID, and does not come with semantic annotations.

\textbf{Baselines.} 
We include representative codecs from several families.
VTM is a state-of-the-art hand-crafted codec used in the VVC video codec~\citep{vtm}.
Among neural compressors using adversarial losses, we include  MS-ILLM~\citep{muckley2023improving} which improves over HiFiC~\citep{mentzer2020high} using an  adversarial loss based on multi-class classification.
 For HiFiC we include their model with the lowest available bitrate, and add MS-PatchGAN, which is a close reproduction of HiFiC by \cite{muckley2023improving}. 
We compare to multi-realism approach of \cite{agustsson2023multi}, at their lowest bitrate with the best realism.
Text-Sketch~\citep{lei23ncw} is the only prior work we are aware of that evaluates image compression at bitrates below 0.01, it is based on text-conditioned diffusion models.
We  include two other recent diffusion-based approaches: DIRAC~\citep{ghouse2023residual} and HFD/DDPM~\citep{hoogeboom2023high}. 

\begin{figure}
\def\myim#1{ \includegraphics[width=0.245\textwidth]{figs/plot_metrics/#1}}
\def\myimLeg#1{ \includegraphics[width=0.5\textwidth]{figs/plot_metrics/#1}}
\centering
\setlength\tabcolsep{0.5 pt}
\renewcommand{\arraystretch}{0.2}
\scriptsize
\begin{tabular}{cccc}
\multicolumn{4}{c}{\myimLeg{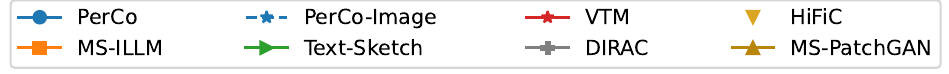}}\\
Kodak& MS-COCO 30k  & MS-COCO 30k & MS-COCO 30k\\
\myim{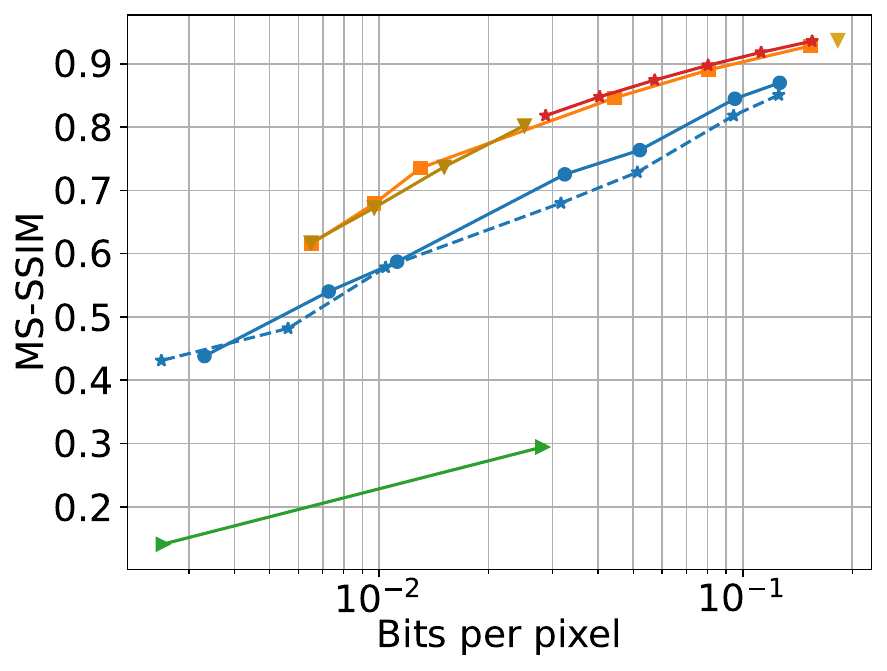} & \myim{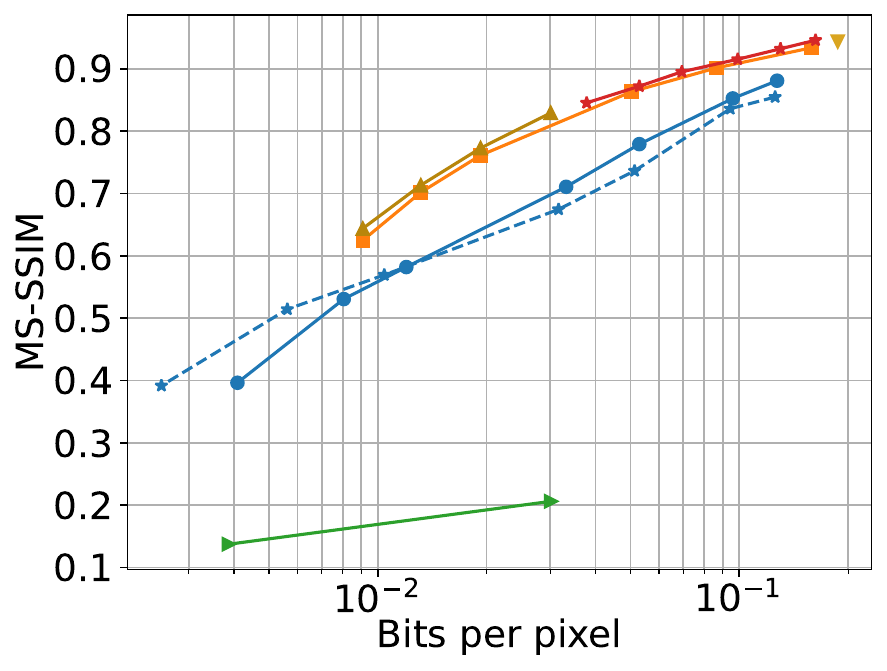} & \myim{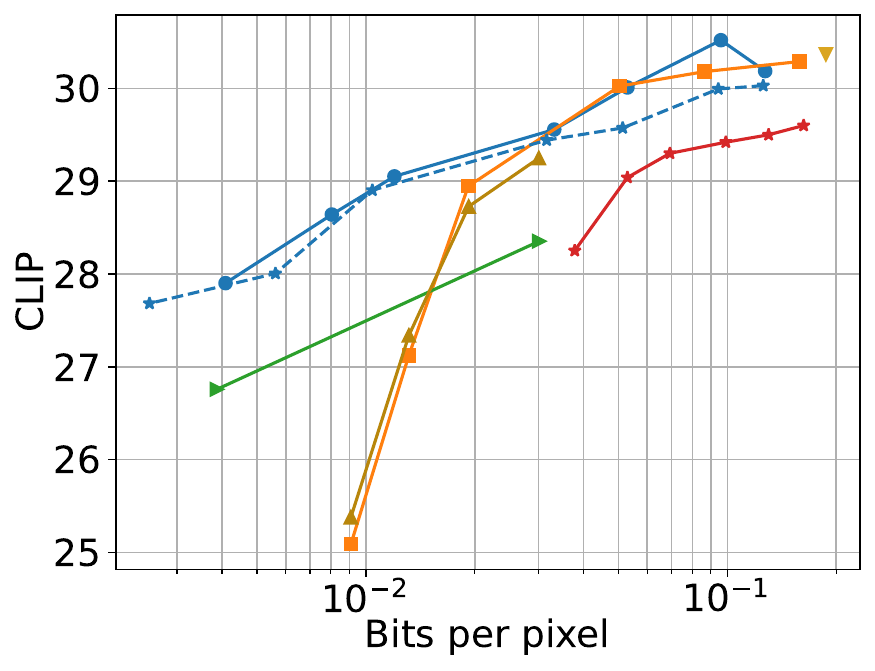} &  \myim{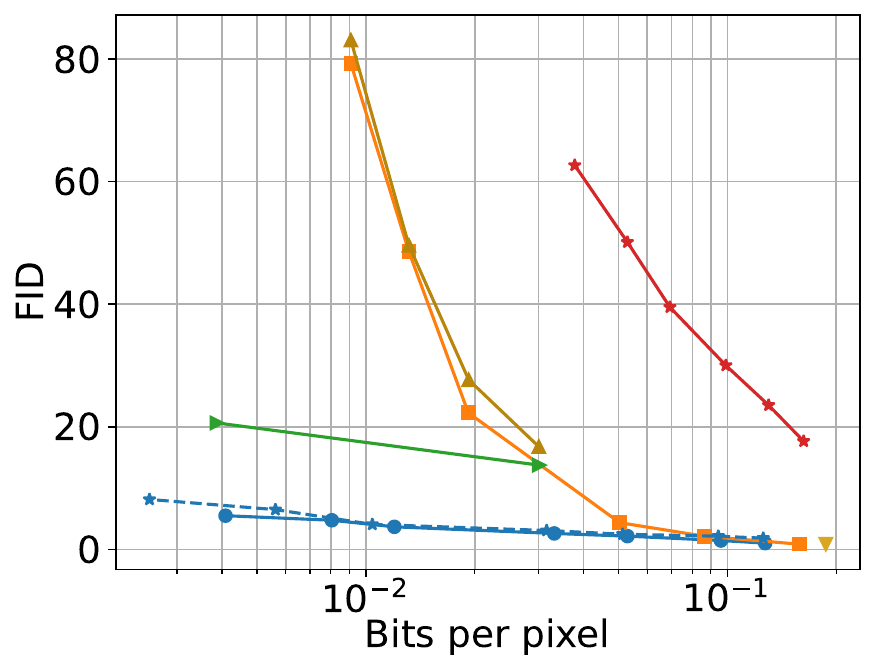}   \\
\myim{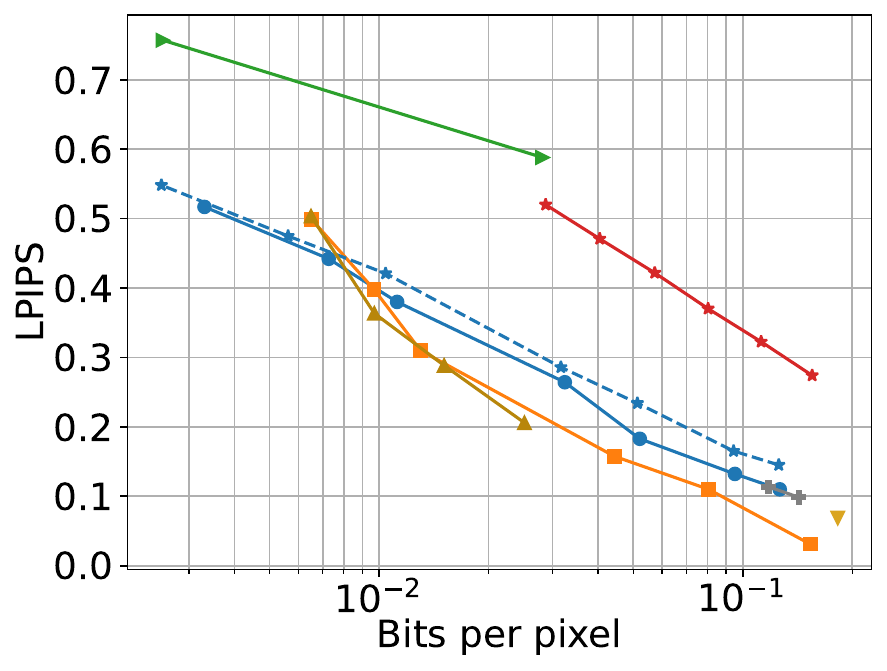} & \myim{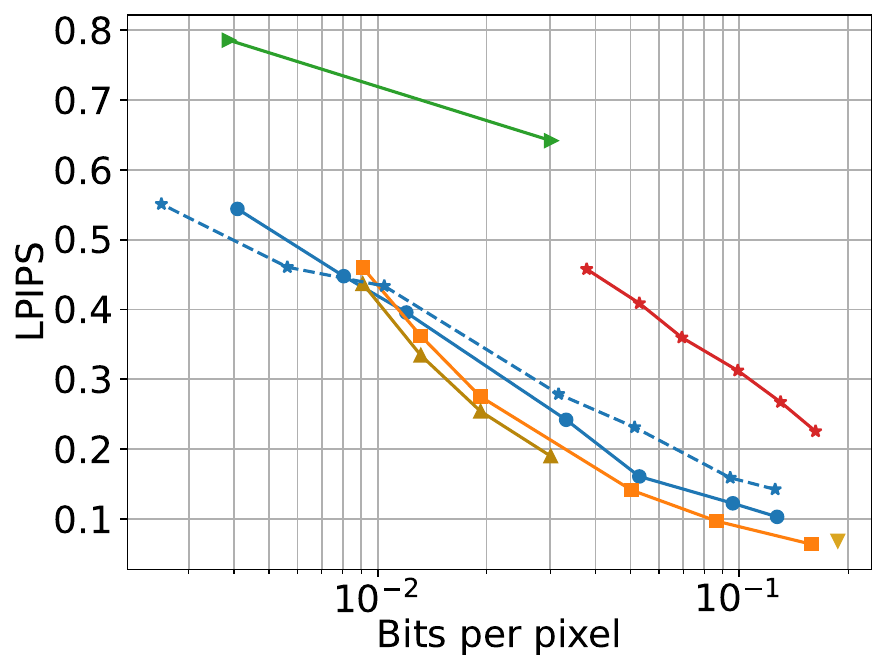} & \myim{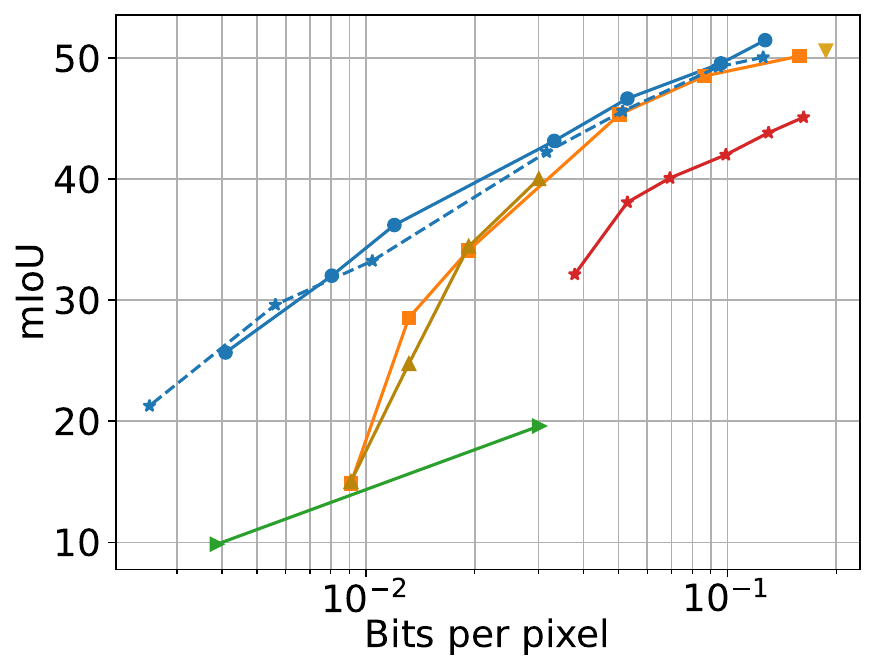} & \myim{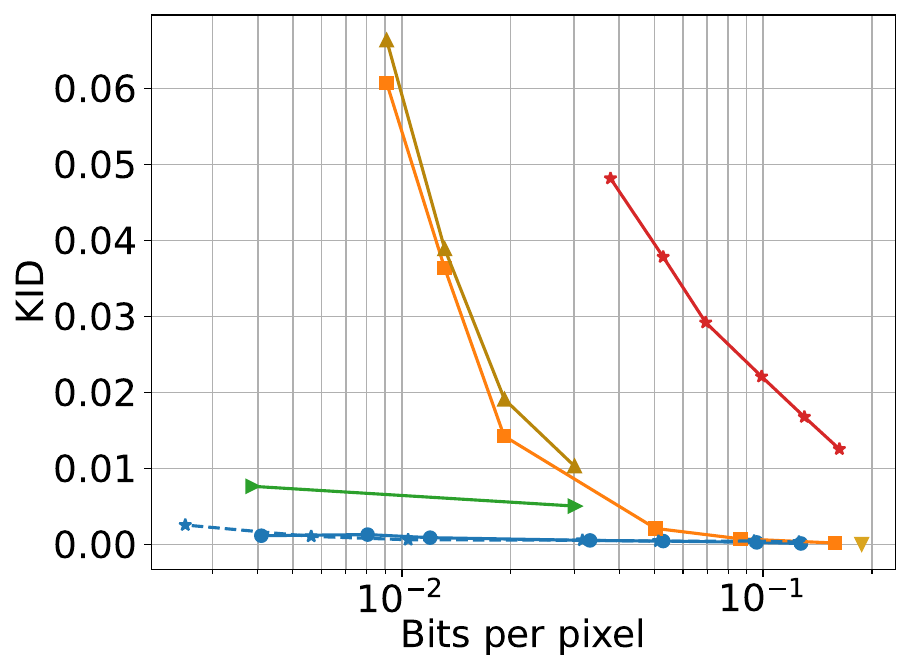}    \\
\end{tabular}
\caption{Evaluation  of \ours and other image compression codecs on Kodak and MS-COCO 30k.}
\label{fig:main_result}
\end{figure}

\subsection{Main results}

\begin{wrapfigure}{r}{0.3\textwidth}
\vspace{-10mm}
\includegraphics[width=0.3\textwidth]{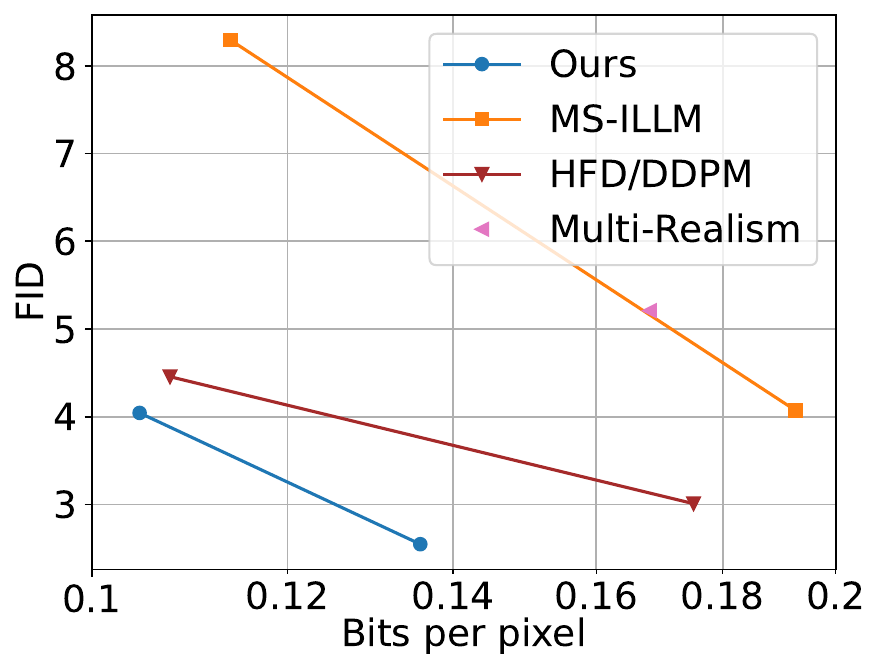}
\caption{Comparisons on COCO at $256\!\times\!256$ resolution. }
\label{fig:HFDComparison}
\end{wrapfigure}

\textbf{Comparison to state-of-the-art methods.} 
In \cref{fig:main_result} we compare our results to  state-of-the-art codecs.
We observe that \ours (ours) yields  significantly lower (better) FID and KID  compared to other approaches at lower bitrates ($<$0.04 bpp), and that our FID and KID curves are much flatter, indicating a decoupling of realism and bitrate that is not seen for other methods.
For the semantics-related CLIP and mIoU metrics we also obtain consistent improvements over all other methods, in particular at low rates.
For LPIPS, \ours is better than Text-Sketch and VTM, but somewhat worse than other approaches for rates $>$0.01 bpp.
Similarly, for MS-SSIM we improve over Text-Sketch, but are worse than other methods. 
 However, as we illustrate in \cref{fig:psnr,fig:psnr2} in \cref{app:qualitative}, such  similarity metrics  are not necessarily meaningful at low bitrates,  as they do not capture  realism. In \cref{fig:HFDComparison} we compare to HFD/DDPM (using their   lowest bitrates: 0.107 and 0.175 bpp) on FID on MS-COCO 30k at $256\!\times\!256$ resolution, as used in their paper.
We also include MS-ILLM and the multi-realism approach of \citet{agustsson2023multi}.
We  find that \ours yields best FID for both rates.

\begin{figure}[t]
\def\myimGT#1{\includegraphics[width=0.24\textwidth,height=0.24\textwidth]{figs/main_fig/kodak/#1}}
\def\myim#1{\includegraphics[width=0.24\textwidth,height=0.24\textwidth]{figs/fig_zoom/#1}}
\def\myimZ#1{\includegraphics[width=0.12\textwidth,height=0.12\textwidth]{figs/fig_zoom/#1}}

\begin{center}
   \setlength\tabcolsep{1 pt}
   \renewcommand{\arraystretch}{0.2}

\scriptsize
\begin{tabular}{cccc}
Original &  MS-ILLM, 0.0065 bpp  & Text-Sketch, PIC, 0.0025 bpp & \ours (ours), 0.0032 bpp  \\ 
\myimGT{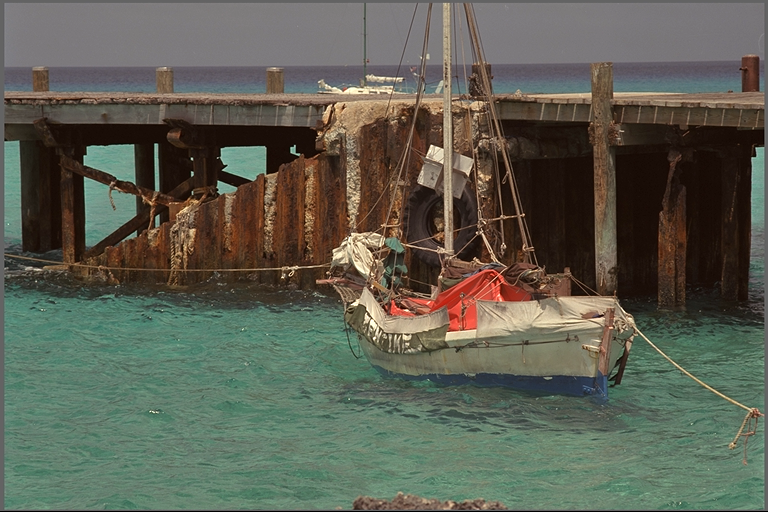} &  \myim{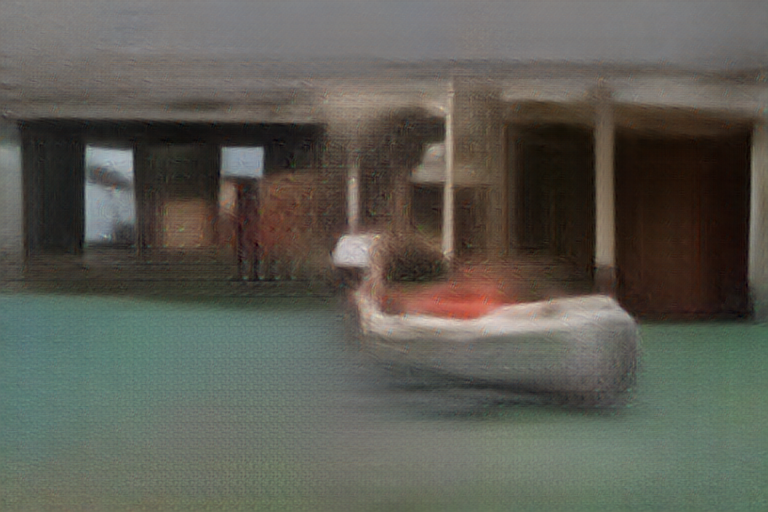} & \myim{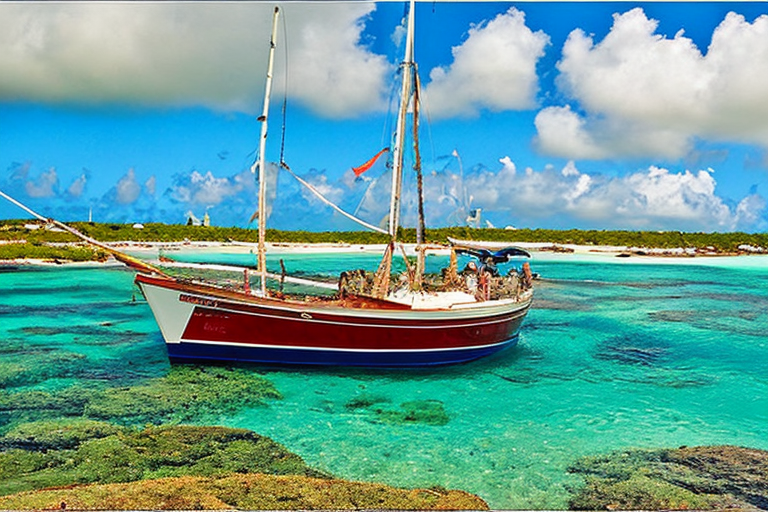}  &
\myim{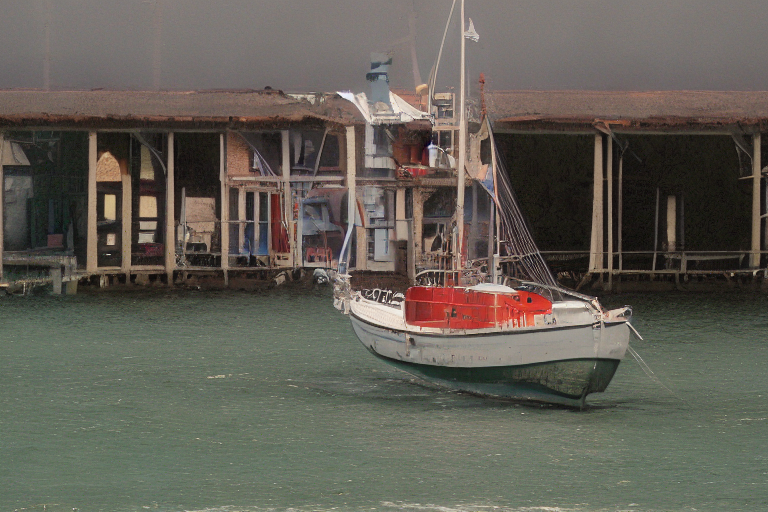}  \\
\myimZ{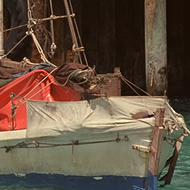}\myimZ{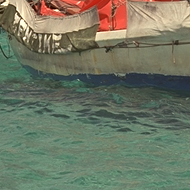} &
\myimZ{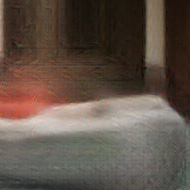}\myimZ{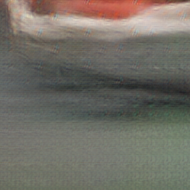} &
\myimZ{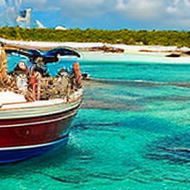}\myimZ{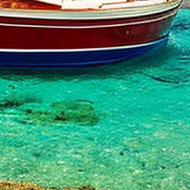} &
\myimZ{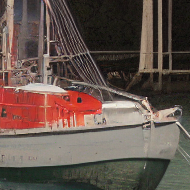}\myimZ{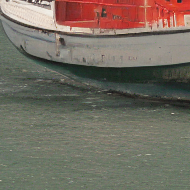} \\
Original &  MS-ILLM, 0.013 bpp  & Text-Sketch, PICS, 0.028 bpp & \ours (ours), 0.011 bpp  \\ 
\myim{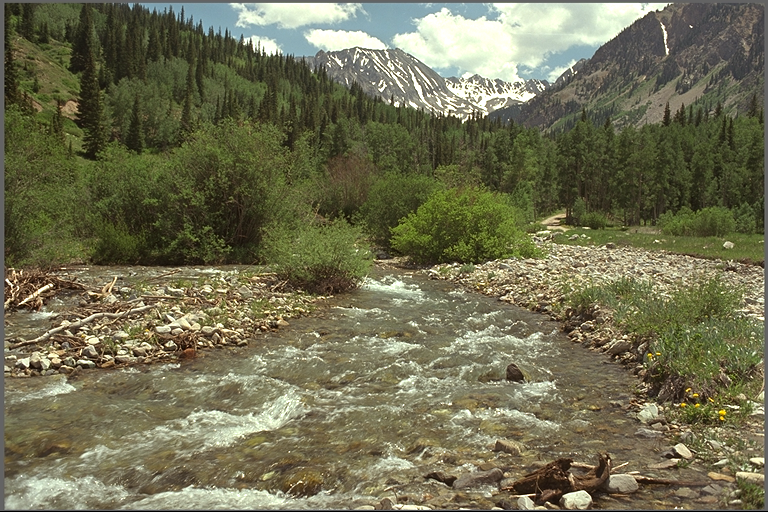} &  \myim{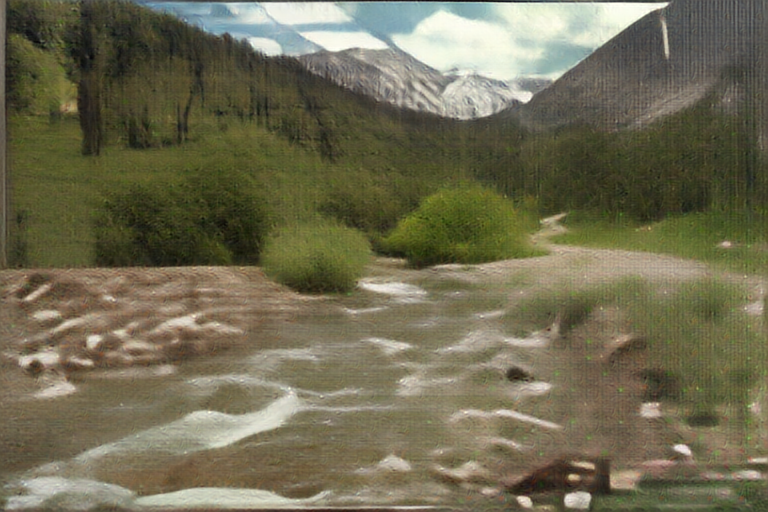} & \myim{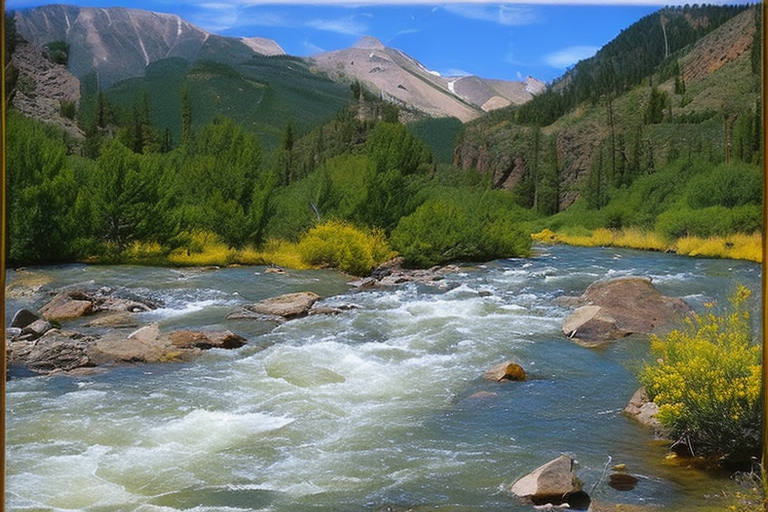}  &
\myim{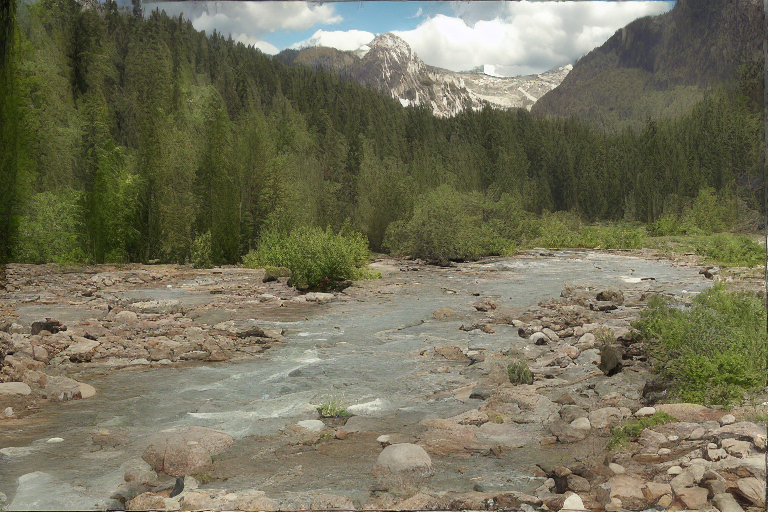}  \\
\myimZ{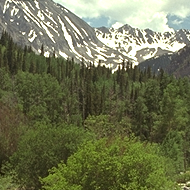}\myimZ{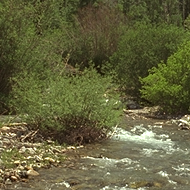} &
\myimZ{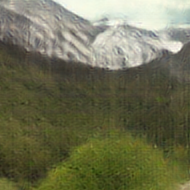}\myimZ{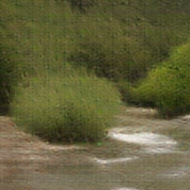} &
\myimZ{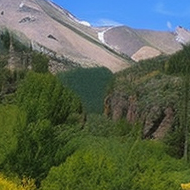}\myimZ{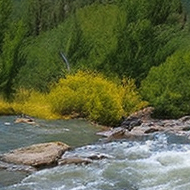} &
\myimZ{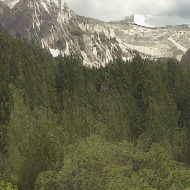}\myimZ{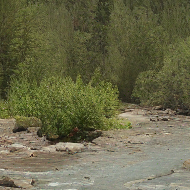} \\
\end{tabular}
\end{center}     
\caption{Comparing \ours on images from the Kodak dataset to MS-ILLM~\citep{muckley2023improving} which leverages an adversarial loss, and the diffusion-based Text-sketch approach~\citep{lei23ncw} conditioned on a text only (top, PIC, 0.0025 bpp) and  text + sketch (bottom, PICS, 0.028 bpp).
}

\label{fig:zoom}
\end{figure}

\begin{figure}[h]
\def\myim#1{ \includegraphics[width=0.16\textwidth,height=0.16\textwidth]{figs/diversity/bpp0p003/#1}}
\def\myimB#1{ \includegraphics[width=0.16\textwidth,height=0.16\textwidth]{figs/diversity/bpp0p01/#1}}
\centering
\setlength\tabcolsep{0.5 pt}
\renewcommand{\arraystretch}{0.2}
\small
\begin{tabular}{ccccccc}
\includegraphics[width=0.16\textwidth,height=0.16\textwidth]{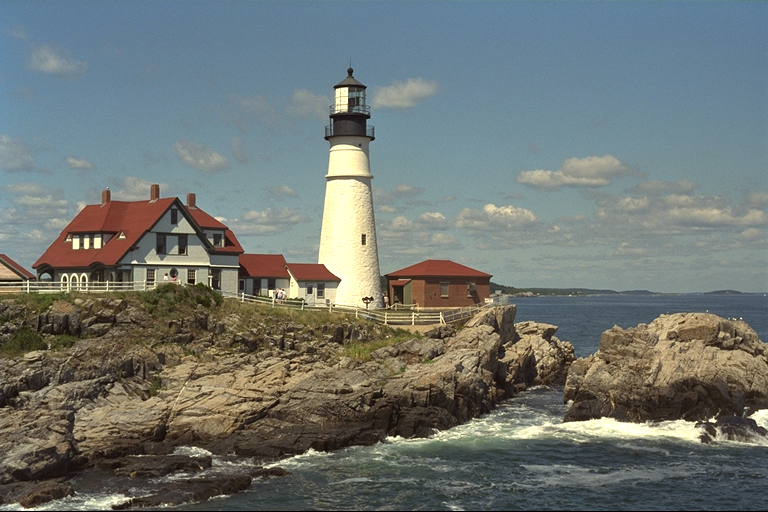} & \myim{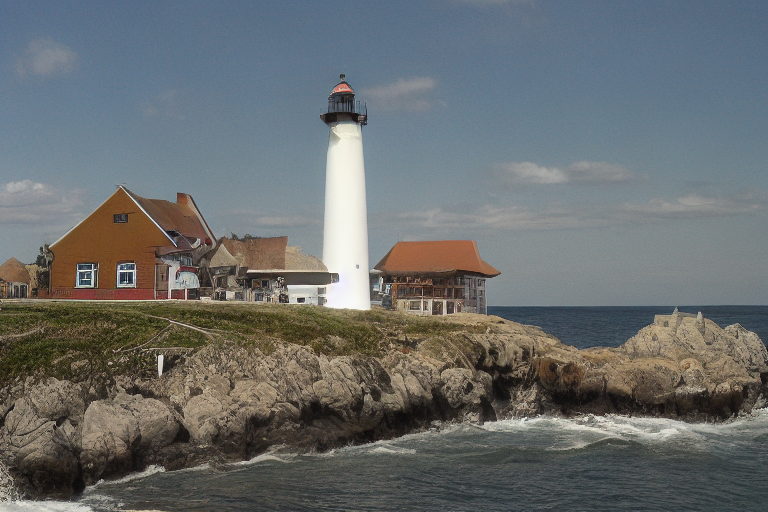} & \myim{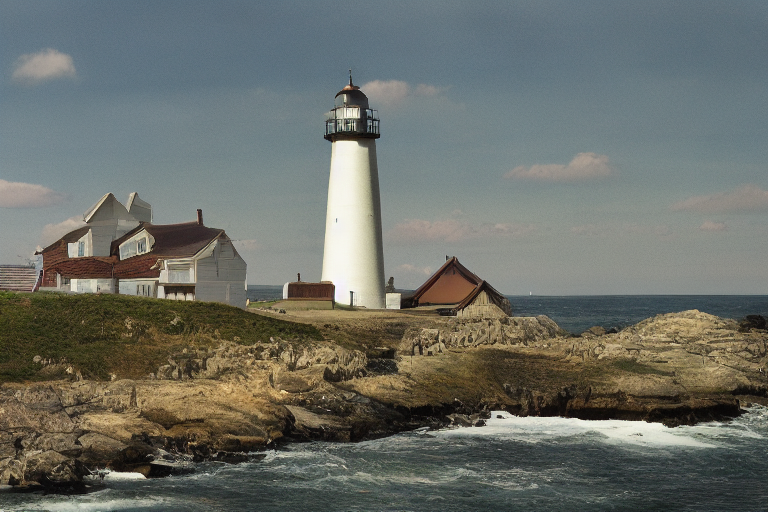} &  \myim{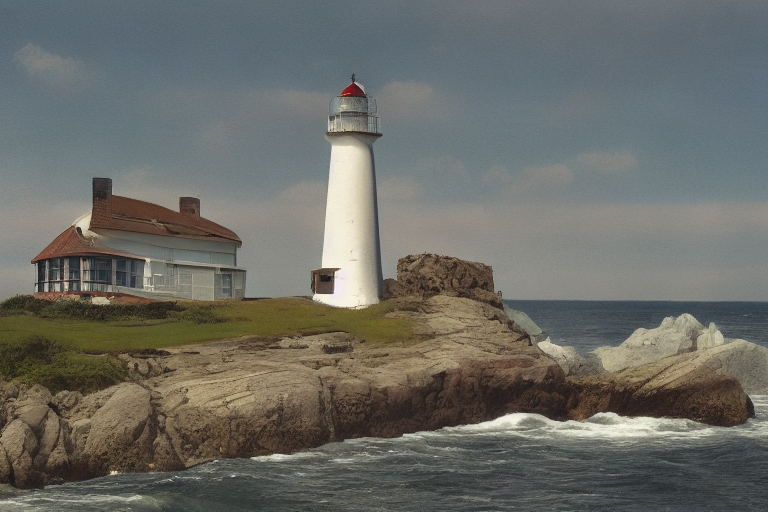} & \myim{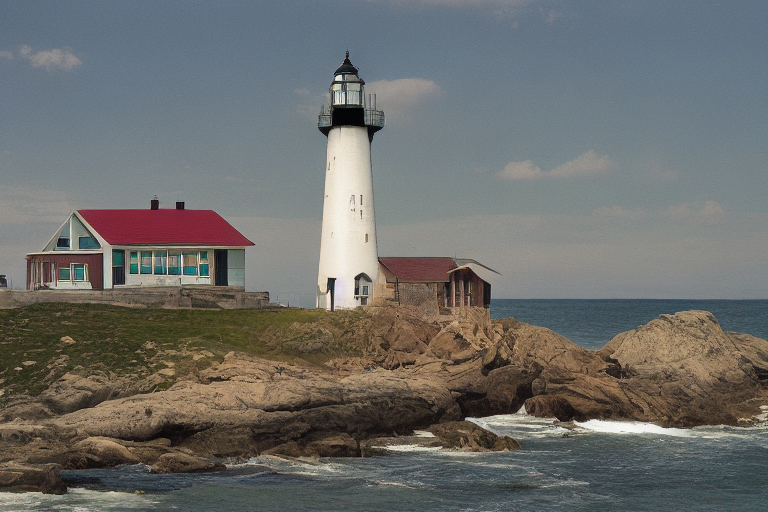}  &
\myim{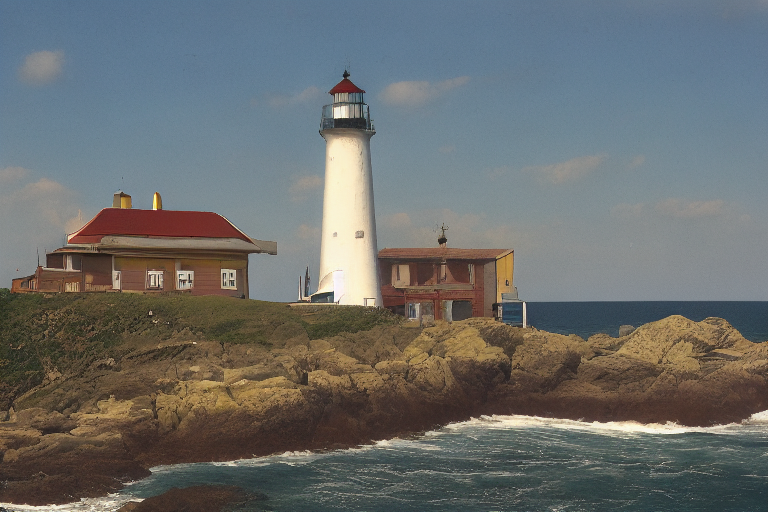}  & \rotatebox{90}{\quad 0.003 bpp}\\
& \myimB{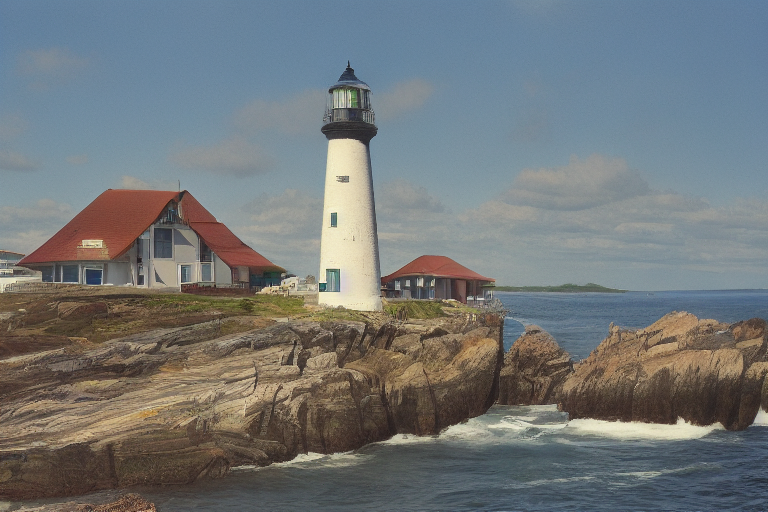} & \myimB{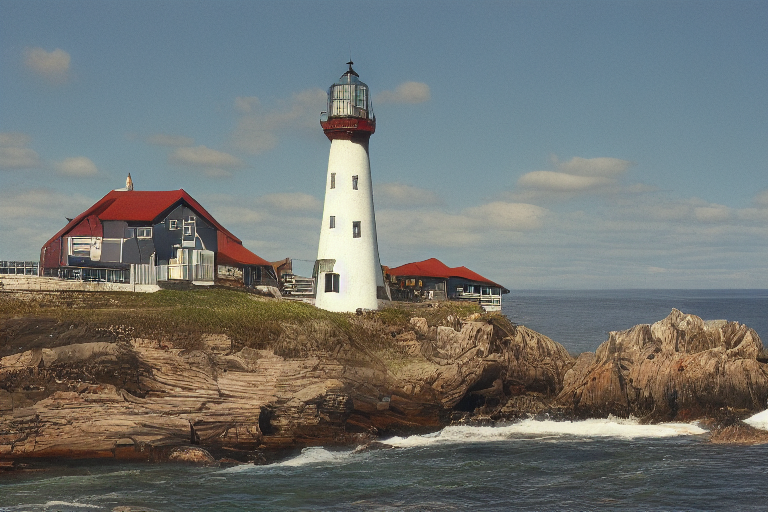} &  \myimB{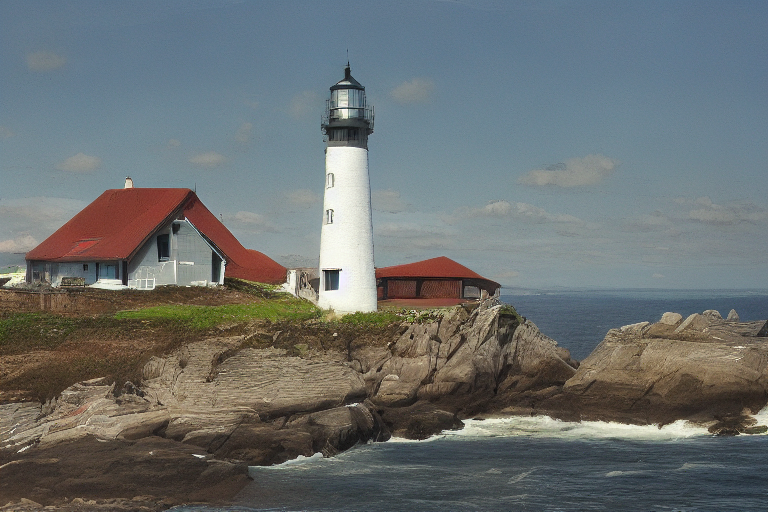} & \myimB{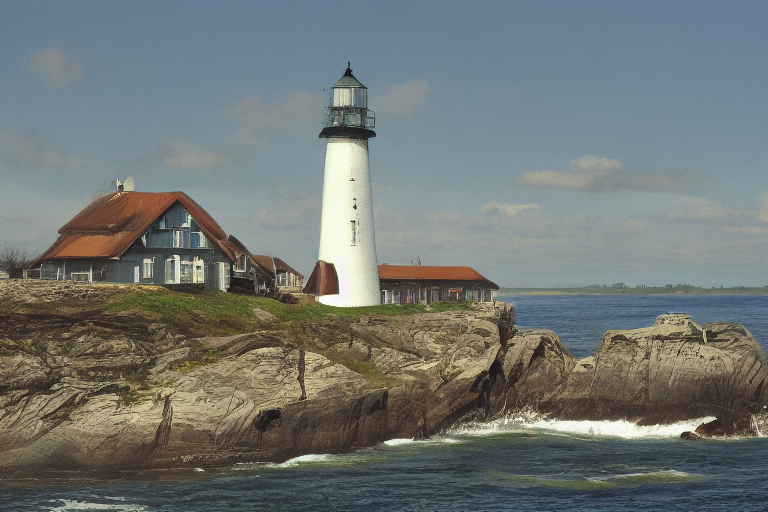}  &
\myimB{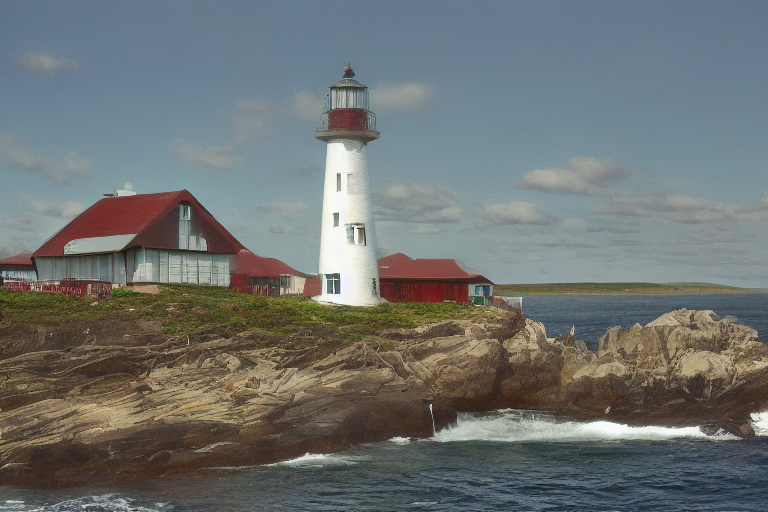} & \rotatebox{90}{\quad 0.011 bpp} \\
\end{tabular}
\caption{Reconstructions  of a Kodak image at 0.003  and 0.011 bpp. 
Best viewed  zoomed in.
}
\label{fig:diversity}
\end{figure}

\textbf{Qualitative comparisons.} We provide qualitative comparisons of \ours to baselines    in \cref{fig:teaser} and  \cref{fig:zoom}. \ours yields reconstructions  of higher quality  with less artefacts, more faithful colors, and realistic details.
VTM and MS-ILLM reconstructions are blurry and/or blocky, while Text-Sketch  strongly deviates from the originals in color. When using text-only conditioning for Text-Sketch (PIC, 0.0025 bpp), it is not able to achieve spatial alignment, by lack of local image encoding.

\subsection{Ablations}

\mypar{Diversity in reconstructions}
With our model, reconstructions are sampled from a conditional diffusion model.
Ideally, the conditional distribution concentrates all mass perfectly on the  original input image. In practice, this does not happen due to limited model capacity, training data, and bitrate. 
In particular, we expect that at lower bitrates ---where there is less information about the original sample--- the conditional distribution would have more variance, and we expect to observe more diversity among the samples.
The reconstruction at two bitrates  in \cref{fig:diversity} confirm this.

\begin{figure}
\def\myim#1{\includegraphics[width=0.25\textwidth]{figs/ablation_text/#1}}
\def\myimLeg#1{\includegraphics[width=.99\textwidth]{figs/ablation_text/#1}}
\centering
\myimLeg{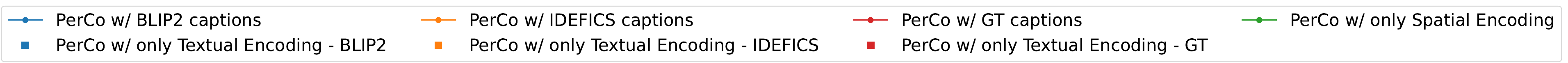} \\
\myim{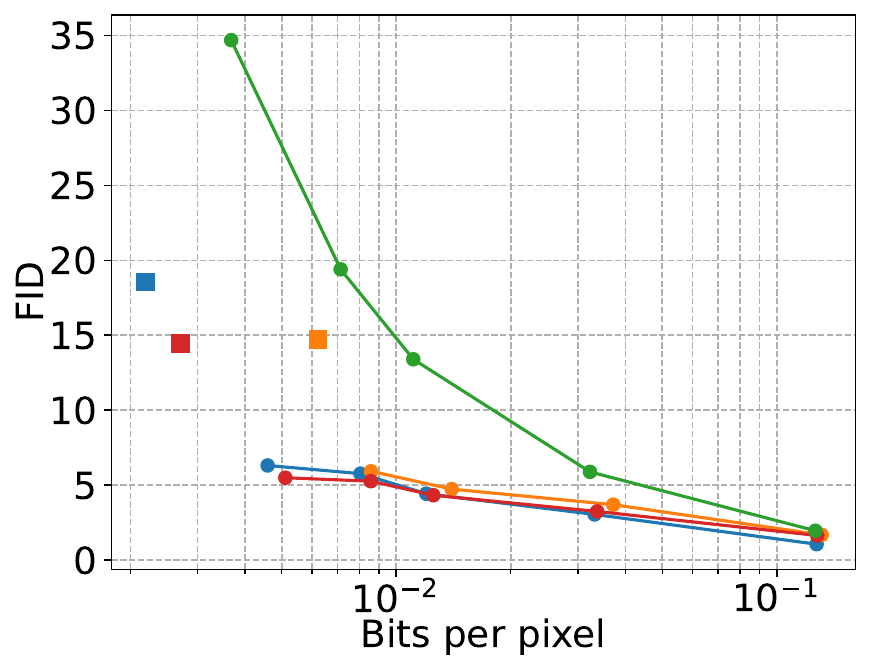}\myim{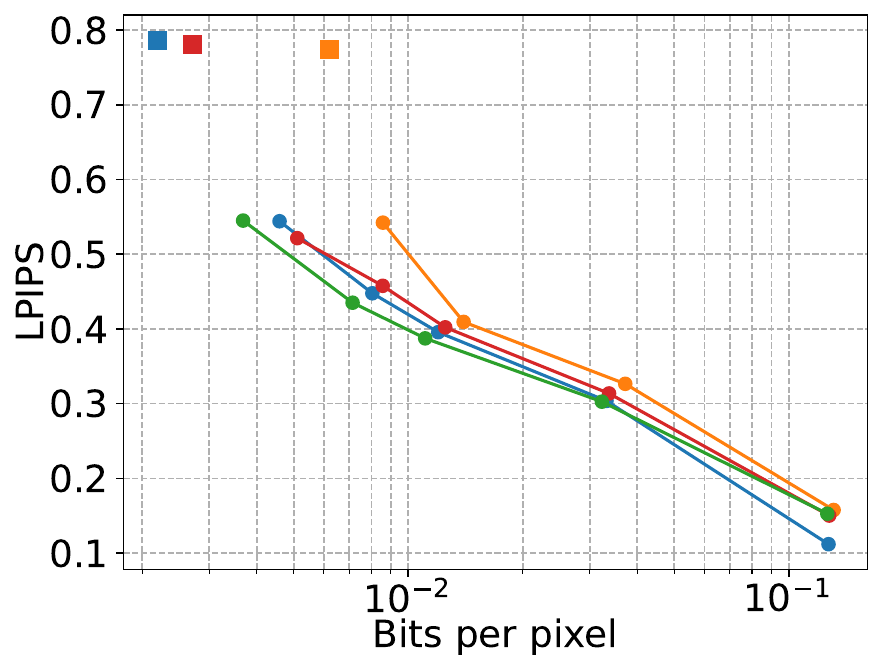}\myim{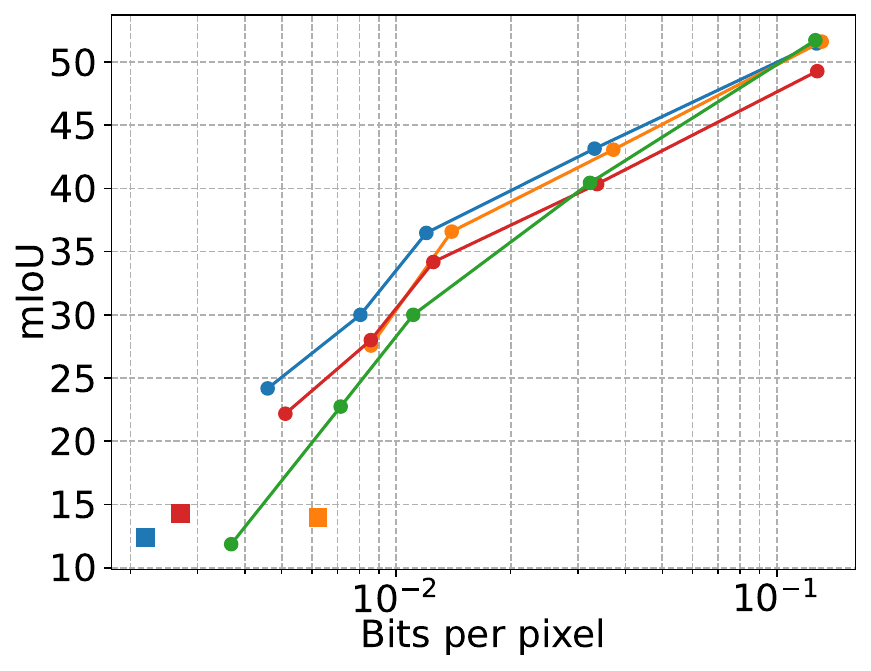}\myim{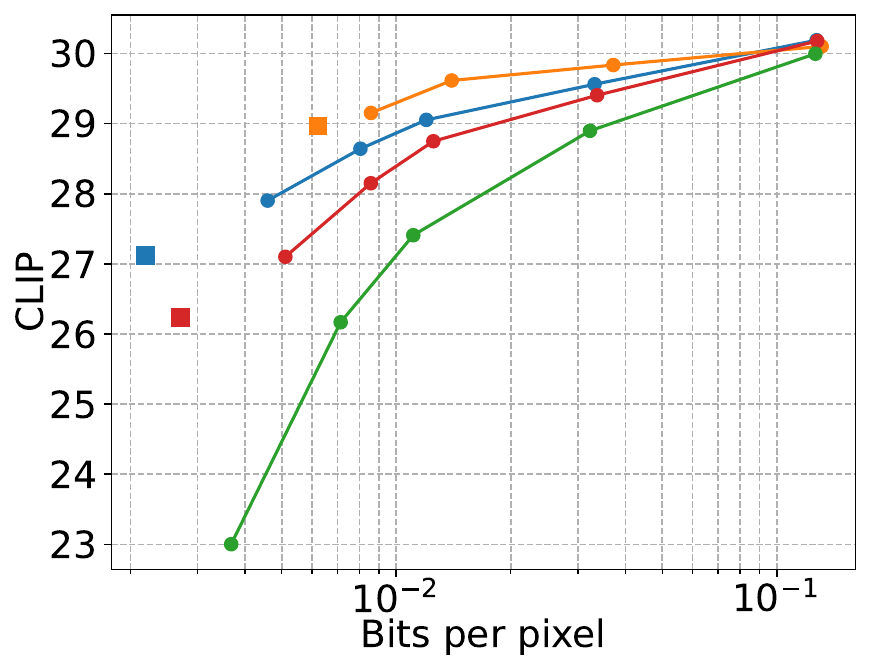}
\caption{Ablation of conditioning on  local  features and/or different captions. See text for detail. 
}
\label{fig:ablat_cond}
\end{figure}

\mypar{Conditioning modalities}
In \cref{fig:ablat_cond}, we evaluate \ours using each conditioning modality separately (textual or spatial), to analyze the contribution of each.
We  compare  BLIP-2~\citep{li23icml}  to generate captions (default) to using IDEFICS~\citep{idefics}  which produces more detailed descriptions, and ground-truth (GT) captions. When not using descriptive captions, we consider conditioning on a learned  textual embedding that is constant across images.  
We also tried using an empty text or on a fixed generic  text ---\emph{``A high quality photograph.''}--- for all images, and observed similar or slightly worse performance as the learned text embedding setting.

The model using the constant text embedding (\ours w/ only Spatial Encoding) performs significantly worse than the other models in terms of FID and CLIP score. 
For the LPIPS and mIoU metrics, which rely on the alignment of the original and reconstructed images, the  results are comparable to other models. 
This can be attributed to the fact that accurate alignment is primarily achieved through local feature conditioning, which remains unchanged in this model.
Consistently, we also notice that when using textual condition only (blue, orange,  and red square), the performance in terms of LPIPS and mIoU is  worse than for the models that use spatial conditioning, while the  CLIP scores are in line with the models using both textual and spatial conditioning. 
Both types of conditioning contribute to improving FID, and the text-only conditioned models perform in between the models with only spatial conditioning and those with both types of conditioning.

Between models using captions from BLIP (blue curves) and IDEFICS (orange curves) in combination with spatial conditioning, we observe similar FID  scores and somewhat better CLIP scores for the more detailed IDEFICS captions.
In terms of LPIPS and mIoU we find that the models with BLIP captions perform better, which is largely due to bitrate required to encode the longer captions: leaving out the textual encoding cost, both models perform very similarly.
Overall, using ground-truth captions (GT, red curves) yields similar results as using BLIP, while BLIP and IDEFICS yield slightly better mIoU and CLIP scores. 
As COCO contains five GT captions per image, we ensure that when  conditioning on a GT caption, a different one is used to compute the CLIP score.

\mypar{Classifier-free guidance} 
In \cref{tab:clsfree_abl} (left), we compare three options regarding classifier-free guidance (CFG) in our model  (i) not using CFG, (ii) applying CFG on the text and spatial feature maps, \ie contrasting full conditioning to no conditioning (we use an constant learned text embedding, and fixed  all-zeros feature maps for local conditioning), and (iii) applying CFG to text-conditioning only, \ie always conditioning on the spatial feature map. 
We find that best results are obtained when CFG is applied to textual conditioning only, which yields significantly better FID and mIoU metrics.

\mypar{Impact of  autoencoder and quantization}
To assess to which extent performance is bounded by the LDM autoencoder or the use of limited bitrate, we compare \ours to the LDM encoder/decoder reconstruction and to \ours trained without quantization in \cref{tab:clsfree_abl} (right).
We observe that the LDM autoencoder significantly improves all metrics, and that 
\ours trained without quantization obtains roughly the same performance as the autoencoder. 
This shows that \ours is a strong conditional model, and that its performance is mainly bounded by the quantization module which throttles the bitrate of the image encoding, but could also lead to suboptimal training through noisy gradients.

\begin{table}
\begin{center}  
\resizebox{\linewidth}{!}{
\scriptsize
\setlength{\tabcolsep}{1pt}
\begin{tabular}{lrrrrr}
 \toprule
    \textbf{Guidance type}  & $\downarrow$\textbf{FID} &   $\downarrow$\textbf{LPIPS} & $\uparrow$\textbf{SSIM}  &$\uparrow$\textbf{mIoU} &$\uparrow$\textbf{CLIP}  \\ 
    \midrule
    None             &  7.89 & 0.40 & 0.58 & 30.84 &28.36\\
    Text and spatial & 13.84 & 0.41 & 0.58 & 32.26 & 28.87\\
    \ours: Text only &  4.42 & 0.39& 0.58 & 46.64 & 29.05\\
 \bottomrule
\end{tabular}
\;\;\; 
\begin{tabular}{lrrrrr}
 \toprule
    \textbf{Method}  & $\downarrow$\textbf{FID} &  $\downarrow$\textbf{LPIPS} & $\uparrow$\textbf{SSIM }& $\uparrow$\textbf{mIoU} & $\uparrow$\textbf{CLIP}   \\ 
    \midrule
    LDM autoencoder          & 0.67& 0.07 & 0.92 & 51.04 & 30.19  \\ 
    \ours w/o quant. & 0.68 & 0.07 & 0.92 & 51.03 & 30.18\\
  
  \ours  (0.012 bpp) & 4.42 & 0.39& 0.58 & 46.64 & 29.05\\
 \bottomrule
\end{tabular} 
}
\end{center}
\caption{
Ablations of  classifier-free guidance forms (left), and impact of conditional diffusion model and quantization on reconstruction abilities (right). 
All models at 0.0112 bpp, except for LDM autoencoder and \ours w/o quantization which  use single precision for non-quantized latents.
}
\label{tab:clsfree_abl}
\end{table}

\begin{figure}
 \def\myim#1{ \includegraphics[width=0.25\textwidth]{figs/abl_numsteps/#1}}
 \def\myimLeg#1{ \includegraphics[width=.9\textwidth]{figs/abl_numsteps/#1}}
     \centering
   \setlength\tabcolsep{0.5 pt}
   \renewcommand{\arraystretch}{0.2}
\scriptsize
\begin{tabular}{cccc}
\multicolumn{4}{c}{\myimLeg{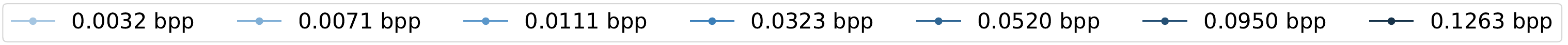}}\\
\myim{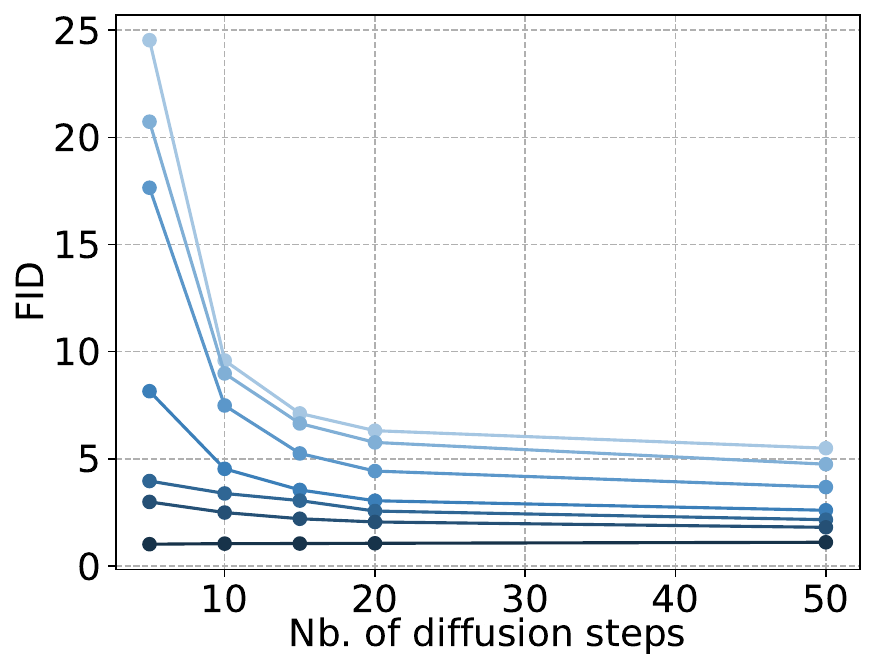} &\myim{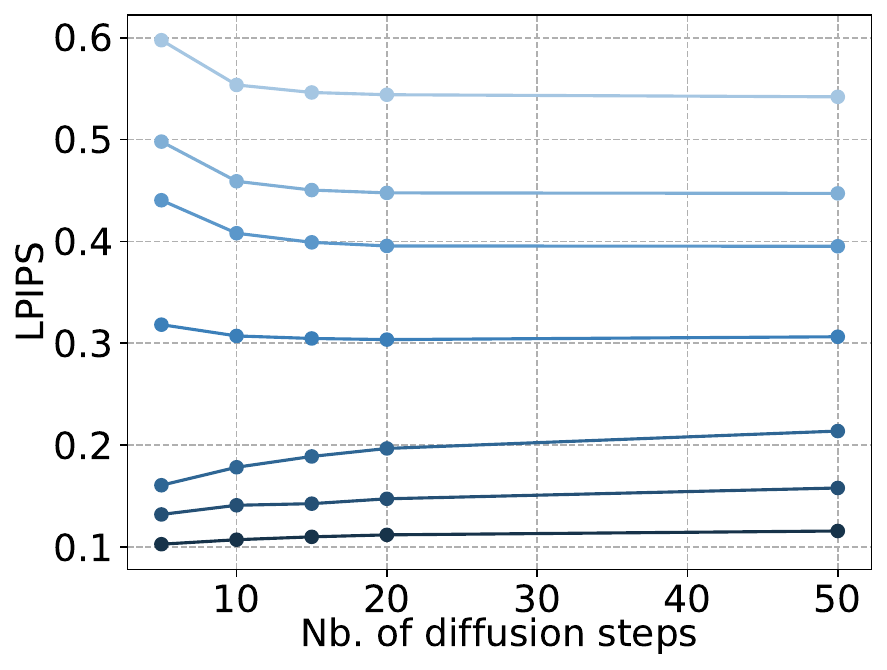} &  \myim{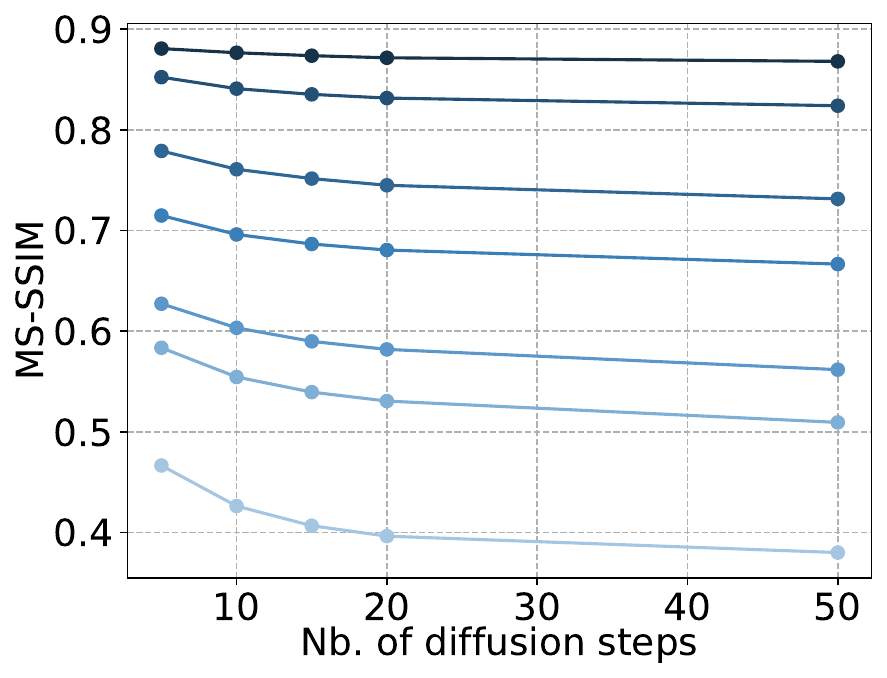} & \myim{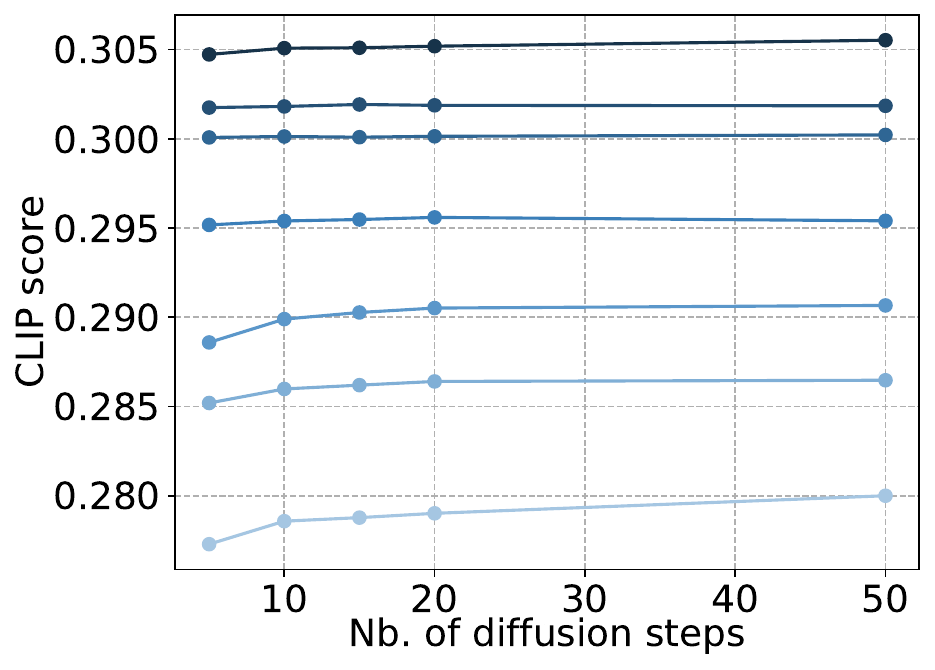} 
\end{tabular}
\caption{Evolution of FID, LPIPS, MS-SSIM and CLIP scores depending on the number of denoising steps for different bitrates.
In all cases the model has been trained using 50 denoising steps.
}
\label{fig:diff_iters}
\end{figure}

\mypar{Number of timesteps as a tradeoff between perception and distortion} In Figure \ref{fig:diff_iters}, we show how FID, LPIPS, MS-SSIM and CLIP scores evolve along the  denoising steps. 
We do the evaluation for seven different bitrates, ranging from 0.0032 to 0.1263 bpp. 
First, we notice that for higher bitrates, the metrics are quite constant across the  number of timesteps, which means that the model has less uncertainty about the image to reconstruct and obtains optimal performance after only few denoising steps. 
Secondly, we observe a tradeoff between distortion and realism for lower bitrates. 
Indeed, when the number of timesteps is small, FID is worse, but MS-SSIM is optimal, while this phenomenon reverses when increasing the number of denoising steps.

\mypar{Conditioning on global image features} 
Besides the text-conditioned model described above and used in most experiments (\ours), we also experimented with a  model conditioned on image embeddings extracted from   
an image backbone network.
While we use lossless Lempel-Ziv coding to compress the captions for the text-conditioned model,  we  use product quantization~\citep{5432202} for lossy compression of the image embeddings. 
We divide the  embeddings into $M$ subvectors and quantize each subvector using a separate codebook with $V$ elements. 
We empirically found that we achieve similar performance compared to training with  non-quantized image embeddings with $M=16$ and $V=1024$. 
For  $512\!\times\!512$ images,  the bitrate obtained for the image embeddings is 0.00061 bpp, which is 3.5$\times$ lower than the bitrate of 0.00219 bpp for BLIP image captions. 
From the results in \cref{fig:main_result}, we observe that the model with global image embeddings obtains competitive results, but has slightly worse pairwise image reconstruction metrics at high bitrates in terms of LPIPS and MS-SSIM on Kodak and MS-COCO 30k.

\section{Conclusion}

We proposed \ours, an image compression model that combines a VQ-VAE-like encoder with a diffusion-based decoder, and includes an second conditioning stream based on  textual image descriptions.
The iterative diffusion decoder allows for more realistic reconstructions, in particular at very low bitrates, as compared to feed-forward decoders used in previous work, even when trained with perceptual and adversarial losses.
With our work we make step towards perfect realism codecs: we observe realistic reconstructions at bitrates as low as 0.003 bits per pixel, a bitrate regime that to our knowledge is explored in only one prior work~\citep{lei23ncw}.
We find that 
semantics-driven metrics such as CLIP score and mIoU are improved overall and in particular for low rates, and  FID and KID  are dramatically improved at low bitrates and much more stable across different bitrates. 

\mypar{Limitations}
In this study we focused on medium sized images up to $768\!\times\!512$ resolution, similar to prior work on diffusion-based methods~\citep{theis2022lossy,lei23ncw} that evaluated on $64\!\times\!64$ and $512\!\times\!512$ images, respectively. 
Extension to higher resolutions can possibly be achieved using a patch-based approach, see \eg \cite{hoogeboom2023high}. 
\ours exhibits somewhat poorer reconstruction performance in terms MS-SSIM, PSNR and LPIPS (the latter for rates $>$0.01), than existing approaches. 
This seems at least in part to be due to limitations of the LDM autoencoder, see \cref{app:qualitative}. It is also probably explained by the tradeoff existing between distortion and perception~\citep{blau2019rethinking}.
We leave detailed study of these two points to future work.

\subsection*{Ethics statement}
The use of machine learning models, in general and for image compression in particular, can lead to biases in the model related to the training data, and potentially work better on data similar to those used to train the models.
Therefore the models as described in this paper should be used for research purposes only, and not deployed in practice without extensive analysis of their potential biases.

\subsection*{Acknowledgements}
The photo in \Cref{fig:flowchart} is by Emily Allen, licensed as CC BY 2.0, and taken from \url{https://www.flickr.com/photos/99453901@N00/20263907283}.
We would like to thank Oron Ashual, Uriel Singer, Adam Polyak and Shelly Sheynin for sharing the text-to-image model.

\bibliography{iclr2024_conference}

\begin{thebibliography}{51}
\providecommand{\natexlab}[1]{#1}
\providecommand{\url}[1]{\texttt{#1}}
\expandafter\ifx\csname urlstyle\endcsname\relax
  \providecommand{\doi}[1]{doi: #1}\else
  \providecommand{\doi}{doi: \begingroup \urlstyle{rm}\Url}\fi

\bibitem[vtm()]{vtm}
{VTM} reference software for {VVC}.
\newblock URL \url{https://vcgit.hhi.fraunhofer.de/jvet/VVCSoftware_VTM}.

\bibitem[zli()]{zlib}
zlib - compression compatible with gzip.
\newblock URL \url{https://docs.python.org/3/library/zlib.html}.

\bibitem[Agustsson et~al.(2019)Agustsson, Tschannen, Mentzer, Timofte, and Gool]{agustsson2019generative}
Eirikur Agustsson, Michael Tschannen, Fabian Mentzer, Radu Timofte, and Luc~Van Gool.
\newblock Generative adversarial networks for extreme learned image compression.
\newblock In \emph{ICCV}, 2019.

\bibitem[Agustsson et~al.(2023)Agustsson, Minnen, Toderici, and Mentzer]{agustsson2023multi}
Eirikur Agustsson, David Minnen, George Toderici, and Fabian Mentzer.
\newblock Multi-realism image compression with a conditional generator.
\newblock In \emph{CVPR}, 2023.

\bibitem[Ball{\'e} et~al.(2017)Ball{\'e}, Laparra, and Simoncelli]{balle2017end}
Johannes Ball{\'e}, Valero Laparra, and Eero~P Simoncelli.
\newblock End-to-end optimized image compression.
\newblock In \emph{ICLR}, 2017.

\bibitem[Ball\'e et~al.(2018)Ball\'e, Minnen, Singh, Hwang, and Johnston]{balle2018variational}
Johannes Ball\'e, David Minnen, Saurabh Singh, Sung~Jin Hwang, and Nick Johnston.
\newblock Variational image compression with a scale hyperprior.
\newblock In \emph{ICLR}, 2018.

\bibitem[Bi{\'n}kowski et~al.(2018)Bi{\'n}kowski, Sutherland, Arbel, and Gretton]{binkowski2018demystifying}
Miko{\l}aj Bi{\'n}kowski, Danica~J Sutherland, Michael Arbel, and Arthur Gretton.
\newblock Demystifying {MMD} {GANs}.
\newblock In \emph{ICLR}, 2018.

\bibitem[Blau \& Michaeli(2018)Blau and Michaeli]{blau2018perception}
Yochai Blau and Tomer Michaeli.
\newblock The perception-distortion tradeoff.
\newblock In \emph{CVPR}, 2018.

\bibitem[Blau \& Michaeli(2019)Blau and Michaeli]{blau2019rethinking}
Yochai Blau and Tomer Michaeli.
\newblock Rethinking lossy compression: The rate-distortion-perception tradeoff.
\newblock In \emph{ICML}, 2019.

\bibitem[Caesar et~al.(2018)Caesar, Uijlings, and Ferrari]{caesar2018coco}
Holger Caesar, Jasper Uijlings, and Vittorio Ferrari.
\newblock {COCO-Stuff}: Thing and stuff classes in context.
\newblock In \emph{CVPR}, 2018.

\bibitem[Chen et~al.(2023)Chen, Duan, Wang, He, Lu, Dai, and Qiao]{chen2022vitadapter}
Zhe Chen, Yuchen Duan, Wenhai Wang, Junjun He, Tong Lu, Jifeng Dai, and Yu~Qiao.
\newblock Vision transformer adapter for dense predictions.
\newblock In \emph{ICLR}, 2023.

\bibitem[Cheng et~al.(2020)Cheng, Sun, Takeuchi, and Katto]{cheng2020joint}
Zhengxue Cheng, Heming Sun, Masaru Takeuchi, and Jiro Katto.
\newblock Learned image compression with discretized {Gaussian} mixture likelihoods and attention modules.
\newblock In \emph{CVPR}, 2020.

\bibitem[Dhariwal \& Nichol(2021)Dhariwal and Nichol]{dhariwal2021diffusion}
Prafulla Dhariwal and Alex Nichol.
\newblock Diffusion models beat {GAN}s on image synthesis.
\newblock In \emph{NeurIPS}, 2021.

\bibitem[El-Nouby et~al.(2023)El-Nouby, Muckley, Ullrich, Laptev, Verbeek, and J\'egou]{elnouby2023image}
Alaaeldin El-Nouby, Matthew~J. Muckley, Karen Ullrich, Ivan Laptev, Jakob Verbeek, and Herv\'e J\'egou.
\newblock Image compression with product quantized masked image modeling.
\newblock \emph{Trans. Mach. Learn. Res.}, 2023.

\bibitem[Franzen(1999)]{kodak}
Rich Franzen.
\newblock Kodak {PhotoCD} dataset, 1999.
\newblock URL \url{http://r0k.us/graphics/kodak/}.

\bibitem[Ghouse et~al.(2023)Ghouse, Petersen, Wiggers, Xu, and Sauti{\`e}re]{ghouse2023residual}
Noor~Fathima Ghouse, Jens Petersen, Auke Wiggers, Tianlin Xu, and Guillaume Sauti{\`e}re.
\newblock A residual diffusion model for high perceptual quality codec augmentation.
\newblock \emph{arXiv preprint}, 2301.05489, 2023.

\bibitem[He et~al.(2022{\natexlab{a}})He, Yang, Peng, Ma, Qin, and Wang]{he2022elic}
Dailan He, Ziming Yang, Weikun Peng, Rui Ma, Hongwei Qin, and Yan Wang.
\newblock {ELIC}: Efficient learned image compression with unevenly grouped space-channel contextual adaptive coding.
\newblock In \emph{CVPR}, June 2022{\natexlab{a}}.

\bibitem[He et~al.(2022{\natexlab{b}})He, Yang, Yu, Xu, Luo, Chen, Gao, Shi, Qin, and Wang]{he2022po}
Dailan He, Ziming Yang, Hongjiu Yu, Tongda Xu, Jixiang Luo, Yuan Chen, Chenjian Gao, Xinjie Shi, Hongwei Qin, and Yan Wang.
\newblock {PO-ELIC}: Perception-oriented efficient learned image coding.
\newblock In \emph{CVPR}, 2022{\natexlab{b}}.

\bibitem[Hessel et~al.(2021)Hessel, Holtzman, Forbes, Bras, and Choi]{hessel21emnlp}
Jack Hessel, Ari Holtzman, Maxwell Forbes, Ronan~Le Bras, and Yejin Choi.
\newblock {CLIPScore}: A reference-free evaluation metric for image captioning.
\newblock In \emph{EMNLP}, 2021.

\bibitem[Heusel et~al.(2017)Heusel, Ramsauer, Unterthiner, Nessler, and Hochreiter]{heusel2017gans}
Martin Heusel, Hubert Ramsauer, Thomas Unterthiner, Bernhard Nessler, and Sepp Hochreiter.
\newblock {GANs} trained by a two time-scale update rule converge to a local {Nash} equilibrium.
\newblock \emph{NeurIPS}, 2017.

\bibitem[Ho \& Salimans(2021)Ho and Salimans]{ho21cfg}
J.~Ho and T.~Salimans.
\newblock Classifier-free diffusion guidance.
\newblock In \emph{NeurIPS Workshop on Deep Generative Models and Downstream Applications}, 2021.

\bibitem[Ho et~al.(2020)Ho, Jain, and Abbeel]{ho2020denoising}
Jonathan Ho, Ajay Jain, and Pieter Abbeel.
\newblock Denoising diffusion probabilistic models.
\newblock In \emph{NeurIPS}, volume~33, 2020.

\bibitem[Hoogeboom et~al.(2023)Hoogeboom, Agustsson, Mentzer, Versari, Toderici, and Theis]{hoogeboom2023high}
Emiel Hoogeboom, Eirikur Agustsson, Fabian Mentzer, Luca Versari, George Toderici, and Lucas Theis.
\newblock High-fidelity image compression with score-based generative models.
\newblock \emph{arXiv preprint}, 2305.18231, 2023.

\bibitem[Jégou et~al.(2011)Jégou, Douze, and Schmid]{5432202}
Herve Jégou, Matthijs Douze, and Cordelia Schmid.
\newblock Product quantization for nearest neighbor search.
\newblock \emph{IEEE TPAMI}, 33\penalty0 (1):\penalty0 117--128, 2011.

\bibitem[Kingma et~al.(2021)Kingma, Salimans, Poole, and Ho]{kingma2021variational}
Diederik Kingma, Tim Salimans, Ben Poole, and Jonathan Ho.
\newblock Variational diffusion models.
\newblock In \emph{NeurIPS}, 2021.

\bibitem[Kuznetsova et~al.(2020)Kuznetsova, Rom, Alldrin, Uijlings, Krasin, Pont-Tuset, Kamali, Popov, Malloci, Kolesnikov, et~al.]{kuznetsova2020open}
Alina Kuznetsova, Hassan Rom, Neil Alldrin, Jasper Uijlings, Ivan Krasin, Jordi Pont-Tuset, Shahab Kamali, Stefan Popov, Matteo Malloci, Alexander Kolesnikov, et~al.
\newblock The {Open} {Images} {Dataset} {V4}: Unified image classification, object detection, and visual relationship detection at scale.
\newblock \emph{IJCV}, 128\penalty0 (7):\penalty0 1956--1981, 2020.

\bibitem[Lauran\c{c}on et~al.(2023)Lauran\c{c}on, van Strien, Bekman, Tronchon, Saulnier, Wang, Karamcheti, Singh, Pistilli, Jernite, and Sanh]{idefics}
Hugo Lauran\c{c}on, Daniel van Strien, Stas Bekman, Leo Tronchon, Lucile Saulnier, Thomas Wang, Siddharth Karamcheti, Amanpreet Singh, Giada Pistilli, Yacine Jernite, and Victor Sanh.
\newblock Introducing {IDEFICS}: An open reproduction of state-of-the-art visual language model, 2023.
\newblock URL \url{https://huggingface.co/blog/idefics}.

\bibitem[Lei et~al.(2023)Lei, Uslu, Hassani, and Bidokht]{lei23ncw}
Eric Lei, Yigit~Berkay Uslu, Hamed Hassani, and Shirin~Saeedi Bidokht.
\newblock Text + sketch: Image compression at ultra low rates.
\newblock In \emph{{ICML} Neural Compression Workshop}, 2023.

\bibitem[Li et~al.(2023)Li, Li, Savarese, and Hoi]{li23icml}
Junnan Li, Dongxu Li, Silvio Savarese, and Steven Hoi.
\newblock {BLIP-2}: Bootstrapping language-image pre-training with frozen image encoders and large language models.
\newblock In \emph{ICML}, 2023.

\bibitem[Lin et~al.(2014)Lin, Maire, Belongie, Bourdev, Girshick, Hays, Perona, Ramanan, Doll{\'{a}}r, and Zitnick]{lin14eccv}
T.{-}Y. Lin, M.~Maire, S.~Belongie, L.~Bourdev, R.~Girshick, J.~Hays, P.~Perona, D.~Ramanan, P.~Doll{\'{a}}r, and C.~Zitnick.
\newblock Microsoft {COCO:} common objects in context.
\newblock In \emph{ECCV}, 2014.

\bibitem[Mentzer et~al.(2020)Mentzer, Toderici, Tschannen, and Agustsson]{mentzer2020high}
Fabian Mentzer, George~D Toderici, Michael Tschannen, and Eirikur Agustsson.
\newblock High-fidelity generative image compression.
\newblock In \emph{NeurIPS}, 2020.

\bibitem[Minnen \& Singh(2020)Minnen and Singh]{minnen2020channel}
David Minnen and Saurabh Singh.
\newblock Channel-wise autoregressive entropy models for learned image compression.
\newblock In \emph{ICIP}, 2020.

\bibitem[Minnen et~al.(2018)Minnen, Ball\'{e}, and Toderici]{minnen2018joint}
David Minnen, Johannes Ball\'{e}, and George~D Toderici.
\newblock Joint autoregressive and hierarchical priors for learned image compression.
\newblock In \emph{NeurIPS}, 2018.

\bibitem[Muckley et~al.(2023)Muckley, El-Nouby, Ullrich, J\'egou, and Verbeek]{muckley2023improving}
Matthew~J. Muckley, Alaaeldin El-Nouby, Karen Ullrich, Herv\'e J\'egou, and Jakob Verbeek.
\newblock Improving statistical fidelity for neural image compression with implicit local likelihood models.
\newblock In \emph{ICML}, 2023.

\bibitem[Nichol et~al.(2022)Nichol, Dhariwal, Ramesh, Shyam, Mishkin, {McGrew}, Sutskever, and Chen]{nichol21arxiv}
Alex Nichol, Prafulla Dhariwal, Aditya Ramesh, Pranav Shyam, Pamela Mishkin, Bob {McGrew}, Ilya Sutskever, and Mark Chen.
\newblock {GLIDE}: Towards photorealistic image generation and editing with text-guided diffusion models.
\newblock In \emph{ICML}, 2022.

\bibitem[Pan et~al.(2022)Pan, Zhou, and Tian]{pan2022extreme}
Zhihong Pan, Xin Zhou, and Hao Tian.
\newblock Extreme generative image compression by learning text embedding from diffusion models.
\newblock \emph{arXiv preprint}, 2211.07793, 2022.

\bibitem[Razavi et~al.(2019)Razavi, van~den Oord, and Vinyals]{razavi19nips}
Ali Razavi, Aaron van~den Oord, and Oriol Vinyals.
\newblock Generating diverse high-fidelity images with {VQ-VAE-2}.
\newblock In \emph{NeurIPS}, 2019.

\bibitem[Saharia et~al.(2022)Saharia, Ho, Chan, Salimans, Fleet, and Norouzi]{saharia2022image}
Chitwan Saharia, Jonathan Ho, William Chan, Tim Salimans, David~J Fleet, and Mohammad Norouzi.
\newblock Image super-resolution via iterative refinement.
\newblock \emph{IEEE TPAMI}, 45\penalty0 (4):\penalty0 4713--4726, 2022.

\bibitem[Salimans \& Ho(2022)Salimans and Ho]{salimans22iclr}
Tim Salimans and Jonathan Ho.
\newblock Progressive distillation for fast sampling of diffusion models.
\newblock In \emph{ICLR}, 2022.

\bibitem[Shannon(1948)]{shannon1948mathematical}
C.~E. Shannon.
\newblock A mathematical theory of communication.
\newblock \emph{Bell Syst. Tech. J.}, 27\penalty0 (3):\penalty0 379--423, 1948.

\bibitem[Sohl-Dickstein et~al.(2015)Sohl-Dickstein, Weiss, Maheswaranathan, and Ganguli]{sohl2015deep}
Jascha Sohl-Dickstein, Eric Weiss, Niru Maheswaranathan, and Surya Ganguli.
\newblock Deep unsupervised learning using nonequilibrium thermodynamics.
\newblock In \emph{ICML}, 2015.

\bibitem[Song et~al.(2021)Song, Meng, and Ermon]{song2020denoising}
Jiaming Song, Chenlin Meng, and Stefano Ermon.
\newblock Denoising diffusion implicit models.
\newblock In \emph{ICLR}, 2021.

\bibitem[Sushko et~al.(2022)Sushko, Sch{\"o}nfeld, Zhang, Gall, Schiele, and Khoreva]{sushko2022oasis}
Vadim Sushko, Edgar Sch{\"o}nfeld, Dan Zhang, Juergen Gall, Bernt Schiele, and Anna Khoreva.
\newblock {OASIS}: Only adversarial supervision for semantic image synthesis.
\newblock \emph{IJCV}, 130\penalty0 (12):\penalty0 2903--2923, 2022.

\bibitem[Theis et~al.(2022)Theis, Salimans, Hoffman, and Mentzer]{theis2022lossy}
Lucas Theis, Tim Salimans, Matthew~D Hoffman, and Fabian Mentzer.
\newblock Lossy compression with {G}aussian diffusion.
\newblock \emph{arXiv preprint}, 2206.08889, 2022.

\bibitem[van~den Oord et~al.(2017)van~den Oord, Vinyals, and Kavukcuoglu]{oord2017neural}
Aaron van~den Oord, Oriol Vinyals, and koray Kavukcuoglu.
\newblock Neural discrete representation learning.
\newblock In \emph{NeurIPS}, 2017.

\bibitem[Wang et~al.(2003)Wang, Simoncelli, and Bovik]{wang2003multiscale}
Z.~Wang, E.P. Simoncelli, and A.C. Bovik.
\newblock Multiscale structural similarity for image quality assessment.
\newblock In \emph{ACSSC}, volume~2, pp.\  1398--1402, 2003.

\bibitem[Wen et~al.(2023)Wen, Jain, Kirchenbauer, Goldblum, Geiping, and Goldstein]{wen23arxiv}
Yuxin Wen, Neel Jain, John Kirchenbauer, Micah Goldblum, Jonas Geiping, and Tom Goldstein.
\newblock Hard prompts made easy: Gradient-based discrete optimization for prompt tuning and discovery.
\newblock \emph{arXiv preprint}, 2302.03668, 2023.

\bibitem[Xu et~al.(2023)Xu, Zhang, Li, He, Wang, Wang, Qin, Wang, Liu, and Zhang]{xu2023conditional}
Tongda Xu, Qian Zhang, Yanghao Li, Dailan He, Zhe Wang, Yuanyuan Wang, Hongwei Qin, Yan Wang, Jingjing Liu, and Ya-Qin Zhang.
\newblock Conditional perceptual quality preserving image compression.
\newblock \emph{arXiv preprint}, 2308.08154, 2023.

\bibitem[Yang \& Mandt(2023)Yang and Mandt]{yang2022lossy}
Ruihan Yang and Stephan Mandt.
\newblock Lossy image compression with conditional diffusion models.
\newblock In \emph{{ICML} Neural Compression Workshop}, 2023.

\bibitem[Yu et~al.(2022)Yu, Li, Koh, Zhang, Pang, Qin, Ku, Xu, and Jason~Baldridge]{yu22iclr}
Jiahui Yu, Xin Li, Jing~Yu Koh, Han Zhang, Ruoming Pang, James Qin, Alexander Ku, Yuanzhong Xu, and Yonghui~Wu Jason~Baldridge.
\newblock Vector-quantized image modeling with improved {VQGAN}.
\newblock In \emph{ICLR}, 2022.

\bibitem[Zhang et~al.(2018)Zhang, Isola, Efros, Shechtman, and Wang]{zhang2018unreasonable}
Richard Zhang, Phillip Isola, Alexei~A. Efros, Eli Shechtman, and Oliver Wang.
\newblock The unreasonable effectiveness of deep features as a perceptual metric.
\newblock In \emph{CVPR}, 2018.

\end{thebibliography}
\bibliographystyle{iclr2024_conference}

\appendix

\section{Experimental details}
\label{app:exp_details}

\mypar{Evaluation of third-party models}
Where possible we reported evaluation metrics reported in the original papers.
For Text-Sketch~\cite{lei23ncw}, we used the publicly released checkpoints, and used them for evaluation with the provided code. 
Note that Text-Sketch resized all images at resolution 512 for evaluation on the DIV2K and CLIC 2021 datasets, which makes the comparison with other baselines not immediate.
We prefer to evaluate on original images instead, and for that purpose used the Kodak dataset which is commonly used to evaluate image compression models, and contains images of $512\!\times\!768$ pixels. We also evaluate on the MS-COCO 2014 validation set, which is less commonly used, but which offers ground truth semantic segmentation maps and descriptive captions which enable evaluation with the semantics oriented mIoU and CLIP score metrics.

For MS-ILLM and MS-PatchGAN~\citep{muckley2023improving}, checkpoints for models trained at additional low bitrates were obtained from the authors through personal communication.  For HFD/DDPM~\citep{hoogeboom2023high}, DIRAC~\citep{ghouse2023residual} and Multi-Realism~\citep{agustsson2023multi}, we obtained the numbers shown in their paper through personal communication, as there are no available github. 
To evaluate VTM, we followed same the procedure as described by \cite{muckley2023improving} in appendix B of their paper. 
For HifiC~\citep{mentzer2020high}, we took the checkpoints available  from their github.\footnote{\url{https://github.com/tensorflow/compression/tree/master/models/hific}}

 \mypar{Additional implementation details}
We take a pretrained latent diffusion model doing text-to-image generation that contains 1.4B parameters. We fix the weights of the LDM autoencoder and  finetune 15$\%$ of the diffusion model weights. 
We  train our hyper-encoder for 5 epochs with AdamW optimizer at a peak learning rate of 1e-4 and a weight decay of 0.01, and a batch size of 160. 
We apply linear warmup for the first 10k training iterations. For bitrates higher that 0.05, we found it beneficial to add a LPIPS loss in image space to reconstruct more faithfully images. As this loss requires to backpropagate gradients through the LDM decoder (even though the LDM decoder is frozen), it is more memory intensive and we had to decrease the batch size to 40. 
For the vector quantization module, we used the vector-quantize-pytorch
library.\footnote{\url{https://github.com/lucidrains/vector-quantize-pytorch}}

To finetune our latent diffusion model we used a grid of 50 timesteps. 
At inference time, we use 20 denoising steps  to obtain good performance for the lowest bitrates. 
For  bitrates of 0.05 and higher, we use five denoising steps as this is  enough to obtain optimal performance, see \cref{fig:diff_iters} from the main paper. 
Besides, we sample images with DDIM deterministic sampling schedule, \ie by fixing $\sigma_t=0$ for all $t \in [0,T]$, where $T$ is the number of denoising steps, see~\cite{song2020denoising}.

When testing different textual conditioning modalities (either BLIP or IDEFICS) in our ablation study in \cref{fig:ablat_cond}, the conditioning used at inference time is also used when training the models. When using human-specified captions, we use \ours trained on BLIP captions, as there are no ground-truth captions for the OpenImagesv6 dataset we used to train our model.

We use the clean-fid library to compute FID and KID.\footnote{\url{https://github.com/GaParmar/clean-fid}}. For CLIP score, we compute image and text embeddings with the CLIP backbone ViT-B/32. LPIPS is calculated using the original github\footnote{ \url{https://github.com/richzhang/PerceptualSimilarity}}, and MS-SSIM from the FAIR Neural Compression library.\footnote{ \url{https://github.com/facebookresearch/NeuralCompression}}

\begin{wraptable}{R}{0.4\textwidth}

\small
\centering
  {
\begin{tabular}{ccc}
 \toprule
   \textbf{Spatial size}  & \textbf{Codebook size} &    \textbf{ Spatial bpp} \\

    \midrule
    \multirow{2}{*}{ $64\!\times\!64$} &  256 & 0.1250\\ 
    
    &64 & 0.0937 \\
   
   \midrule
\multirow{2}{*}{ $32\!\times\!32$} &  8196& 0.0507 \\

& 256 & 0.0313  \\

   \midrule

\multirow{1}{*}{ $16\!\times\!16$}   &  1024 & 0.0098  \\
   \midrule
\multirow{2}{*}{ $8\!\times\!8$} &  1024 & 0.0024\\
& 256 & 0.0019  \\

 \bottomrule
\end{tabular} }
\caption{Summary of the different combinations of spatial grid size and codebook size used for spatial encoding of $512\!\times\!512$ images.
}
\label{bpp-table}
\end{wraptable}

 \mypar{Details on the bitrates  used to evaluate our models}
To achieve a specific target bitrate, we take into account the average bitrates required to encode the textual conditioning,  and then set the spatial grid size and vocabulary size of the local conditioning accordingly. 
More specifically, for COCO at resolution $512\!\times\!512$, the average textual bpp for BLIP is 0.00219, while it is 0.00622 for IDEFICS, and 0.00271 for ground truth captions. For Kodak at resolution $768\!\times\!512$, it is 0.00141 with BLIP and 0.00415 bpp with IDEFICS. 
In the experiments, we select the spatial size and codebook dimension of the quantization module to adjust the spatial encoding bitrate depending on the targeted total bitrate. 
We found that codebooks with less than  64 elements  were detrimental to the training, so we only used larger  codebooks.
\Cref{bpp-table} lists the bpp for the spatial coding for the configurations we used to proeuce the results in our experiments.
For the ablation study in \cref{fig:ablat_cond} we used other combinations of spatial grid size and vocabulary size for the tested variants to approximately match the bitrates of our default \ours model.

 \mypar{Inference Speed}
We run benchmarking on A100 GPUs using 20 denoising steps, and 5 denoising steps for bitrates higher than 0.05 bits per pixel. 
We compute the runtime for all the Kodak images at resolution $512\times 768$ and provide the average time in seconds with its standard deviation in \cref{speed-table}.
The encoding complexity of PerCo is quite similar to adversarial-based compression methods such as MS-ILLM and HiFiC (which has the same architecture as MS-ILLM). Encoding with Text-Sketch (PIC/PICS) is orders of magnitude slower because it is optimizing a text embedding for each image.
The decoding complexity of PerCo using 20 denoising steps (bitrates below 0.05 bpp) is comparable to that of Text-Sketch (PIC) and about 8 times faster than Text-Sketch (PICS), because PICS is conditioning on sketches which takes time to decompress.
When using only five denoising steps, the decoding complexity drops rouhgly by a factor four, showing that the decoding is dominated by evaluations of the denoising network.
Compared to HiFiC and MS-ILLM, the PerCo decoder using 5 denoising steps is about eight times slower, and when using 20 steps this becomes a factor 32.
Please note that the running time of VTM is computed on CPU, so it may explain the longer runtime.
We could not evaluate  the diffusion-based  DIRAC and HFD/DDPM models as there is no public codebase available, but we found some evaluations in DIRAC paper about decoding time. They report a decoding time which is about four times slower than what they report for HiFiC, which also shows that diffusion-based methods are slower than adversarial-based ones.

The average decoding times for the 24 images of Kodak dataset are 0.67 secs.\ and   2.54 secs.\ when decoding with 5 and 20 steps, respectively. Timings is performed when decoding one image at a time.

\begin{table}[h]
\centering
  { \small
\begin{tabular}{lrr}
\toprule
   \textbf{Model}  & \textbf{Encoding Speed (in sec.)} &    \textbf{Decoding Speed (in sec.)} \\
\midrule
VTM     & 16.892 $\pm$ 7.574	& 0.135 $\pm$ 0.002\\
MS-ILLM &  0.084 $\pm$ 0.021	& 0.080 $\pm$ 0.007\\

Text-Sketch (PIC) & 163.070 $\pm$ 0.380	&  2.725 $\pm$ 0.012 \\
Text-Sketch (PICS)& 190.231 $\pm$ 2.476 & 19.288 $\pm$ 0.251 \\
 \midrule
PerCo - 5 denoising steps & 
0.080 $\pm$ 0.018	& 0.665 $\pm$ 0.009 \\
PerCo - 20 denoising steps & 
0.080 $\pm$ 0.018	& 2.551 $\pm$ 0.018 \\
 \bottomrule
\end{tabular} }
\caption{Encoding and decoding Speed (in seconds).
}
\label{speed-table}
\end{table}

\section{Additional experimental results}

\begin{figure}
\def\myim#1{ \includegraphics[width=0.35\textwidth]{figs/appendix/psnr/#1}}
\def\myimLeg#1{ \includegraphics[width=0.7\textwidth]{figs/appendix/psnr/#1}}
\centering
\myimLeg{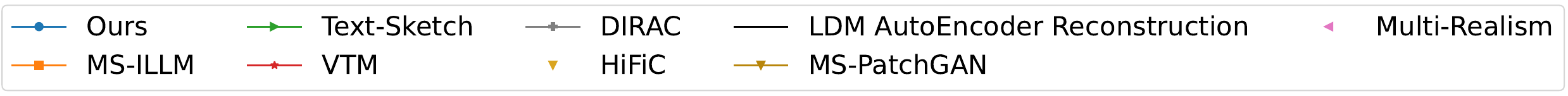} \\
\myim{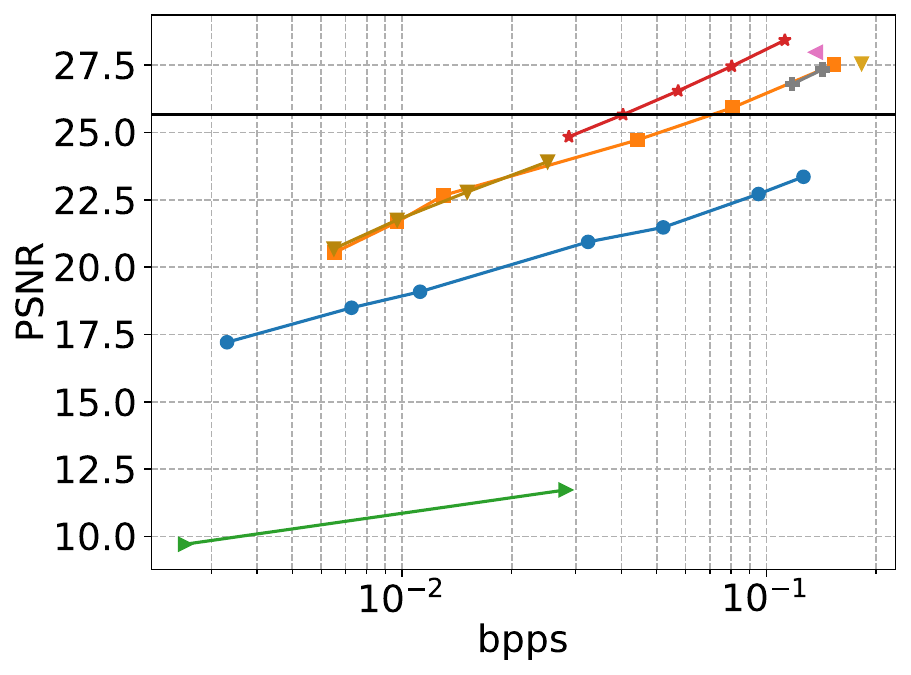} \quad   \myim{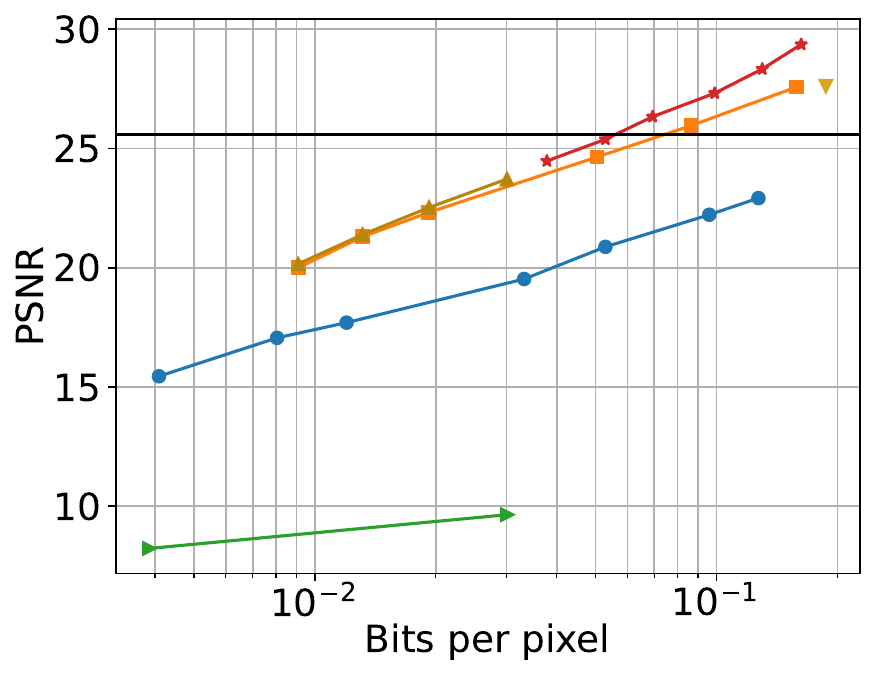}   
\caption{PSNR on Kodak (left) and MS-COCO30 (right). 
}
\label{fig:otherMetrics}
\end{figure}

\subsection{Quantitative results} 
\label{app:quant}
\begin{figure}
\def\myim#1{\includegraphics[height=0.27\textwidth]{figs/appendix/psnr_quali/#1}}
\begin{center}
   \setlength\tabcolsep{1 pt}
   \renewcommand{\arraystretch}{0.2}
\scriptsize
\begin{tabular}{ccc}
\myim{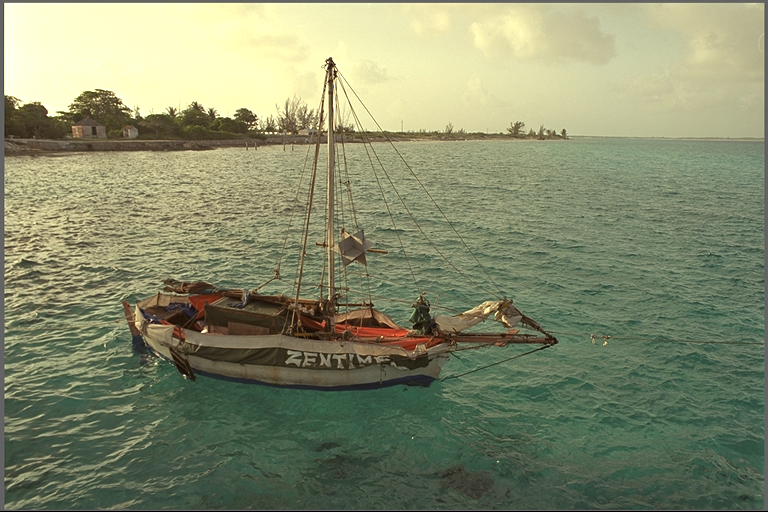}  & \myim{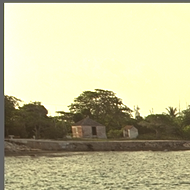} & \myim{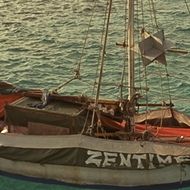} \\
\multicolumn{3}{c}{Original}\\
\myim{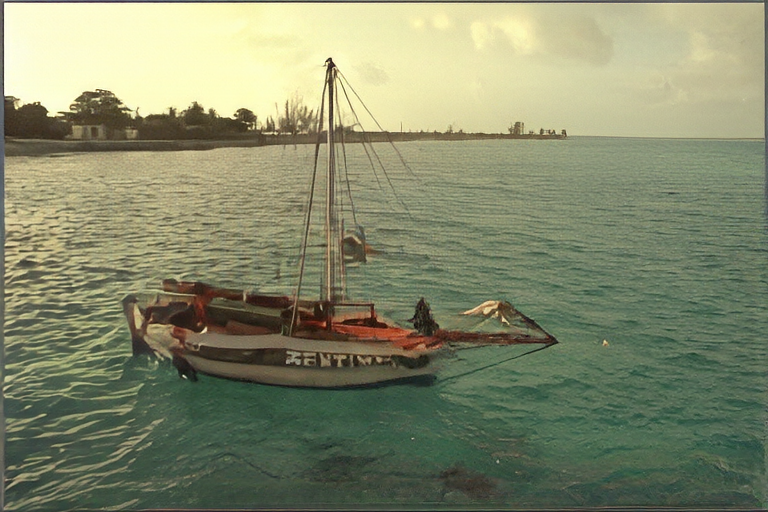} & \myim{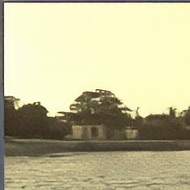} & \myim{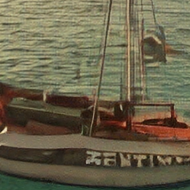}\\
  \multicolumn{3}{c}{MS-ILLM, 0.0806 bpp, 24.44  PSNR, 0.8547 MS-SSIM, 0.1095 LPIPS}  \\

\myim{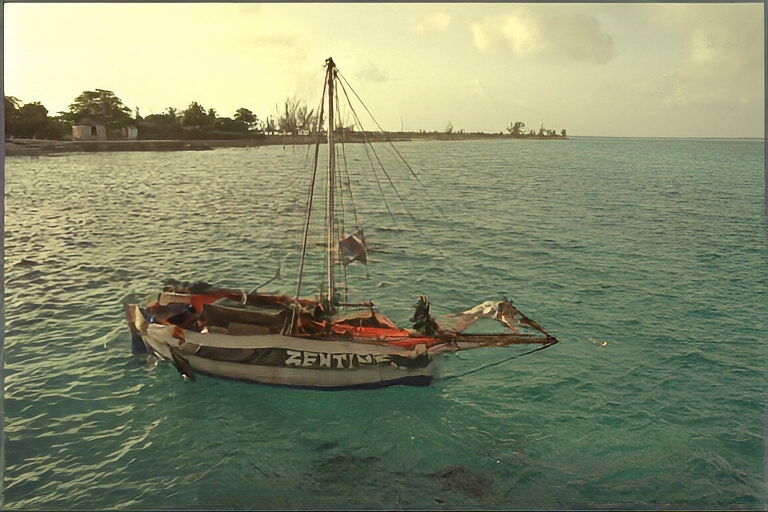} & \myim{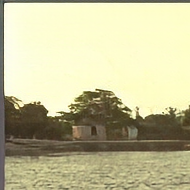} & \myim{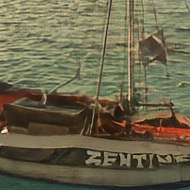}\\
  \multicolumn{3}{c}{MS-ILLM, 0.1535 bpp,  25.97 PSNR, 0.9055 MS-SSIM, 0.0771 LPIPS}  \\

\myim{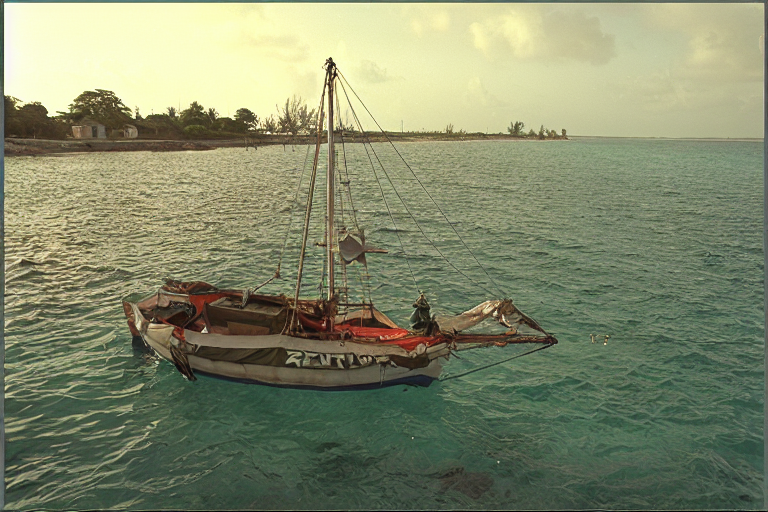}  & \myim{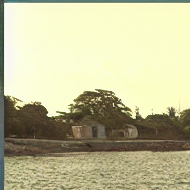} & \myim{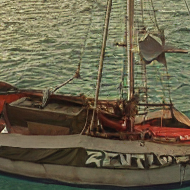} \\

 \multicolumn{3}{c}{ LDM autoencoder,  23.17 PSNR, 0.8674 MS-SSIM, 0.0880 LPIPS }
\end{tabular}     
\end{center}     
\caption{Qualitative comparison of reconstructions using  MS-ILLM and our LDM autoencoder at comparable PSNR of a Kodak image (left) along with two zooms (middle, and right). 
}
\label{fig:psnr}
\end{figure}

In \Cref{fig:otherMetrics}, we complete \Cref{fig:main_result} from the main paper with the PSNR metric computed on Kodak and COCO. 
We observe that \ours yields consistently better (higher) PSNR than Text-Sketch, but also consistently  worse (lower) than MS-ILLM. 
To understand the reason behind this relatively low PSNR, we also include  the PSNR of our LDM autoencoder (black horizontal line), which does not involve quantization. 
Interestingly, we observe that the LDM autoencoder  achieves a  PSNR performance comparable to MS-ILLM at a bitrate of 0.08 and VTM at a bitrate of 0.04. 
The PSNR of the LDM autoencoder presumably explains the lower PSNR performance we observe with \ours across different bitrates. 
To gain further insights into this behavior, we visually inspect the reconstruction for a Kodak image in \Cref{fig:psnr}. 
From the example, the LDM autoencoder seems to preserve more relevant detail and texture than the MS-ILLM reconstruction, despite the lower PSNR.

\begin{figure}
\def\myim#1{ \includegraphics[height=0.35\textwidth]{figs/appendix/lpips_loss/#1}}

\centering
\setlength\tabcolsep{0.5 pt}
\renewcommand{\arraystretch}{0.2}
\footnotesize

\begin{tabular}{cccc}
\myim{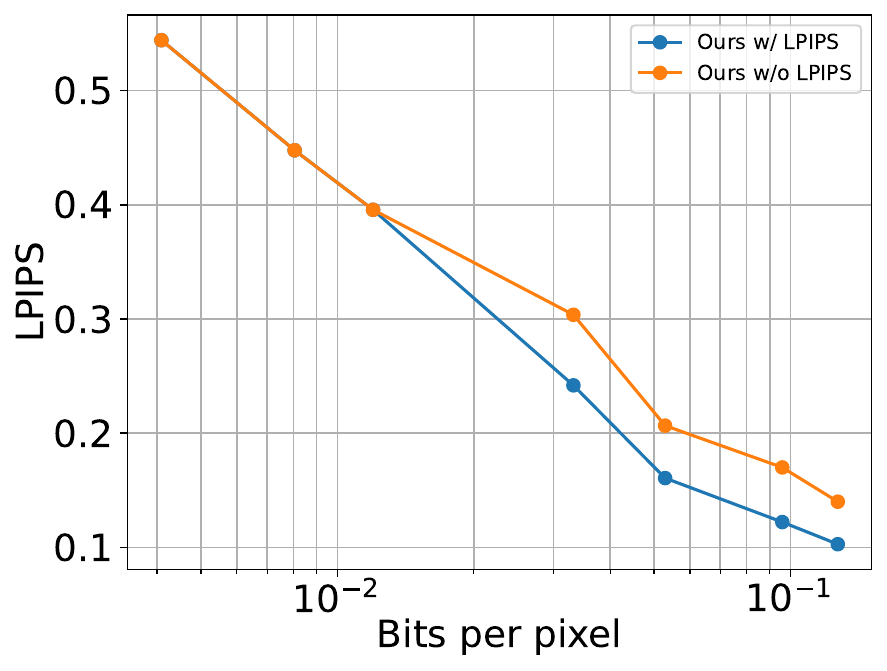} &   \myim{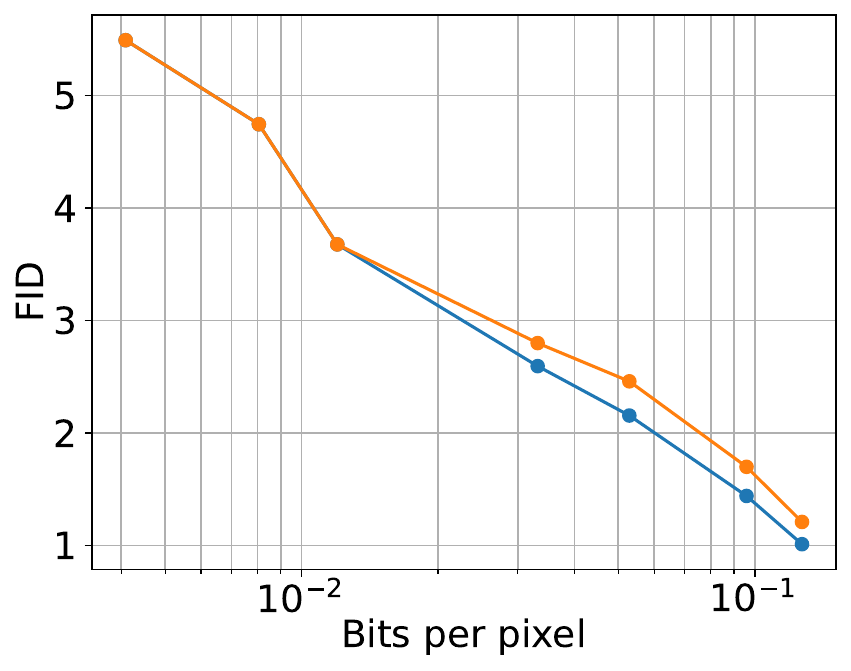}  \\ 
\end{tabular}
\caption{Ablation of adding the LPIPS loss to train the diffusion decoder.
}
\label{fig:abllpipsloss}
\end{figure}

In \Cref{fig:abllpipsloss}, we provide an ablation on the addition of the LPIPS loss, which shows that it leads to small improvements for both LPIPS and FID at higher bitrates.

\begin{figure}
\def\myim#1{ \includegraphics[width=0.35\textwidth]{figs/appendix/clip_image_metric/#1}}
\def\myimLeg#1{ \includegraphics[width=0.5\textwidth]{figs/plot_metrics/#1}}
\centering
\myimLeg{legend_stable1.pdf}\\
\myim{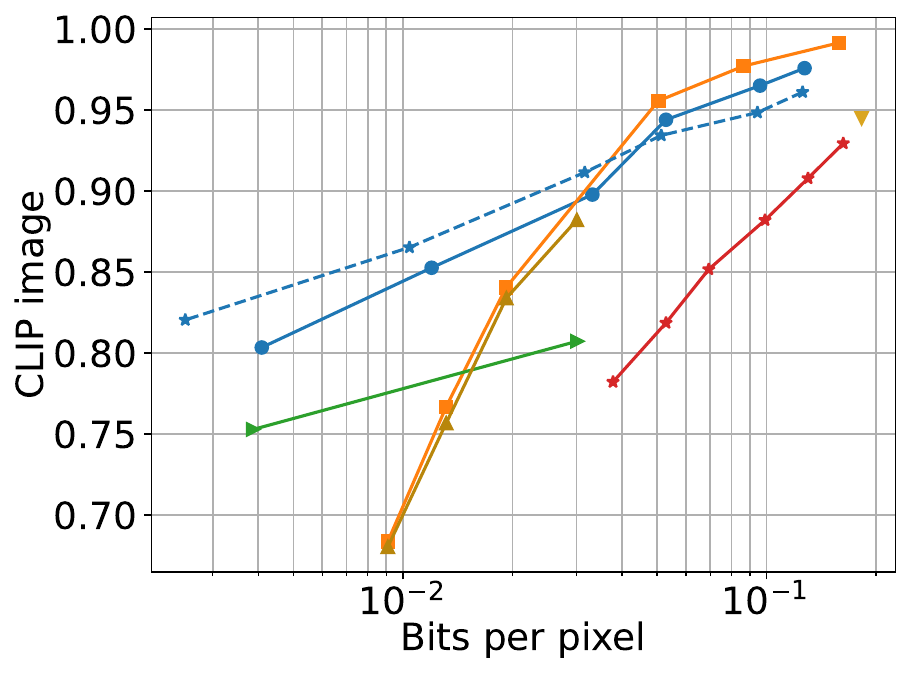}  
\caption{Evaluation in terms of the CLIP image metric.
}
\label{fig:clipimgmetric}
\end{figure}

In~\Cref{fig:clipimgmetric}, we provide evaluation on an additional global image similarity metric that consists in computing CLIP scores with a cosine similarity between CLIP image embeddings of original and reconstructed images, instead of cosine similarity between text captions and reconstructed images. This is motivated by the fact that we noticed that CLIP text score was saturating at high bitrates, as it reaches the upper bound of 30.4 when computed on real images instead of reconstructed ones, while CLIP image score threshold is 1 when evaluated on real images which is higher than what is obtained at high bitrates for all the baselines.

\subsection{Qualitative results}
\label{app:qualitative}

\begin{figure}
\def\myim#1{ \includegraphics[width=0.33\textwidth]{figs/appendix/ablation_text/#1}}

\centering
\setlength\tabcolsep{0.5 pt}
\renewcommand{\arraystretch}{0.2}
\tiny
\begin{tabular}{ccc}
Original & No text & 'A lighthouse with a white picket fence' (BLIP) \\
 &  \textbf{Spatial bpp: 0.0019, Textual bpp: 0.} & \textbf{Spatial bpp: 0.0019, Textual bpp: 0.0016} \\
\myim{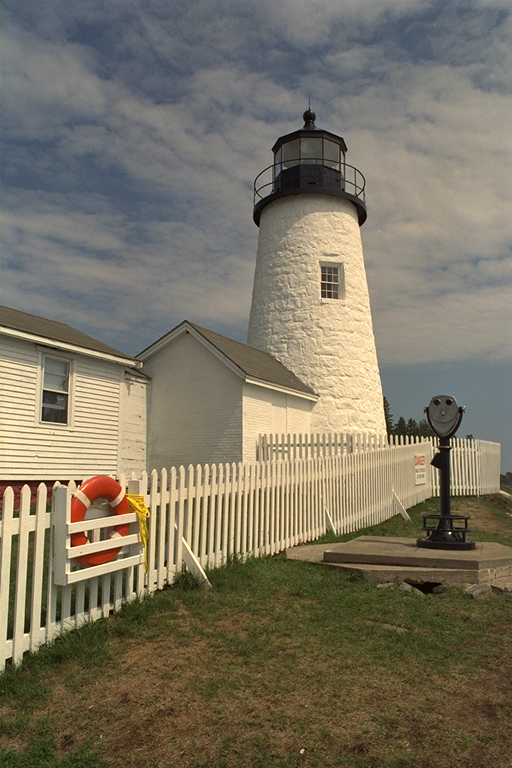} & \myim{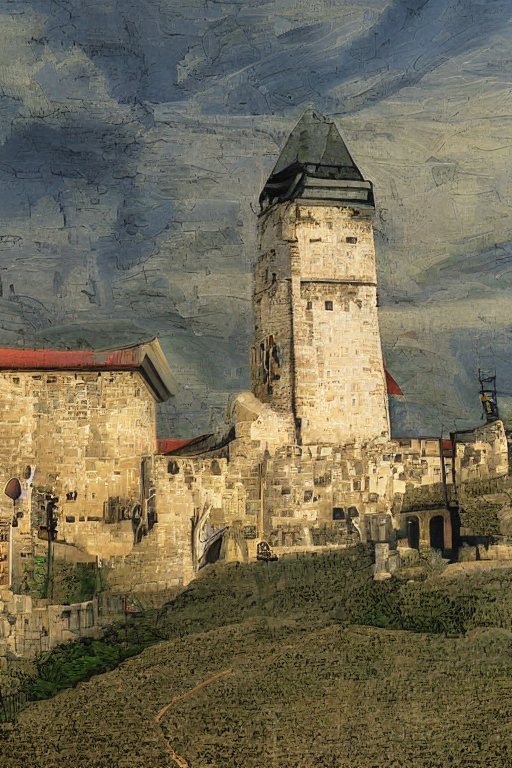} & \myim{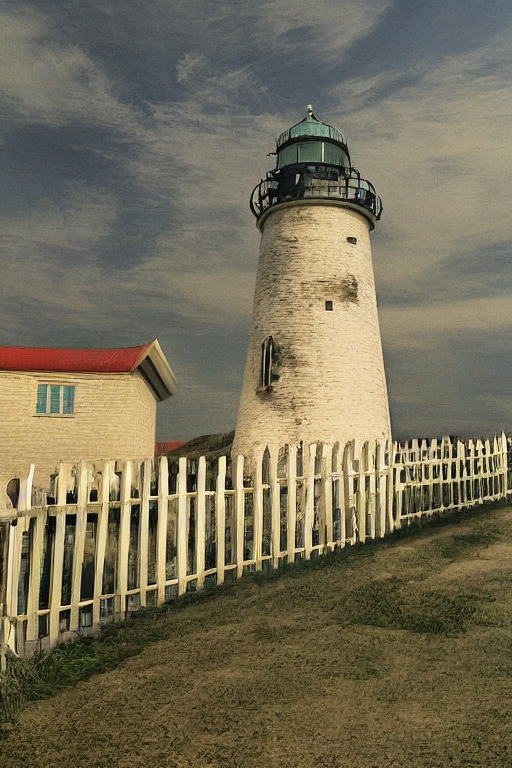} \\

 \multirow{2}{*}{'A lighthouse'} & 'A house and a red safety ring  &  \multirow{2}{*}{'An old castle'} \\
 & on a white picket fence with a lighthouse'&\\
 \textbf{Spatial bpp: 0.0019, Textual bpp: 0.0010} & \textbf{Spatial bpp: 0.0019, Textual bpp: 0.0020} & \textbf{Spatial bpp: 0.0019, Textual bpp: 0.0011} \\
  \myim{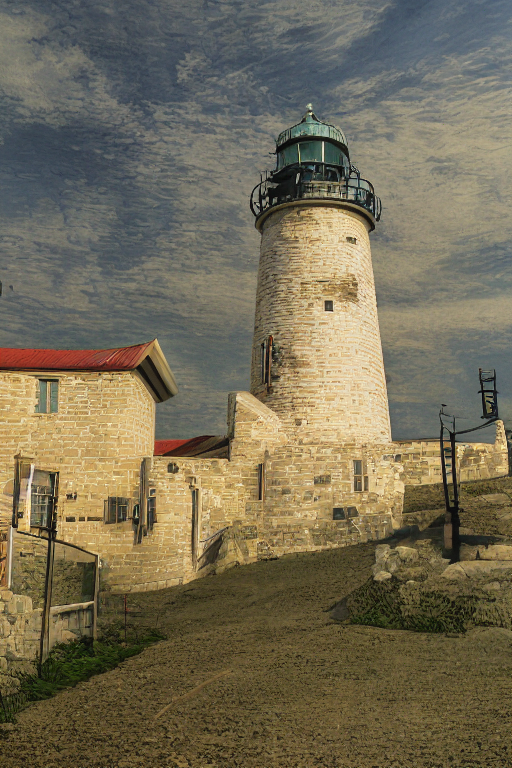} & \myim{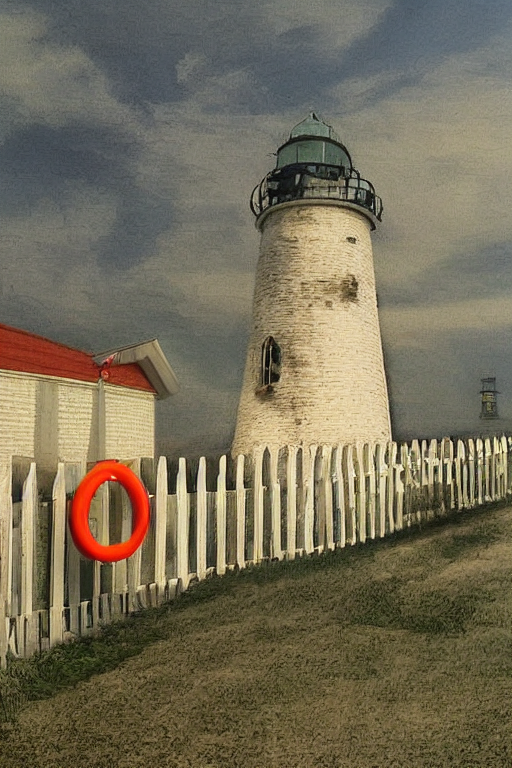}  &
\myim{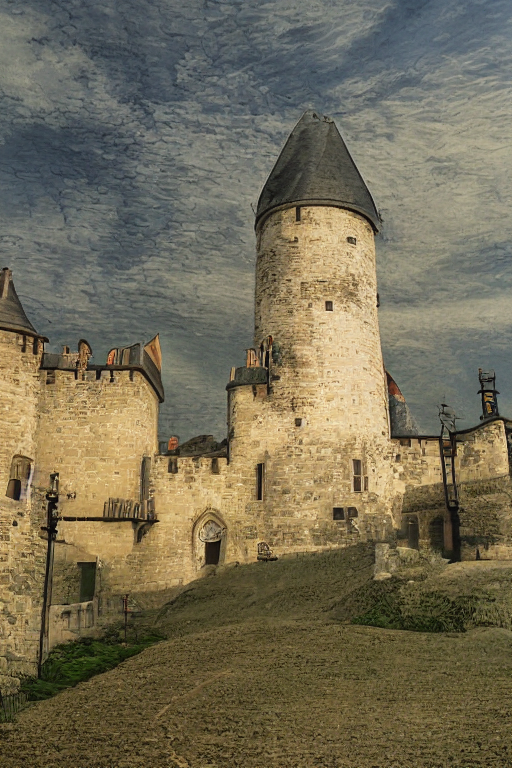} \\
No text & 'A lighthouse with a white picket fence' (BLIP) & 'An old castle' \\
\textbf{Spatial bpp: 0.1250, Textual bpp: 0.} & \textbf{Spatial bpp: 0.1250, Textual bpp: 0.0016}  & \textbf{Spatial bpp: 0.1250, Textual bpp: 0.0010}  \\
\myim{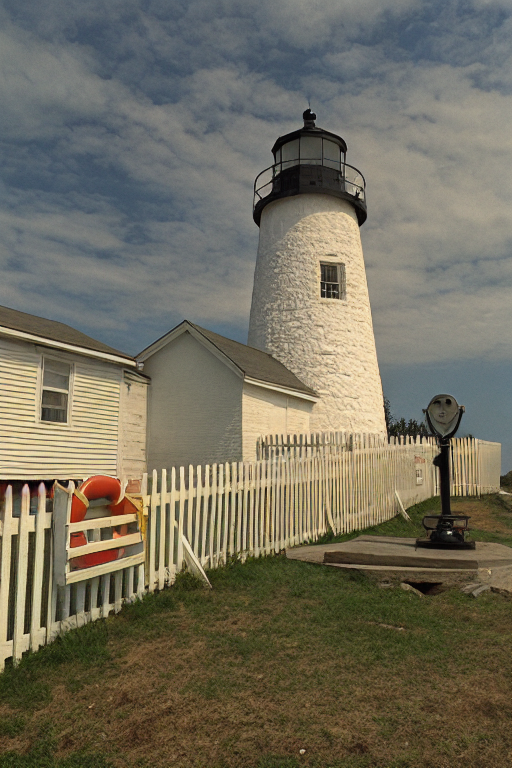} & \myim{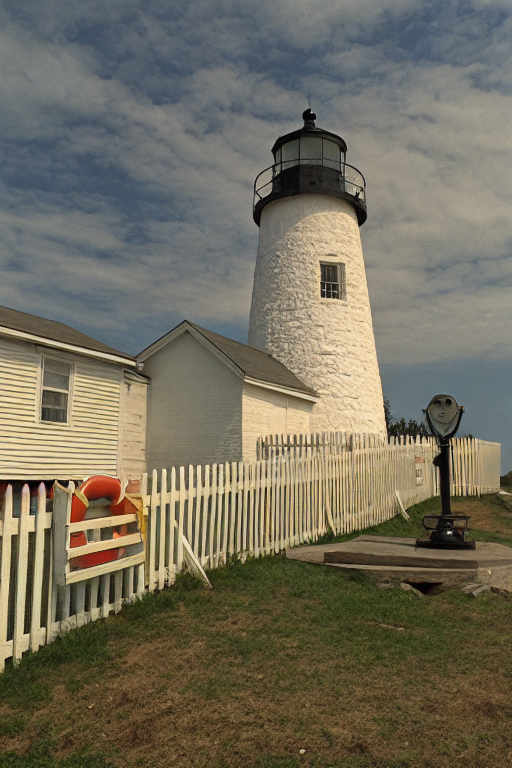} & \myim{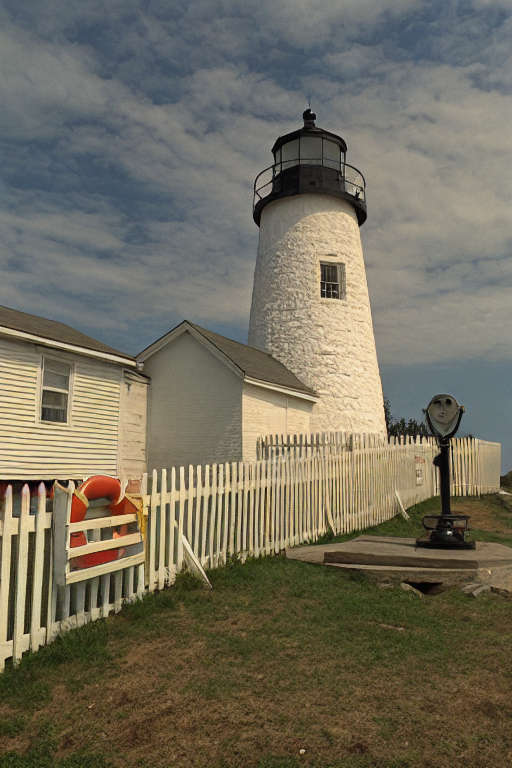}
\end{tabular}
\vspace{-4mm}
\caption{Reconstructions of a Kodak image  when varying the global textual conditioning with a 
 spatial bpp of 0.0019 (first two rows) and 0.1250 bpp (last row). 
Samples with the same spatial bpp are reconstructed from the same initial Gaussian noise from the same model.
}
\label{fig:text}
\end{figure}

 \mypar{Ablations on text}
In this ablation study, we conduct a detailed analysis to comprehend the effects of employing text as a global embedding. In \Cref{fig:text}, we present qualitative examples across two bitrates, where we manipulate the conditioning text.
At lower  bitrates for the spatial conditioning (first and second rows), we observe that textual conditioning has a significant influence on the reconstructed image. 
It can even alter the semantics  of the scene, such as transforming a lighthouse into a castle (second row, third column).
Furthermore, we note the positive impact of text conditioning, as image quality substantially deteriorates when text is not used (first row, second column). 
Conversely, when the bitrate of the local encoding is higher (last row), we notice that text has minimal influence on the reconstructed image.

\begin{figure}
\def\myim#1{ \includegraphics[width=0.33\textwidth,height=0.33\textwidth]{figs/appendix/vary_bpp/#1}}

\centering
\setlength\tabcolsep{0.5 pt}
\renewcommand{\arraystretch}{0.2}
\small
\begin{tabular}{ccc}
Original & 0.0048 bpp & 0.0087 bpp \\
\myim{} & 
\myim{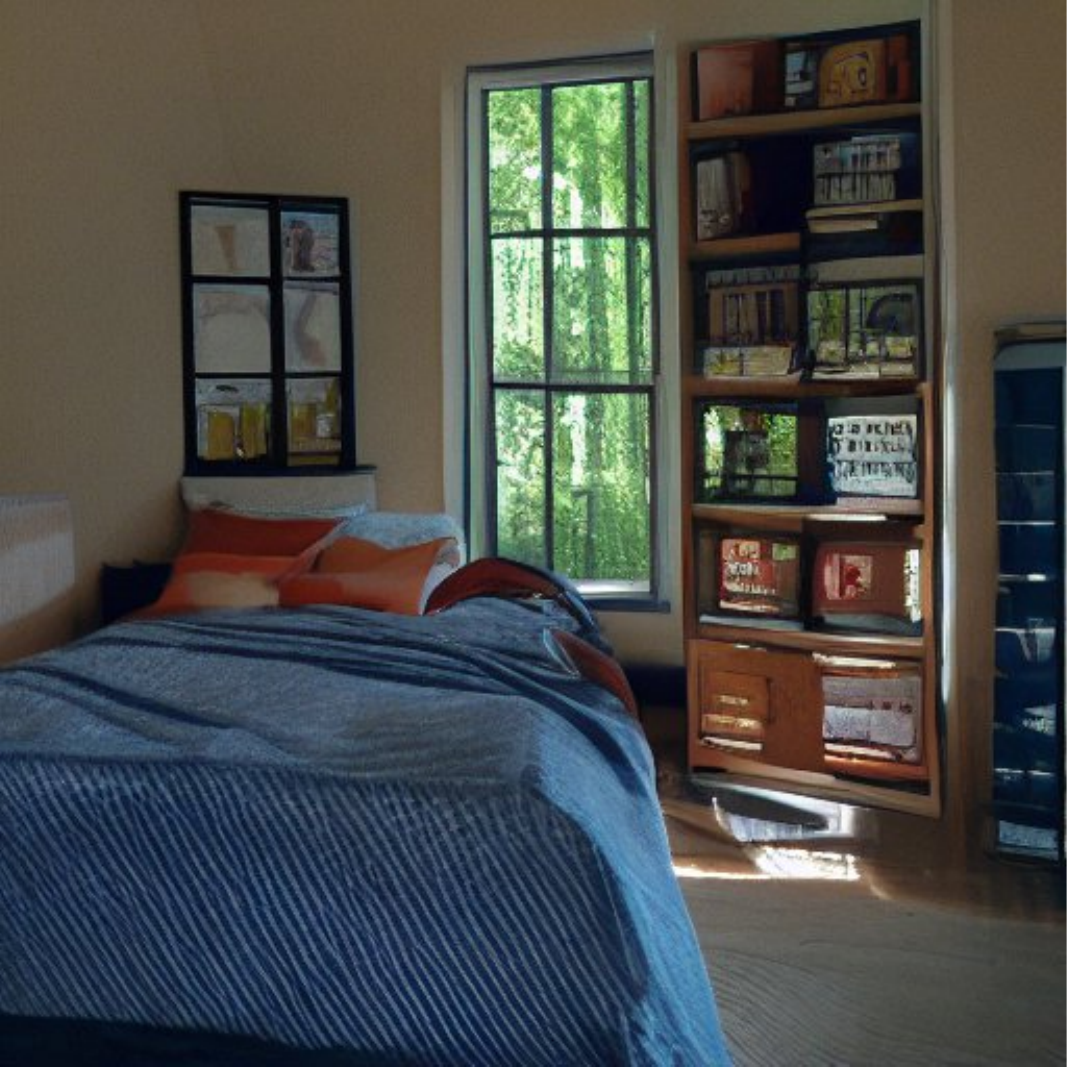} & \myim{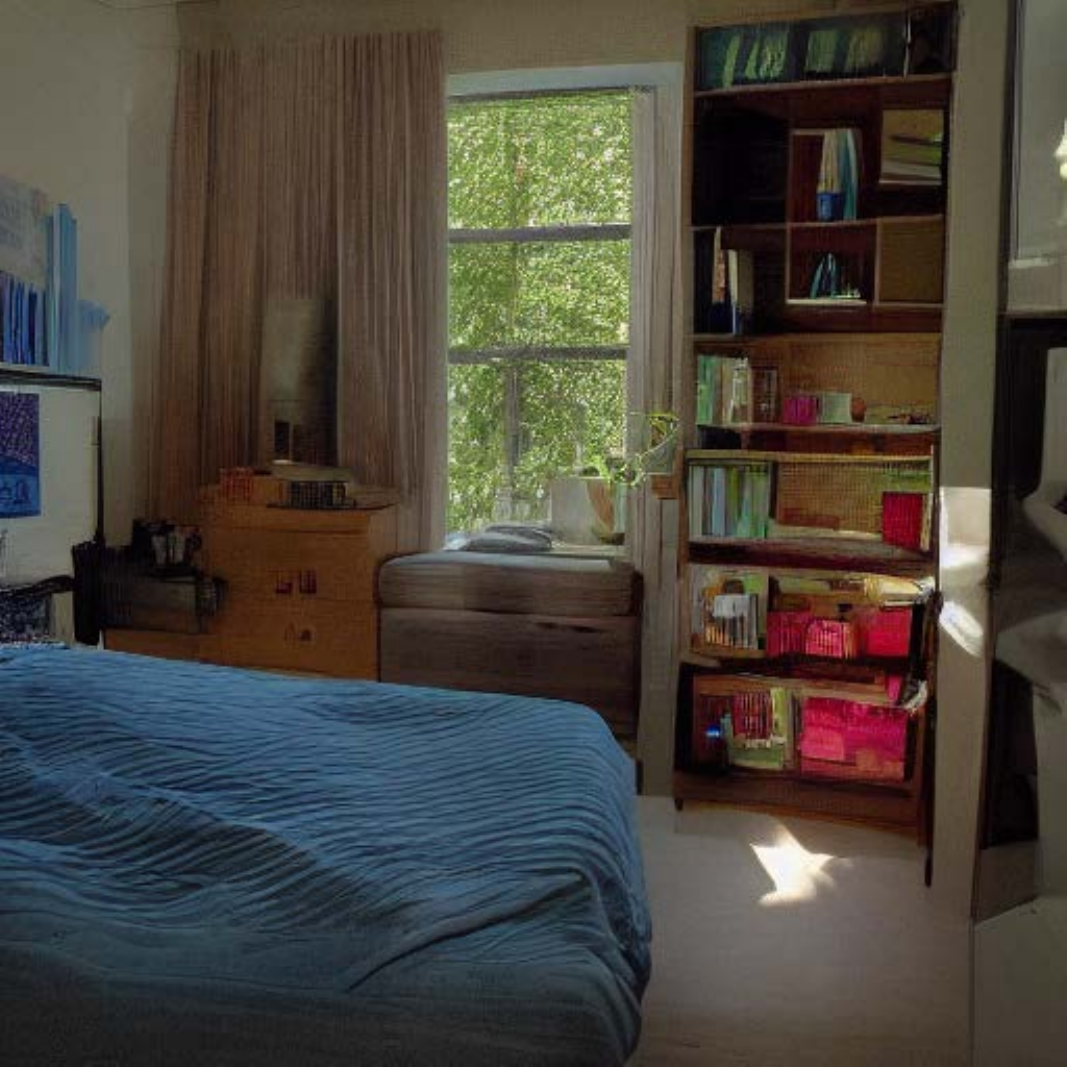} \\[1ex]
0.0127 bpp &0.0536 bpp & 0.1279 bpp \\
\myim{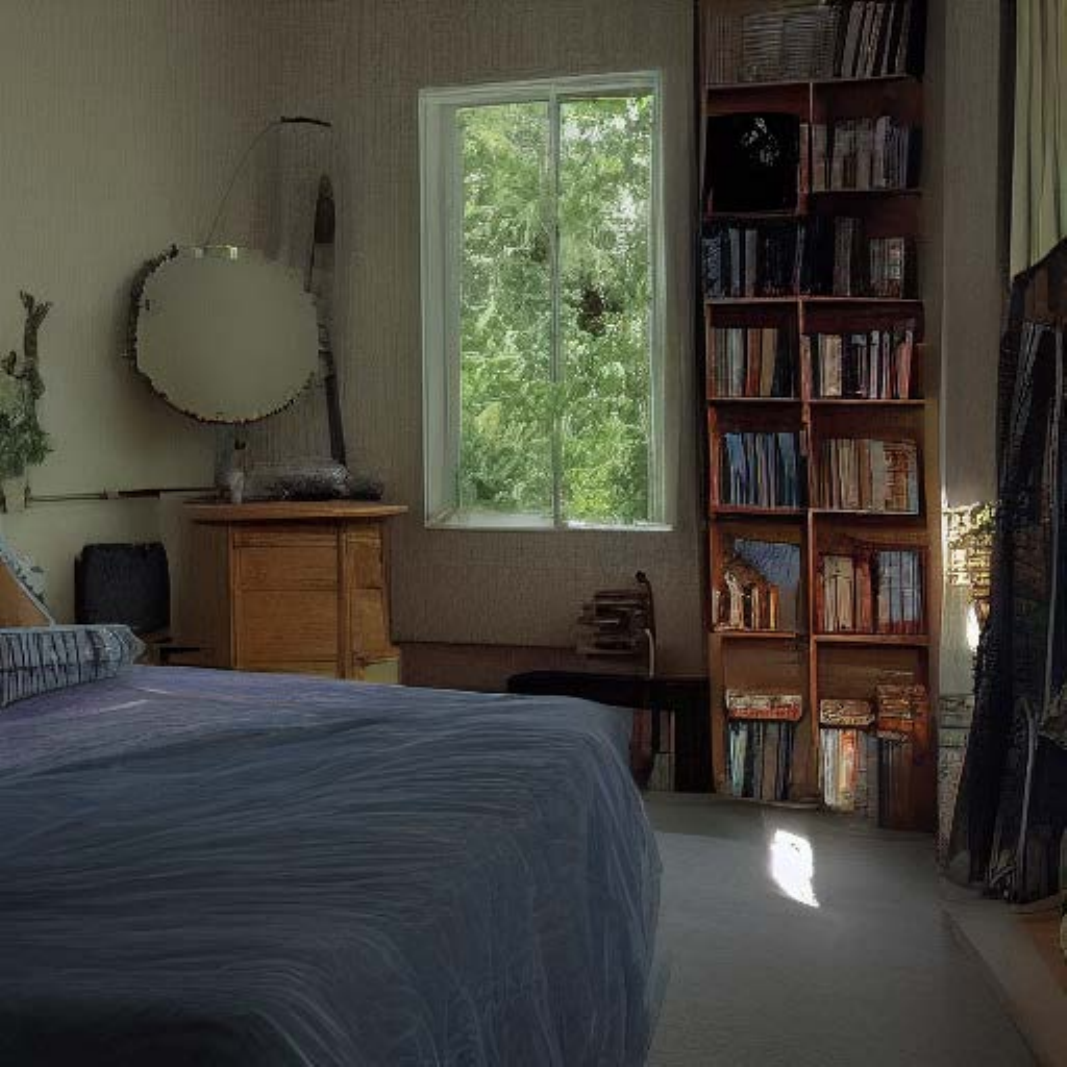}  &  \myim{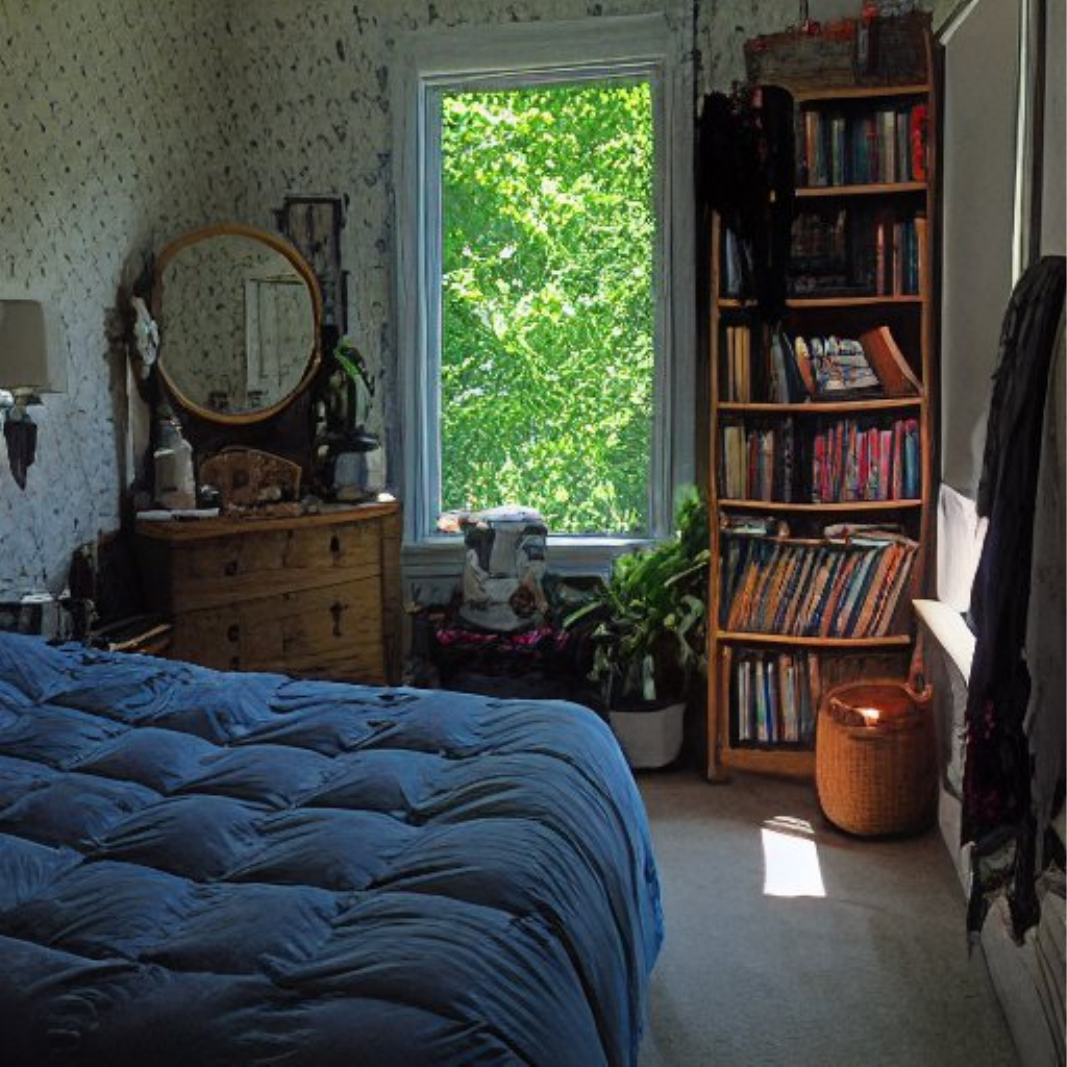}  &
\myim{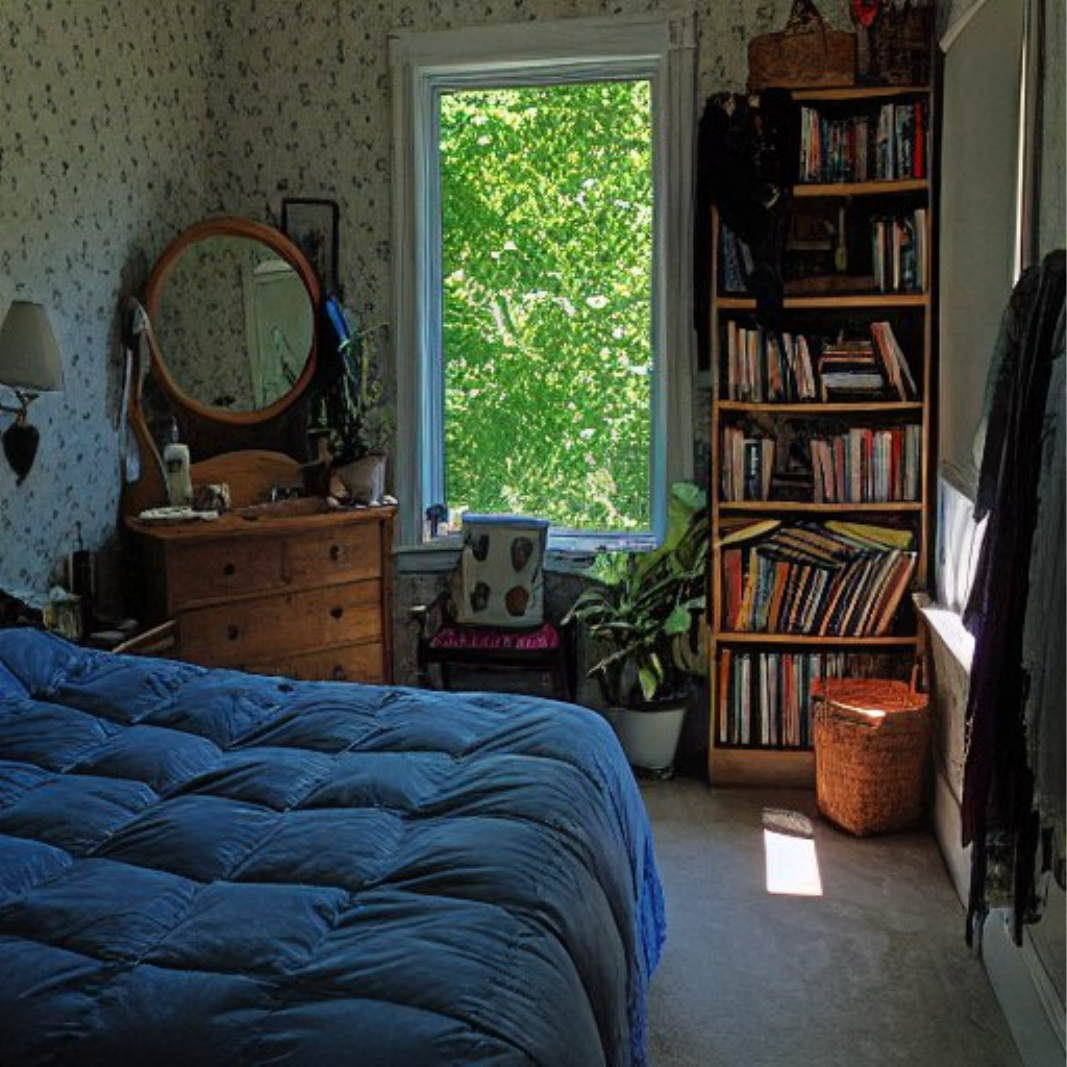}  \\[1ex]
Original & 0.0046 bpp & 0.0085 bpp \\
\myim{} & 
\myim{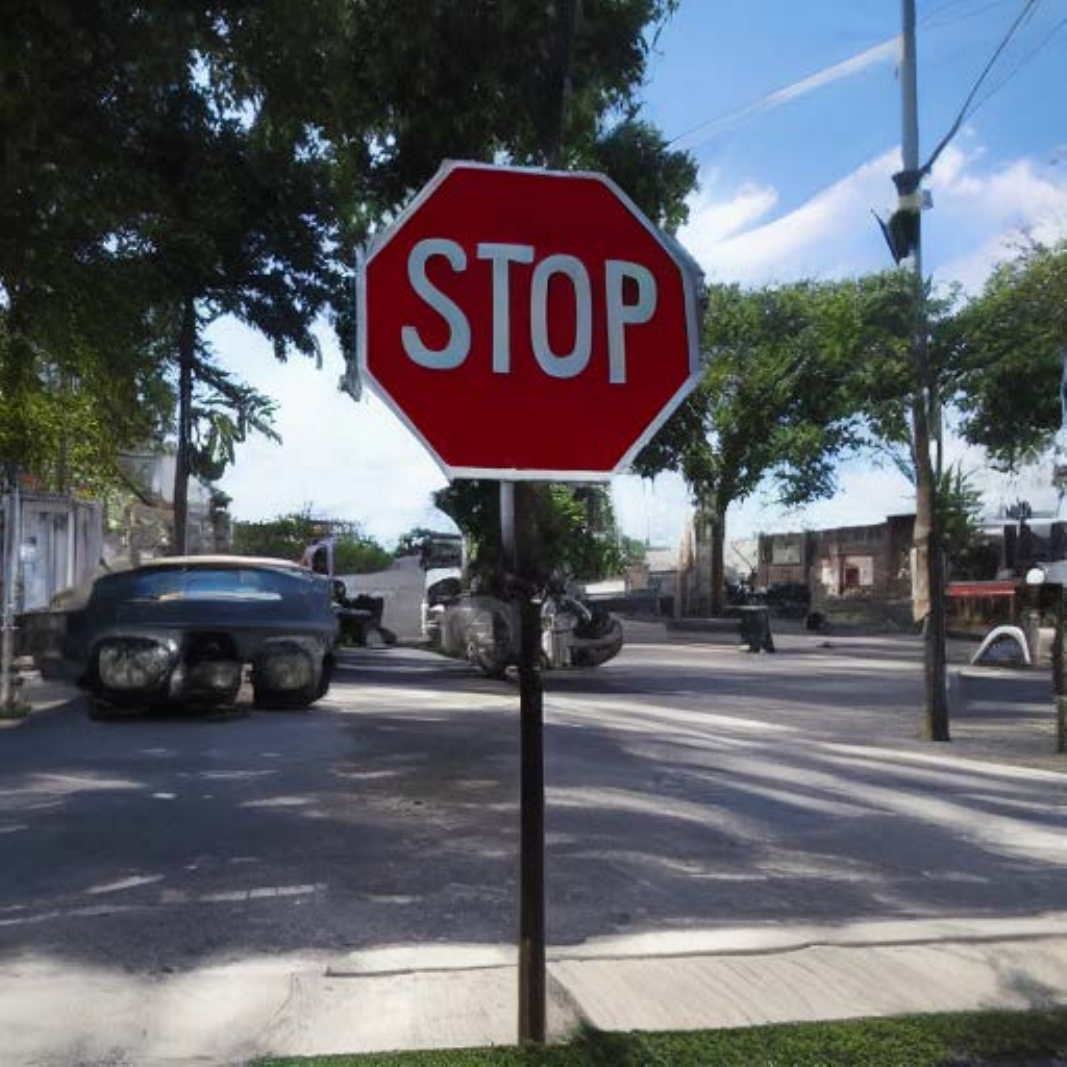} & \myim{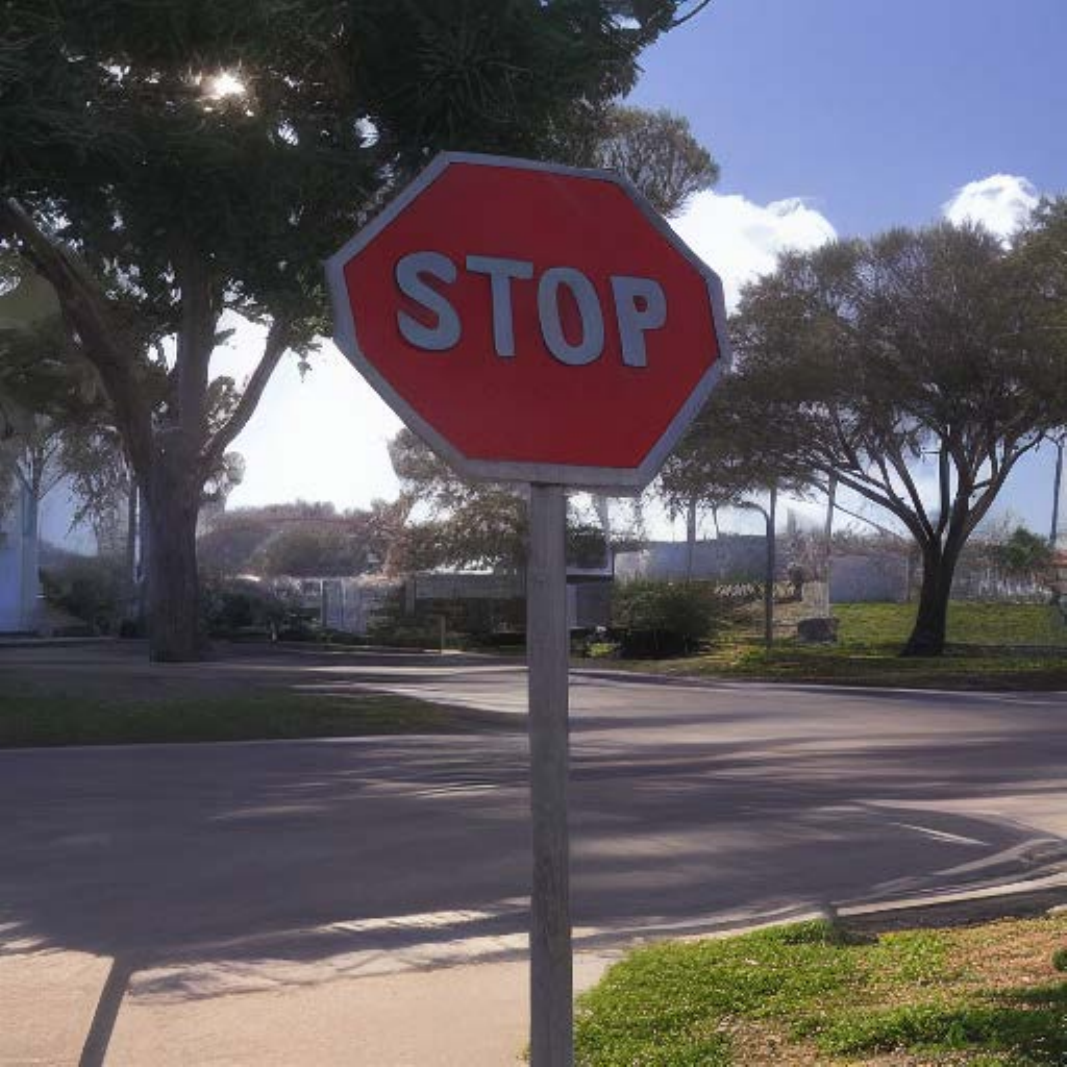} \\[1ex]
0.0125 bpp &0.0534 bpp & 0.1277 bpp \\
\myim{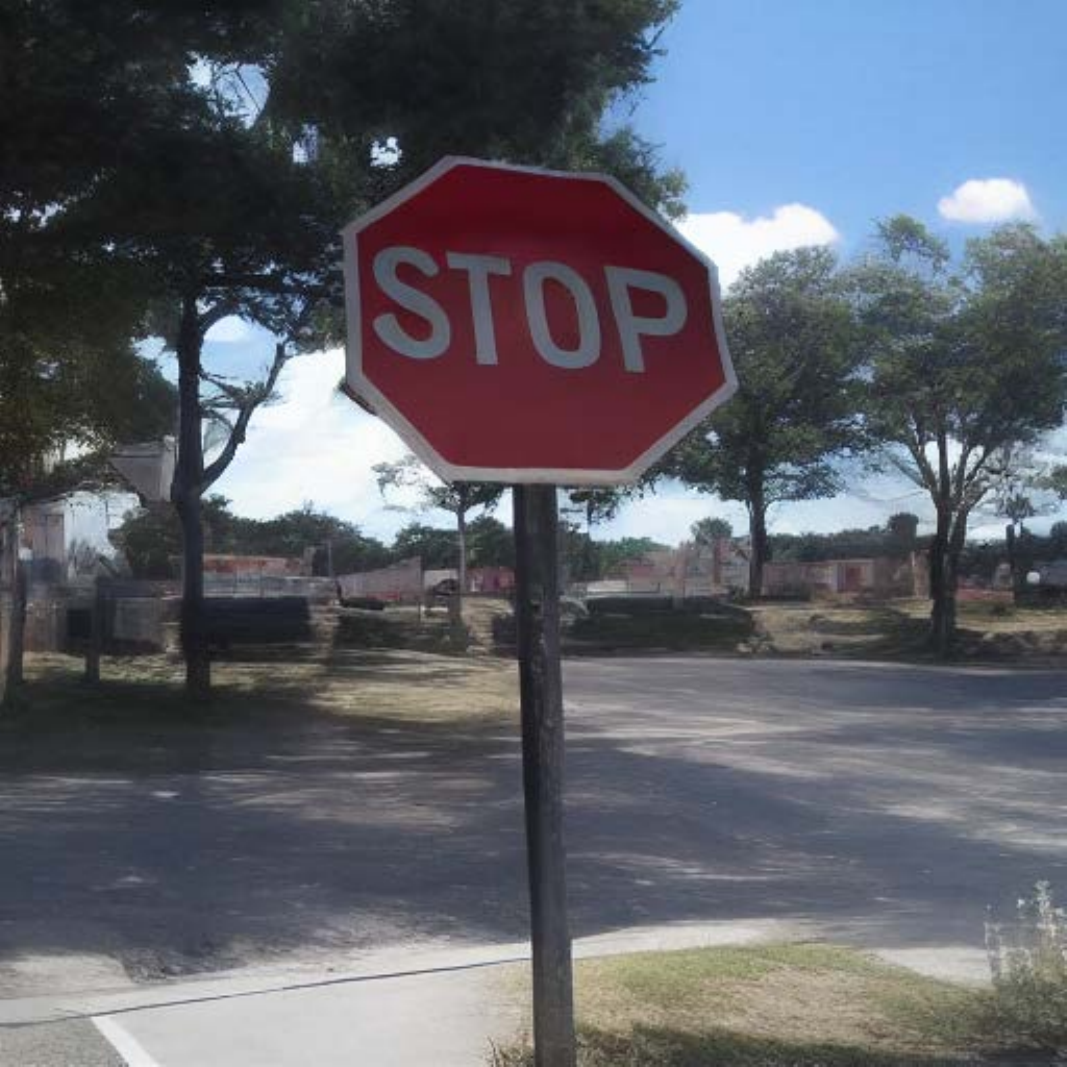}  &  \myim{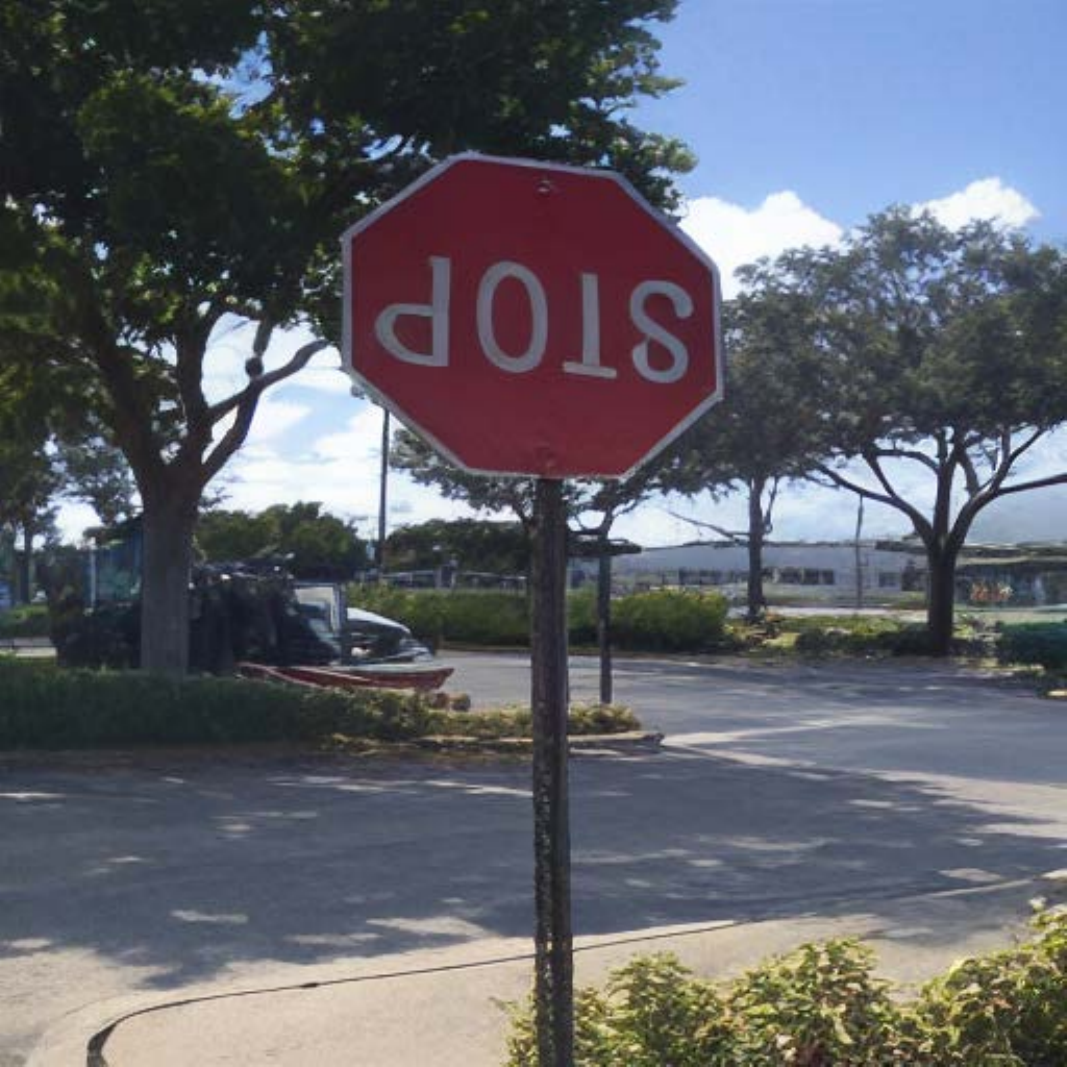}  &
\myim{}  \\

\end{tabular}

\caption{Reconstructions of two MS-COCO 30k images by \ours at different bitrates. 
Textual conditionings are \emph{``Bedroom scene with a bookcase, blue comforter and window''}  and \emph{``A stop sign is mounted upside-down on it's post''}, respectively. 
}
\label{fig:VarBitrateComp}
\end{figure}

 \mypar{Reconstruction at different bitrates}
In \Cref{fig:VarBitrateComp}, we show reconstructions of two MS-COCO 30k images  using \ours trained for different targeted spatial rates, ranging from 0.0046 to 0.1279 bpps. 
The image quality is quite stable across bitrates, while the similarity with the original images increases with the bitrate.

 \mypar{Qualitative analysis of LDM autoencoder reconstruction}
When computing the averaged similarity scores on Kodak dataset for the LDM autoencoder, we obtain 0.0775 LPIPS, 0.9131 MS-SSIM and 25.6003 PSNR which is lower than MS-ILLM at 0.1535 bpp for all the scores and lower than MS-ILLM at 0.0806 bpp on PSNR.
In \Cref{fig:psnr} we show a Kodak image illustrating that.
Even if the LDM autoencoder produces highly detailed images, we observe that reconstruction fidelity is somewhat worse than MS-ILLM at higher bitrates. Indeed, when looking at image details such as the waves, we notice that MS-ILLM exhibits some blurring at 0.0806 bpp while LDM produces sharp details for a similar PSNR. When comparing with MS-ILLM at the higher bitrate of 0.1535 bpp, we notice that the letters on the boat are less faithful for our LDM Autoencoder than MS-ILLM with respect to the original image.

\begin{figure}
\def\myim#1{\includegraphics[height=0.27\textwidth]{figs/appendix/psnr_quali/#1}}
\begin{center}
   \setlength\tabcolsep{1 pt}
   \renewcommand{\arraystretch}{0.2}
\scriptsize
\begin{tabular}{ccc}
 \myim{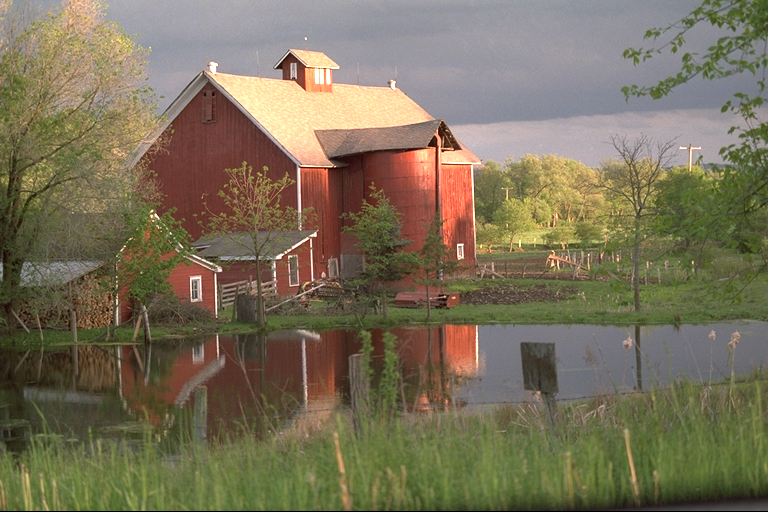} & \myim{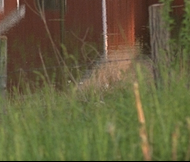} & \myim{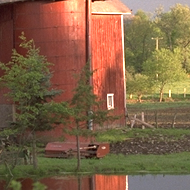} \\
   \multicolumn{3}{c}{Original} \\
 \myim{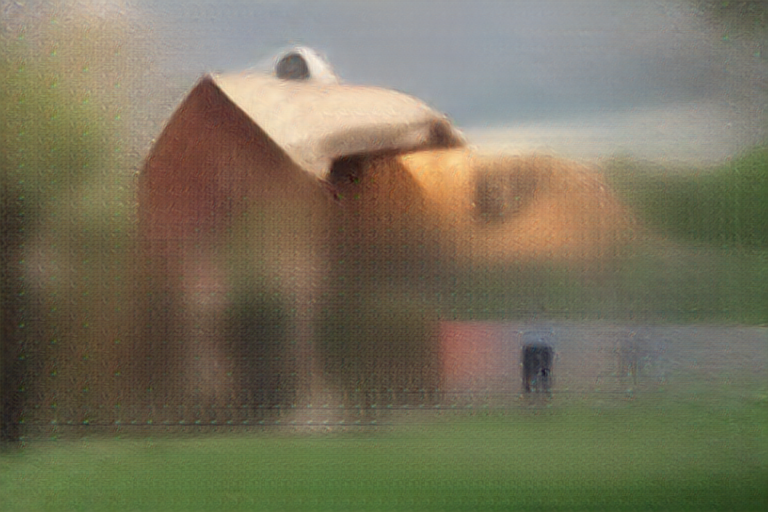} & \myim{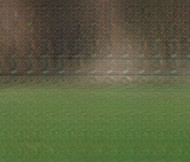} &\myim{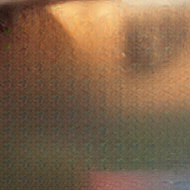} \\
   \multicolumn{3}{c}{MS-ILLM, 0.0065 bpp, 20.54 PSNR, 0.5672 MS-SSIM, 0.5372 LPIPS}   \\
    \myim{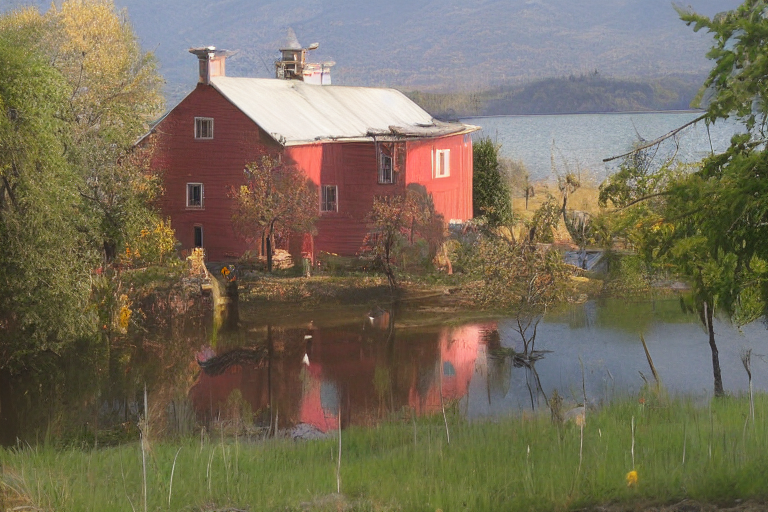} & \myim{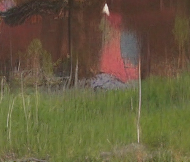} &\myim{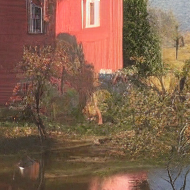} \\
    \multicolumn{3}{c}{\ours 0.0070 bpp, 18.49 PSNR, 0.4700 MS-SSIM, 0.5239 LPIPS} \\
  \myim{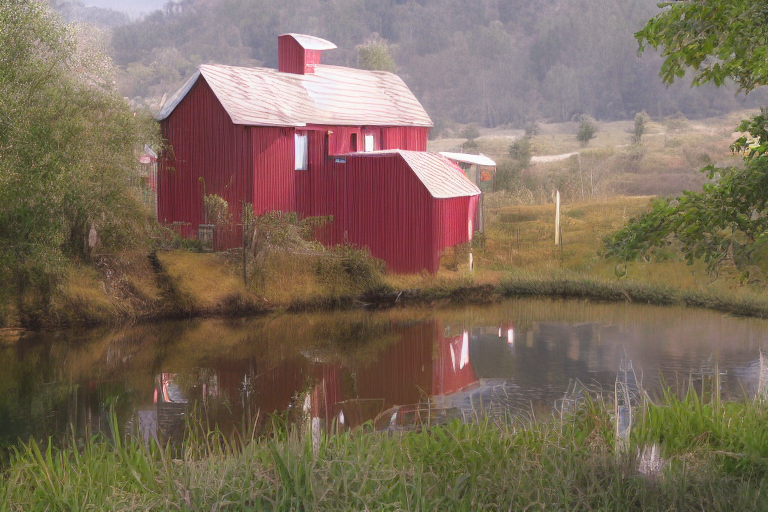} & \myim{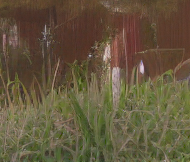} &\myim{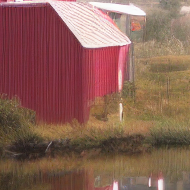} \\
   \multicolumn{3}{c}{ \ours 0.0031 bpp, 17.21 PSNR, 0.3906 MS-SSIM, 0.6222 LPIPS}
\end{tabular}
\end{center}     
\caption{Qualitative comparison of reconstructions using MS-ILLM  and \ours at low bitrates of a Kodak image (left) along with two zooms (middle, and right). 
MS-ILLM yields a very blurry recontruction at this operating point with \ours yields a more realistic reconstruction, yet \ours obtains  a substantially lower PSNR, MS-SSIM and LPIPS. 
}
\label{fig:psnr2}
\end{figure}

 \mypar{Qualitative analysis of pixelwise similarity metrics}
We observed that pixelwise distortion metrics such as PSNR, MS-SSIM and LPIPS are not very meaningful at very low bitrates, as already pointed out by~\cite{lei23ncw}. In the example in \Cref{fig:psnr2}, \ours at  0.0031 bpp is 3.33 points worse in PSNR, 0.1766 points worse in MS-SSIM and 0.085 points worse in LPIPS than MS-ILLM at  0.0065 bpp. However, at these very low bit-rates \ours yields a much sharper and realistic reconstruction than MS-ILLM, and  preserves most of the semantics of the scene.
These examples provide some indication  that these metrics do not  align well with  human preference for image quality at very low bitrates.

\begin{figure}
\def\myim#1{ \includegraphics[width=0.33\textwidth,height=0.22\textwidth]{figs/appendix/comp_highbitrate/#1}}

\centering
\setlength\tabcolsep{0.5 pt}
\renewcommand{\arraystretch}{0.2}
\small
\begin{tabular}{ccc}
Original & VTM (0.1552 bpp) & HifiC (0.18 bpp) \\
\myim{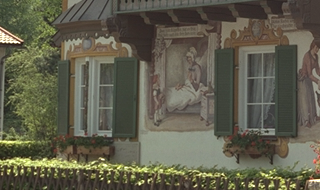} & 
\myim{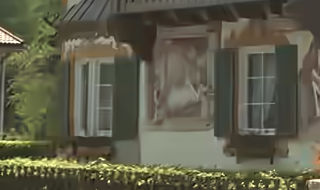} & \myim{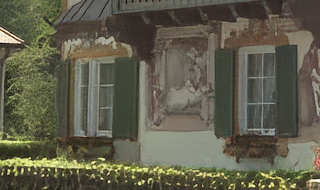} \\[1ex]
&MS-ILLM (0.15347 bpp) & Ours (0.1265 bpp) \\
&  \myim{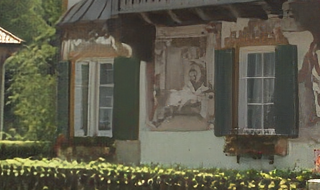}  &
\myim{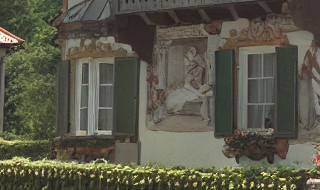} \\
\end{tabular}

\vspace{4mm}

\begin{tabular}{ccc}
Original & VTM (0.1552 bpp) & HifiC (0.18 bpp) \\
\myim{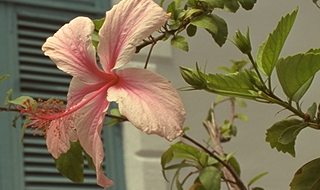} & 
\myim{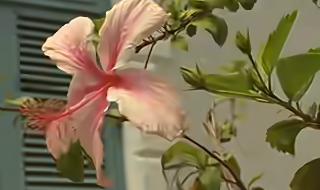} & \myim{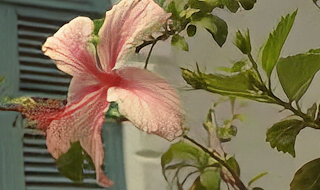} \\[1ex]
&MS-ILLM (0.15347 bpp) & Ours (0.1264 bpp) \\
&  \myim{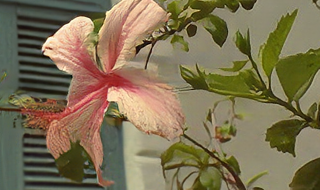}  &
\myim{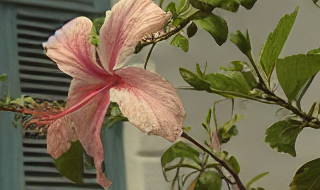} 
\end{tabular}
\caption{Reconstructions of Kodak images at the highest bitrate range we evaluated models on.
}
\label{fig:highBitrateComp}
\end{figure}

 \mypar{Additional comparisons to baselines}
In  \Cref{fig:highBitrateComp}, we compare \ours at  0.1264 bpp, the highest bitrate we trained on,  to VTM, HiFiC and MS-ILLM at similar bitrates. 
At this bitrate, \ours   faithfully reconstructs the original images with similarly detailed textures as HiFiC and MS-ILLM, and fewer artifacts than VTM.

\clearpage

\end{document}